\documentclass[twocolumn]{article}
\pdfoutput=1
\usepackage[square,sort,comma,numbers]{natbib}

\usepackage{authblk}
\usepackage[utf8]{inputenc} 
\usepackage[T1]{fontenc}    
\usepackage{hyperref}       
\usepackage{url}            
\usepackage{booktabs}       
\usepackage{amsfonts}       
\usepackage{nicefrac}       
\usepackage{geometry}
\usepackage{titlesec}

\titlespacing{\paragraph}{%
  0pt}{
  0.4\baselineskip}{
  1em}

\usepackage{subcaption}
\usepackage[ruled,vlined]{algorithm2e}
\usepackage{graphicx}

\usepackage{booktabs} 
\usepackage[utf8]{inputenc} 
\usepackage[T1]{fontenc}    
\usepackage{url}            
\usepackage{booktabs}       
\usepackage{amsfonts}       
\usepackage{nicefrac}       
\usepackage{microtype}      
\usepackage[makeroom]{cancel}
\usepackage{kky}
\usepackage[super]{nth}
\usepackage[makeroom]{cancel}

\usepackage{amssymb}
\usepackage{amsthm}
\usepackage{enumerate}
\usepackage{float}
\usepackage{bm}
\usepackage{graphicx}

\usepackage{multirow}
\usepackage{dirtytalk}
\usepackage{array}
\usepackage{soul}
\usepackage{xspace}
\usepackage{xcolor}
\usepackage{lipsum}
\usepackage[inline]{enumitem}

\usepackage[mathscr]{eucal}
\theoremstyle{plain}
\newtheorem*{assumption*}{Assumption}  
\newtheorem{assumption_s}{Assumption}  
\newtheorem*{theorem*}{Theorem}
\newtheorem*{proposition*}{Proposition}  
\newtheorem{proposition_s}{Proposition}[section]  
\newtheorem*{proposition 1*}{Proposition 1}
\newtheorem*{lemma*}{Lemma}  
\newtheorem{lemma_s}{Lemma}[section]  
\usepackage{bm}

\newcommand{\set}[1]{\mathbb{#1}}  

\usepackage{accents}
\newcommand*{\dt}[1]{%
    \accentset{\mbox{\large\bfseries .}}{#1}}
\newcommand{\ours}{GLOCAD\xspace} 
\newcommand{\autoours}{Auto-GLOCAD\xspace}
\DeclareMathOperator*{\expectation}{\mathbb{E}}
\newcommand\pder[2][]{\ensuremath{\frac{\partial#1}{\partial#2}}} 

\usepackage{cleveref}
\usepackage{verbatim}
\let\oldcdot\cdot
\usepackage{breqn}
\let\cdot\oldcdot

\newcommand{\wjsay}[1]{[\textcolor{red}{\textbf{WJ}}: {\color{green!60!black}{#1}}]}

\newcommand{\kri}[1]{{\color{blue}{Krikamol: #1}}}

\newcommand{\ourtitle}[0]{Kernel-Guided Training of Implicit Generative Models \\with Stability Guarantees}

\title{\ourtitle}
\date{\vspace{-5ex}}

\author[1,2]{Arash Mehrjou\thanks{\texttt{amehrjou@tuebingen.mpg.de}}}
\author[1]{Wittawat Jitkrittum\thanks{\texttt{wittawat@tuebingen.mpg.de}}}
\author[1]{Krikamol Muandet\thanks{\texttt{krikamol@tuebingen.mpg.de}}}
\author[1]{Bernhard Sch{\"o}lkopf\thanks{\texttt{bs@tuebingen.mpg.de}}}
\affil[1]{Max Planck Institute for Intelligent Systems }
\affil[2]{ETH Z\"{u}rich}

\begin{document}
\raggedbottom
\maketitle

%

%

\begin{abstract}
  Modern implicit generative models such as generative adversarial networks (GANs) are generally known to suffer from issues such as instability, uninterpretability, and difficulty in assessing their performance. If we see these implicit models as dynamical systems, some of these issues are caused by being unable to control their behavior in a meaningful way during the course of training. In this work, we propose a theoretically grounded method to guide the training trajectories of GANs by augmenting the GAN loss function with a kernel-based regularization term that controls local and global discrepancies between the model and true distributions. This control signal allows us to inject prior knowledge into the model. We provide theoretical guarantees on the stability of the resulting dynamical system and demonstrate different aspects of it via a wide range of experiments.
\end{abstract}

\section{Introduction}
\label{sec:intro}

%
%
Generative adversarial networks (GANs)~\citep{goodfellow_gan_2014} have been widely studied from different perspectives such as information theory~\citep{chen2016infogan}, dynamical systems~\citep{nagarajan2017gradient, MehSch18}, and game theory~\citep{daskalakis2018training}. 
%
Despite the practical successes~\citep{karras2018progressive, zhu2017unpaired, li2017adversarial}, they have suffered from a lack of theoretical understanding of what is actually learned by the generator and how close the distribution of generated samples is to the real one~\citep{arora2017gans}. Moreover, the instability of training GANs and its reliance on immense hyper-parameter tuning also remains a major issue. On the other hand, classical generative algorithms based on kernel methods have well-grounded theories even though their performance has not yet been comparable with GANs~\citep{li2015generative, dziugaite2015training}. This brings us the possibility of combining the best of both worlds~\citep{li2017mmd}.

\paragraph{Previous work.}
Guided training of implicit generative models is mostly seen as conditional models or imposing some sort of inductive bias on the model. In conditional models, some aspects of the prior knowledge (e.g., class label) are provided as the input to the model~\cite{mirza2014conditional, brock2018large, dai2017towards}. However, while imposing inductive bias, structural domain knowledge (e.g., the existence of discrete classes) is incorporated into the model mostly in heuristic ways~\cite{chen2016infogan, hu2018deep}. 
In contrast, our approach incorporates the prior knowledge by controlling the training trajectories using \emph{witness points} whose role is theoretically grounded thanks to the transparency of kernel methods. One might see this under the general umbrella of adding more players to the adversarial training to improve some aspects of training. For instance, \cite{NeyBhoCha2017} proposed the use of multiple discriminators each of which discriminates data distribution projected on a random axis. 
In practice, this requires a sufficiently large number of projections to ensure stability. 

Our approach, on the other hand, takes advantage of \emph{trainable} witness points which make it possible to recover the distribution with much fewer projections. 
It is based on an \emph{unnormalized mean embedding} (UME) which under mild assumptions and any finite number of witness points defines an almost-sure metric on the space of probability measures \cite[Theorem 2]{ChwRamSejGre2015}. 
Another related work is ~\cite{denton2015deep} which proposed to train multiple GANs for different resolutions of images created by the Laplacian pyramid. The pyramid architecture and the resolutions are hard-coded and the algorithm has no freedom to learn its own pyramid. 
Analogously, our proposed witness points can be viewed as an \emph{adaptive} pyramid. 



\paragraph{Our contribution.}
In this work, after reviewing necessary background information (\Cref{sec:background}), we propose a generic game (dynamical system) with new players called \emph{witness points}~\citep{ChwRamSejGre2015,JitSzaChwGre2016} which can be combined with several objectives of implicit generative models to guide the training (\Cref{sec:method}).
This combination provides an explicit trade-off between \emph{global} and \emph{local} differences between the real and generated distributions. This disentanglement between global and local distances enables witness points to guide the training trajectories towards regions of interests in the data space. Hence, one can also incorporate prior knowledge about the true distribution during training using the witness points.
Based on our empirical studies, this prior knowledge carried by the witness points leads to a better convergence of the learning algorithms of GANs by favoring missed regions with stronger training signals.
Witness points also improve the interpretability of the models
by giving a better understanding of the aspects of the target distribution that are missed by the generator. We provide theoretical guarantees (\Cref{sec:theory}) to ensure that adding witness points does not harm the local stability of the model and studies different aspects of it both analytically and experimentally (\Cref{sec:experiments}). Since witness points are generally defined in the data space, and due to the expected challenge in using conventional kernels (e.g., Gaussian) to compute the similarity matrix in the data space, we suggest an idea to deal with high-dimensional data by defining witness points in an abstract low-dimensional latent space associated with an autoencoder (Section \ref{subsec:latent_space}). Lastly, we provide theoretical stability guarantee for this general setting in which the kernel is also trainable (Section \ref{sec:theory}).

\section{Background}
\label{sec:background}
GANs are among the most popular techniques for learning high-dimensional generative models \citep{goodfellow_gan_2014}. The idea is based on an adversarial game between two players, namely, a \emph{generator} and a \emph{discriminator}. Let $\mathcal{Z}$ and $\mathcal{X}$ be latent space and data space, respectively. The generator $G_{\thetabm}:\mathcal{Z}\to\mathcal{X}$ is defined via a generative process $\ybm = G_{\thetabm}(\zbm)$ for $\zbm \sim P_\Zb$ where $G_{\thetabm}$ is usually a feedforward neural network parameterized by $\thetabm$ and $P_\Zb$ is a simple distribution over the latent space, e.g., $P_\Zb=\mathcal{N}(\mathbf{0},\mathbf{I})$ for $\mathcal{Z}=\mathbb{R}^{d_\zbm}$. Hence, it is easy to sample from $G_{\thetabm}$ as it involves only one forward pass through the generator network. The discriminator $D: \mathcal{X}\to\mathbb{R}$ takes a data point $\xbm$ as an input and then outputs a score $D(\xbm)$, e.g., via a classifier $D:\mathcal{X}\to [0, 1]$. 

Given a dataset $D = \{\xbm_1,\ldots,\xbm_n\}$ of i.i.d. samples from $P_\Xb$, the vanilla GAN \citep{goodfellow_gan_2014} learns the generator $G_{\thetabm}$ to approximate $P_\Xb$ by solving the \emph{minimax} optimization problem:
$\min_{G}\max_{D}\; \mathbb{E}_{\xbm}[\log\, D(\xbm)] + \mathbb{E}_{\zbm}[\log(1 - D(G(\zbm)))]$.
Let $Q_\Yb$ be the distribution induced by the generator $G_\thetabm$. Then, it was shown that this problem is equivalent to minimizing the Jensen-Shannon (JS) divergence between $P_\Xb$ and $Q_\Yb$ \citep{goodfellow_gan_2014} which was later extended to more generic distances \citep{NowCseTom2016}.

\paragraph{Maximum mean discrepancy (MMD).}
In this work we focus on the MMD as discrepancy measure between $P_\Xb$ and $Q_\Yb$ \citep{GreBorRasSchSmo2012}. Let $\mathcal{F}$ be a reproducing
kernel Hilbert space (RKHS) defined by the positive definite kernel
$k\colon\mathcal{X}\times\mathcal{X}\to\mathbb{R}$ with the canonical
feature map $\phi\colon\mathcal{X}\to\mathcal{F}$, i.e., $\phi(\xbm) \coloneqq k(\xbm,\cdot)$ and $k(\xbm,\ybm)=\left\langle \phi(\xbm),\phi(\ybm)\right\rangle _{\mathcal{F}}$
for all $\xbm,\ybm\in\mathcal{X}$. The MMD between $P_\Xb$
and $Q_\Yb$ can be defined as 
\begin{align}
    \mathrm{MMD}^{2}(P_\Xb,Q_\Xb) 
    &= \sup_{f\in\mathcal{F},\|f\|\leq 1}\, \left | \int f dP_\Xb - \int f dQ_\Yb \right | \nonumber \\
    &= \left\|\mu_{P_\Xb}-\mu_{Q_\Yb}\right\|_{\mathcal{F}}^{2},\label{eq:mmd} 
\end{align}
where $\mu_{P_\Xb}\coloneqq \mathbb{E}_{\xbm\sim P_\Xb}[k(\xbm,\cdot)]$ and $\mu_{Q_\Yb}\coloneqq \mathbb{E}_{\ybm\sim Q_\Yb}[k(\ybm,\cdot)]$
are the so-called mean embeddings of $P_\Xb$ and $Q_\Yb$ \citep{MuaFukSriSch17}. For a characteristic kernel $k$ \citep{fukumizu,GreBorRasSchSmo2012}, 
$\mathrm{MMD}^{2}(P,Q) = 0$ iff $P_\Xb=Q_\Yb$, i.e., the MMD is a proper metric on a space of distributions. It is instructive to note that the MMD can be viewed as an integral probability metric (IPM) whose function class is a unit ball in the RKHS associated with the kernel $k$ \citep{Mueller97:IPM, NowCseTom2016, arjovsky2017wasserstein}. In the following, we discuss briefly how MMD has been used in generative models to prepare for the presentation of our proposed method in Section~\ref{sec:method}.

\paragraph{MMD-GAN.}
It was originally proposed in \cite{dziugaite2015training} and \cite{li2015generative} that the generator $Q_\Yb=G_\thetabm(\Zb)$ is learned so as to minimize $\mathrm{MMD}^{2}(P_\Xb,Q_\Yb)$ w.r.t. $\thetabm$. Since the discriminator lives in a unit ball of the RKHS, the benefit of this formulation is that the maximization problem w.r.t. the discriminator can be solved analytically, as seen from~\eqref{eq:mmd}.
In higher dimensions, however, the MMD-GAN usually produces a generator that is inferior to those produced by other variants of GANs. 
Hence, several methods have been proposed recently to improve the MMD-GAN including optimized kernels and feature extractors \citep{sutherland2017generative,li2017mmd}, gradient regularization \citep{BinSutArbGre2018,Arbel:2018}, and repulsive loss \citep{wang2018improving}, to enumerate a few examples.



\section{Adversarial Training via Witness Points}
\label{sec:method}

Our idea is to construct a set of points $\Vset = \{\vbm_{1},\ldots \vbm_{J}\}\subset\mathcal{V}$, which we call \emph{witness points}, whose role is to capture \emph{local} differences between the real distribution $P_\Xb$ and the generated one $Q_\Yb$ and to provide means to guide training trajectories or injecting prior knowledge about the distribution during the course of training.

\subsection{Discrepancy Measure via Witness Points} 

We can rewrite the MMD~\eqref{eq:mmd} as $\left\|\mathbf{w}\right\|_{\mathcal{F}}^{2}$ where $\mathbf{w} \coloneqq  \mu_{P_\Xb}-\mu_{Q_\Yb} \in\mathcal{F}$ is known as a witness function which characterizes the differences between $P_{\Xb}$ and $Q_{\Yb}$ (see~\Cref{fig:global-local} in the appendix). For any $\xbm\in\mathcal{X}$, let $\mathbf{w}(\xbm)$ be a witness function evaluation at $\xbm$ which, according to the reproducing property of $\mathcal{F}$, can be computed by 
$\mathbf{w}(\xbm) = \langle \mu_{P_\Xb}-\mu_{Q_\Yb},\phi(\xbm)\rangle_{\mathcal{F}} = 
\mu_{P_\Xb}(\xbm)-\mu_{Q_\Yb}(\xbm)$.
By projecting the witness function $\mathbf{w}$ onto a set of $J$ directions $\{\phi(\vbm_{1}),\ldots,\phi(\vbm_{J})\}$ in the RKHS $\mathcal{F}$, \cite{ChwRamSejGre2015} and \cite{JitSzaChwGre2016} propose a discrepancy measure known as the \emph{unnormalized mean embeddings} (UME)
statisticas which is the Euclidean distance computed w.r.t. the projected values, i.e.,
\begin{align*}
\mathrm{UME}^{2} &=&
{\frac{1}{J}\left\|\left(\begin{array}{c}
 \mu_{P_\Xb}(\vbm_{1}) \\
\vdots\\
 \mu_{P_\Xb}(\vbm_{J})
\end{array}\right)-\left(\begin{array}{c}
 \mu_{Q_\Yb}(\vbm_{1}) \\
\vdots\\
 \mu_{Q_\Yb}(\vbm_{J})
\end{array}\right)\right\|_{2}^{2}} \nonumber \\
 &=&\frac{1}{J}\sum_{j=1}^{J}(\mu_{P_\Xb}(\vbm_{j})-\mu_{Q_\Yb}(\vbm_{j}))^{2}. \label{eq:ume_pop}
\end{align*}
It was shown in \cite{ChwRamSejGre2015} that if
$k$ is
characteristic, translation
invariant, real analytic, and $\{\vbm\}_{j=1}^{J}$ are drawn from any
distribution $\eta$ with a density, then for any $J\ge1$, $\eta$-almost
surely, $\mathrm{ME}^{2}(P_\Xb,Q_\Yb)=0$ iff $P_\Xb=Q_\Yb$. In \cite{JitSzaChwGre2016}, 
the authors extend the statistic by optimizing $\{\vbm_{j}\}_{j=1}^{J}$
so as to maximize the test power of the two-sample test
$H_{0}\colon P_\Xb=Q_\Yb$ against $H_{1}\colon P_\Xb\neq Q_\Yb$. The result is an
interpretable two-sample test which gives an evidence in the form
of optimized witness points showing where $P_\Xb$ and $Q_\Yb$ differ.

\paragraph{Objective function.}

Based on $J$ witness points in the input space $\mathcal{X}$, we propose to learn the generator $G_{\bm{\theta}}$ by solving $\min_{\bm{\theta}}\max_{\vjs}\; \mathcal{L}_{\lambda}(\bm{\theta},\vjs)$ where the loss function is defined as
\begin{equation}
    \label{eq:dual-generator-obj}
    \mathcal{L}_{\lambda}(\bm{\theta},\Vset) \coloneqq  \mathcal{D}(P_\Xb,Q_\Yb) + \lambda\cdot\mathrm{UME}^2(P_\Xb,Q_\Yb,\Vset).
\end{equation}
Here, $\mathcal{D}(P_\Xb,Q_\Yb)$ is an arbitrary GAN objective and $\lambda > 0$ is a trade-off parameter. 
The empirical estimate of~\eqref{eq:dual-generator-obj} when kernel $k(\cdot, \cdot)$ is employed can be obtained
using the observations $\xbm_{1},\ldots,\xbm_{n}$ from $P_\Xb$ and the generated samples
$\ybm_{1},\ldots,\ybm_{m}$ from the generator $Q_\Yb=G_{\bm{\theta}}$ as
\begin{align}
    & \widehat{\mathcal{L}}_{\lambda}(\bm{\theta},\Vset) = 
    \mathcal{D}(\hat{P}_\Xb,\hat{Q}_\Yb) \nonumber \\
    & + 
    \frac{\lambda}{J}\sum_{j=1}^{J} \left(\frac{1}{n}\sum_{i=1}^n k(\xbm_i,\vbm_j) -\right.
    \left. \frac{1}{m}\sum_{j=1}^m k(\ybm_j,\vbm_j)\right)^{2}
    \label{eq:objective_kernel}
\end{align}
where $\hat{P}_\Xb \coloneqq  \frac{1}{n}\sum_{i=1}^n\delta_{\xbm_i}$ and $\hat{Q}_\Yb \coloneqq  \frac{1}{m}\sum_{j=1}^m\delta_{\ybm_j}$ are the empirical distributions of $P_\Xb$ and $Q_\Yb$.

We can observe that
\begin{enumerate*}[label=(\roman*)]
    \item the first term on the rhs of~\eqref{eq:dual-generator-obj} acts as a \emph{global} discrepancy measure between $P_\Xb$ and $Q_\Yb$. For example, one may consider $\Dcal(P_\Xb,Q_\Yb) = \mathrm{MMD}^2(P_\Xb,Q_\Yb)$, i.e., the objective used in the MMD-GAN,
    \item the second term acts as an auxiliary objective that provides a \emph{local} discrepancy measure between $P_\Xb$ and $Q_\Yb$, allowing the generator to learn the fine details of the data,
    \item and by maximizing $\mathrm{UME}^2(P_\Xb,Q_\Yb,\vjs)$ at each iteration w.r.t. the witness points $\vjs$, they tell us where $P_\Xb$ and $Q_\Yb$ differ most.
\end{enumerate*}
For convenience and due to the reasons mentioned above, we call the proposed algorithm \ours which stands for \emph{GLObal/LOCal Adaptive Discrimination}. 

\begin{example}
\label{ex:mmd-ume}
    Consider a special case when we have $\mathcal{D}(P_\Xb,Q_\Yb) = \mathrm{MMD}^2(P_\Xb,Q_\Yb)$
    where $Q_\Yb=G_\thetabm(\Zb)$. 
    Then, our objective \eqref{eq:dual-generator-obj} can be expressed as
        $\mathcal{L}_{\lambda}(\bm{\theta},\vjs) = \|w_{\bm{\theta}}\|_{\mathcal{F}}^2 + \frac{\lambda}{J}\sum_{j=1}^J w_{\bm{\theta}}(\vbm_j)^2,$
    \noindent where $w_{\bm{\theta}} \coloneqq  \mu_{P_\Xb} - \mu_{Q_\Yb}$ is the
    \emph{witness function}. One can see that the objective function allows us
    to capture both local and global difference between $P_\Xb$ and $Q_\Yb$
    simultaneously. 
    
    
\end{example}

\begin{algorithm}[t] 
    \caption{\ours}
    \label{alg:ours}
    \SetKwInOut{Input}{input}\SetKwInOut{Output}{output}
    \Input{ The number of witness points $J$, tradeoff parameter $\lambda$, the learning rate $\gamma$ , the batch size $B$, the number of iterations to optimize the generator and the witness points in each loop $n_g, n_v$.}

     Initialize generator and discriminator parameters $\{\thetagen, \thetad\}$, initialize witness points $\Vset_0=\{\vbm_{j0}\}_{j=1}^J$, define the convergence criterion.
    
    \Output{the generator function $G_{\thetagen}(\zbm)$ and witness points $\Vset=\vjs$.}
    
     \While{convergence criterion is not met}{
        \For{$t=1,\ldots,n_v$}{
        Sample minibatches $\Xset=\{\xbm_i\}_{i=1}^B \sim P_\Xb$ and $\Zset=\{\zbm_j\}_{j=1}^B \sim P_\Zb$\\
            \For{$j=1,\ldots,J$}{
                $\vj \leftarrow \vj + \gamma \cdot \nabla_\vj \Lcal_\lambda\allparamsours$ \\ 
            }
        }
        \For{$t=1,\ldots,n_g$}{
            Sample minibatches $\Xset$ and $\set{Z}$\\
            $\thetagen \leftarrow \thetagen - \gamma \cdot  \nabla_\thetagen \Lcal_\lambda\allparamsours$\\ 
        }
     }
\end{algorithm}

\Cref{alg:ours} summarized the proposed \ours, which interlaces training iterations over the local term $\mathrm{UME}^2(P_\Xb,Q_\Yb,\vjs)$ (w.r.t. the witness points) with training iterations of the total loss $\mathcal{L}_{\lambda}(\bm{\theta},\vjs)$ (w.r.t the parameters of the generator). 



\subsection{Witness Points in a Latent Space}
\label{subsec:latent_space}

The formulation in~\eqref{eq:objective_kernel} places witness points in the data space. 
This works well when the differences between the two distributions are localized in this space.
However, when the data space is complex (e.g., natural images), localizing the
differences in the original domain (e.g., pixel space) may not be effective,
and it becomes necessary to detect differences in another task-dependent 
transformed space (e.g., the feature space of images). 
To this end, we propose to augment ~\eqref{eq:objective_kernel} with extra components to improve its performance on high-dimensional data. 




To deal with more complex data manifolds $\Mcal=\supp(P_\Xb)\subset\Xcal$, we restrict the witness points $\vjs$ to lie in the latent space $\Scal$ of an autoencoder~\citep{hinton2006reducing} where $\text{dim}(\Scal) \ll \text{dim}(\mathcal{X})$. Formally, we denote an autoencoder parameterized by $\thetad=\{\thetade, \thetadd\}$ as $A_\thetad : \Xcal\to\Xcal$, consisting of an encoder $E_{\thetade}:\mathcal{X}\to\Scal$ and a decoder $D_{\thetadd}:\Scal\to\mathcal{X}$ such that $A_\thetad(\xbm)= D_\thetadd(E_\thetade(\xbm))$.
Fig.~\ref{fig:algorithm_schematics} illustrates the algorithm when the autoencoder is incorporated.


The new objective function can then be written as
\begin{equation}
\label{eq:total_loss}
\Lcal(\thetagen,\thetad,\Vset)=\mathcal{L}_{\lambda_1}(\thetagen,\Vset) + \lambda_2 \Lcal^{r}(\thetade, \thetadd),
\end{equation}
\noindent where the first term on the rhs is equivalent to the original loss in ~\eqref{eq:objective_kernel} except that the kernel now depends on the encoder $E_\thetade$, i.e., $k_{E_\thetade}(\xbm, \ybm)=k(E_\thetade(\xbm), E_\thetade(\ybm)).$
The second term on the rhs measures the \emph{reconstruction error} of the autoencoder $A_\thetad$ and can be estimated by $\widehat{\Lcal^{r}}(\thetade, \thetadd) \coloneqq  \frac{1}{B}\sum_{i=1}^B\lVert \xbm_i-D_\thetadd(E_\thetade(\xbm_i))\rVert_2^2 + \frac{1}{B}\sum_{i=1}^B\lVert G_\thetagen(\zbm_i)-D_\thetadd(E_\thetade(G_\thetagen(\zbm_i)))\rVert_2^2$ when applied to the observations $\{\xbm_1,\ldots,\xbm_B\}\cup\{\zbm_1,\ldots,\zbm_B\}$ in a minibatch of size $2B$.

The benefits of our new objective \eqref{eq:total_loss} are twofold. First, the witness points live in the latent space $\Scal$ of the autoencoder whose dimension is lower than that of the data space $\Xcal$, resulting in a more tractable optimization problem. Second, the reconstruction loss encourages the encoder to be injective. Consequently, the encoder maps points from the data space to the new input space of the kernel without a substantial loss of information and allows the trained kernel to remain characteristic~\citep{li2017mmd}. 

We call this version of the algorithm which is built on top of \ours, \autoours due to the incorporated autoencoder architecture. See~\Cref{alg:autoours} (appendix) for more details. Even though most of our experiments are on GLOCAD, the theoretical guarantees of the next section apply to \autoours as a more general case.

\subsection{Stability Analysis}
\label{sec:theory}

In this section, we employ the theory of dynamical systems to investigate the stability properties of our algorithm for~\eqref{eq:total_loss} as a more general form of~\eqref{eq:objective_kernel} when the kernel is also trainable via an autoencoder. Specifically, we establish the existence of a desirable equilibrium point for the \ours, and prove that it is locally stable under some specific conditions.

Note that the gradient descent update rule on~\eqref{eq:total_loss} results in
the following continuous dynamical system with the state space
$\Phi=(\thetagen,\thetad,\Vset)$:
%
\begin{align}
\label{eq:dynamical_system}
        \dot{\bm{\theta}}_g &\coloneqq\shortminus \nabla_{\thetagen}\mathcal{L}(\thetagen,\thetad,\Vset), \quad\\
        \dot{\bm{\theta}}_d &\coloneqq\nabla_{\thetad}\mathcal{L}(\thetagen,\thetad,\Vset), \quad\\
        \dot{\Vset}_\vbm &\coloneqq  \nabla_{\vbm}\mathcal{L}(\thetagen,\thetad,\Vset).
\end{align}
%
To investigate the stability, we consider the behaviour of the dynamical system \eqref{eq:dynamical_system} around its equilibrium state.
Formally, the \emph{equilibrium state} $\Phi^*$ is a desirable equilibrium if $P_\Xb=Q_\Yb=G_{\thetagopt}(\Zb)$, regardless of the values of $\thetad$ and $\Vset$, and also the decoder $D_\thetadd$ is an injective function. 

\begin{proposition}\label{prop:equilibrium}
If the generator is realizable, i.e., there exists a $\thetagopt\in\Theta_g$ such that $P_\Xb=Q_\Yb=G_{\thetagopt}(\Zb)$, the dynamical system of~\eqref{eq:dynamical_system} has a desirable equilibrium $\Phi^*=(\thetagopt,\thetad,\Vset)$ and this equilibrium is locally exponentially stable.
\end{proposition}

{\it Proof idea--- }To prove the existence, the objective function of each term is expanded and differentiated to obtain the dynamics of~\eqref{eq:dynamical_system}. 
It is then showed that the rhs terms vanish at the equilibrium when $P_\Xb = Q_\Yb$. 
To prove the local stability, we linearize the system around the desirable equilibrium point by computing the Jacobian of the dynamics. 
We then prove that the submatrix $\partial\dot{\bm{\theta}}_g/\partial\bm{\theta}_g$ of the Jacobian is \emph{negative definite}, which is sufficient to conclude the stability of the system, e.g., see \cite{nagarajan2017gradient} and \cite[Lemma A.2]{wang2018improving}. 
The detailed proof is given in~\Cref{sec_appendix:stability_analysis}. 

Proposition \ref{prop:equilibrium} implies that introducing trainable witness points in $\Vset$---in addition to the aforementioned benefits---do not harm our algorithm in terms of its stability.

\begin{figure*}[t!]
    \centering
        \begin{subfigure}[c]{0.24\textwidth}
            \centering
            \includegraphics[width=\linewidth]{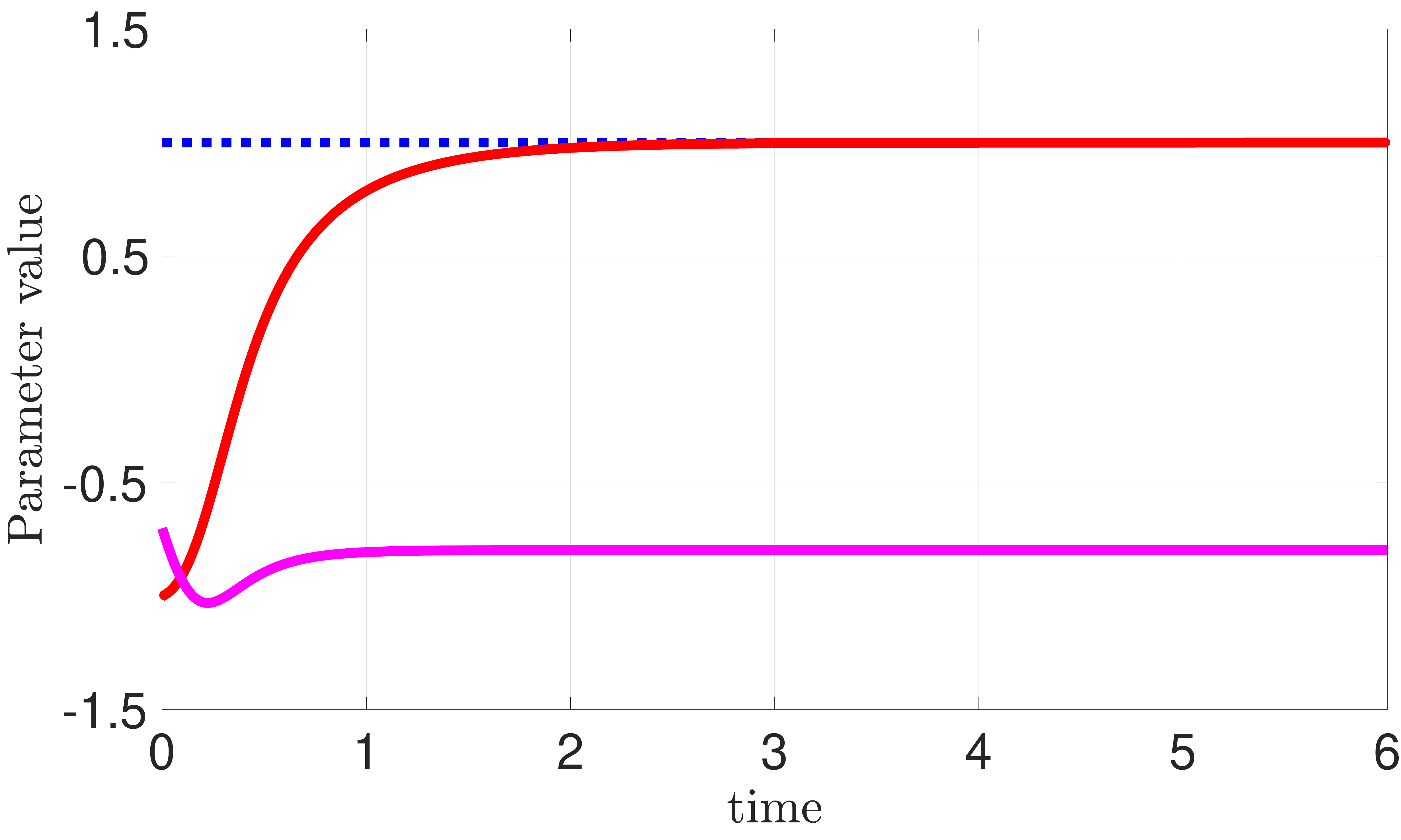}
            \caption{\scriptsize $\lambda=5$}
            \label{fig:single_gaussian_dynamics_lamda_5}
        \end{subfigure}
        \begin{subfigure}[c]{0.24\textwidth}
            \centering
            \includegraphics[width=\linewidth]{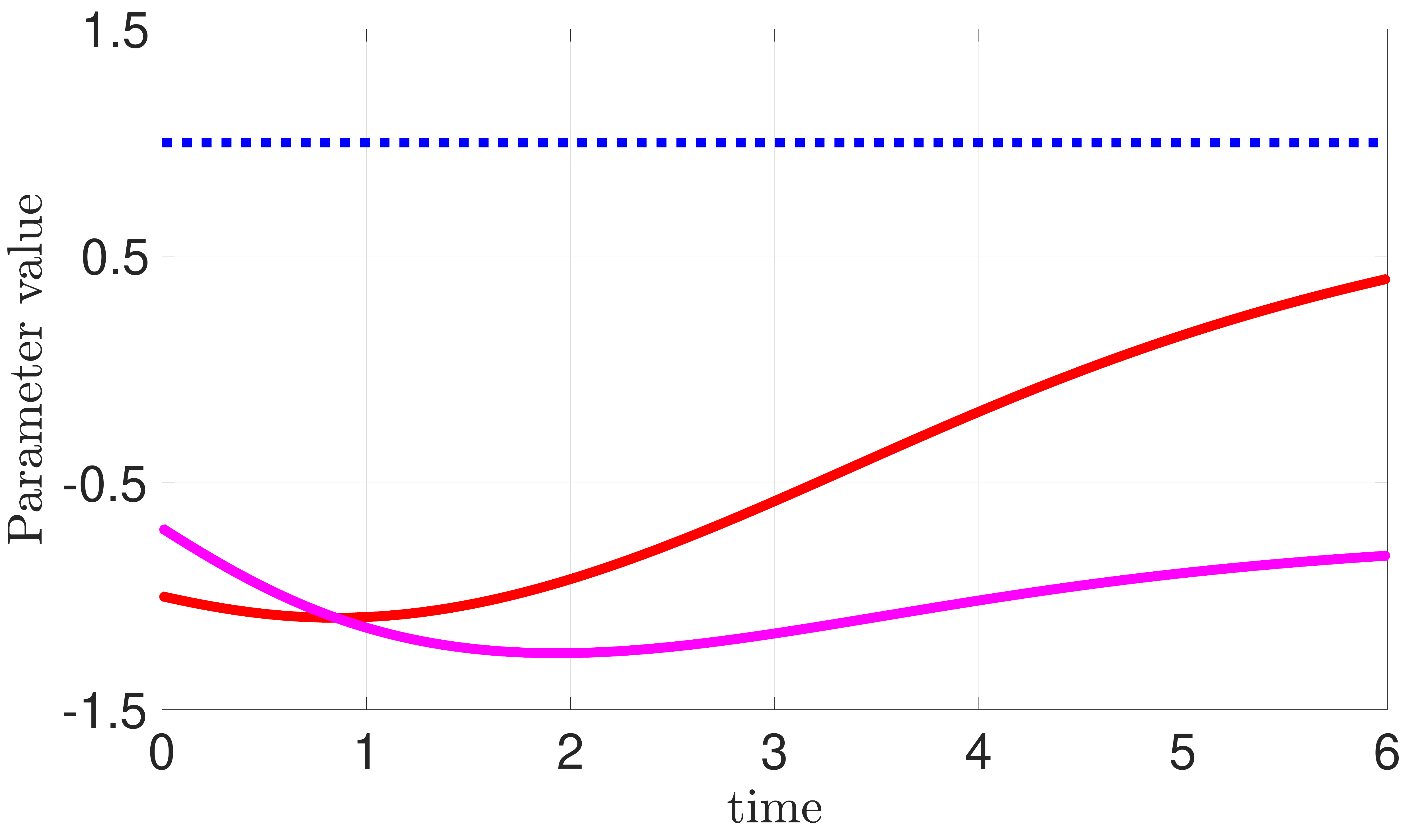}
            \caption{\scriptsize $\lambda=\infty$}
            \label{fig:single_gaussian_dynamics_lamda_inf}
        \end{subfigure}
        \begin{subfigure}[c]{0.25\textwidth}
            \centering
            \includegraphics[width=\linewidth]{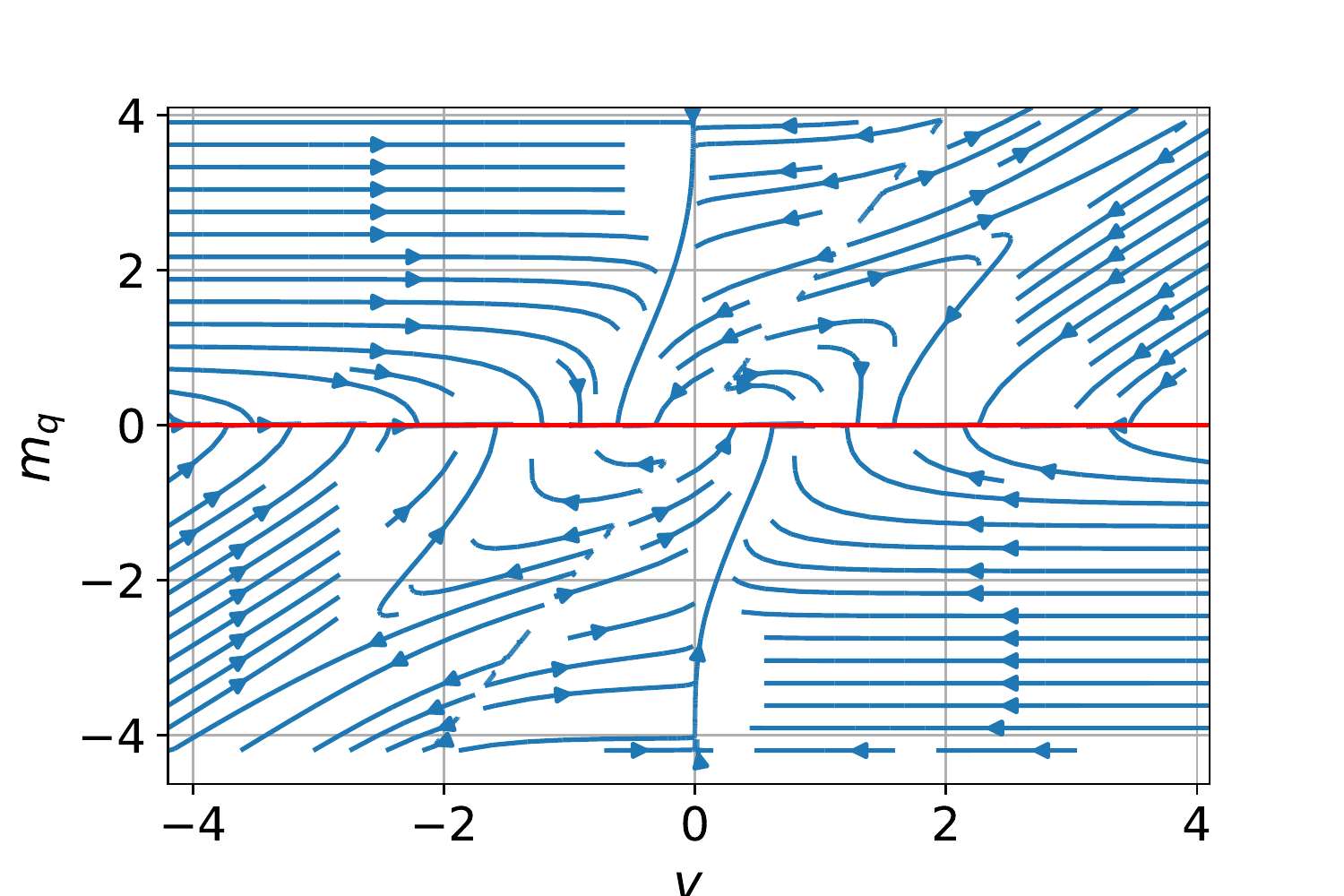}
            \caption{\scriptsize $\sigma$ =  $0.1$}
            \label{fig:phase_portrait_sigma_0.1}
        \end{subfigure}
        \begin{subfigure}[c]{0.25\textwidth}
            \centering
            \includegraphics[width=\linewidth]{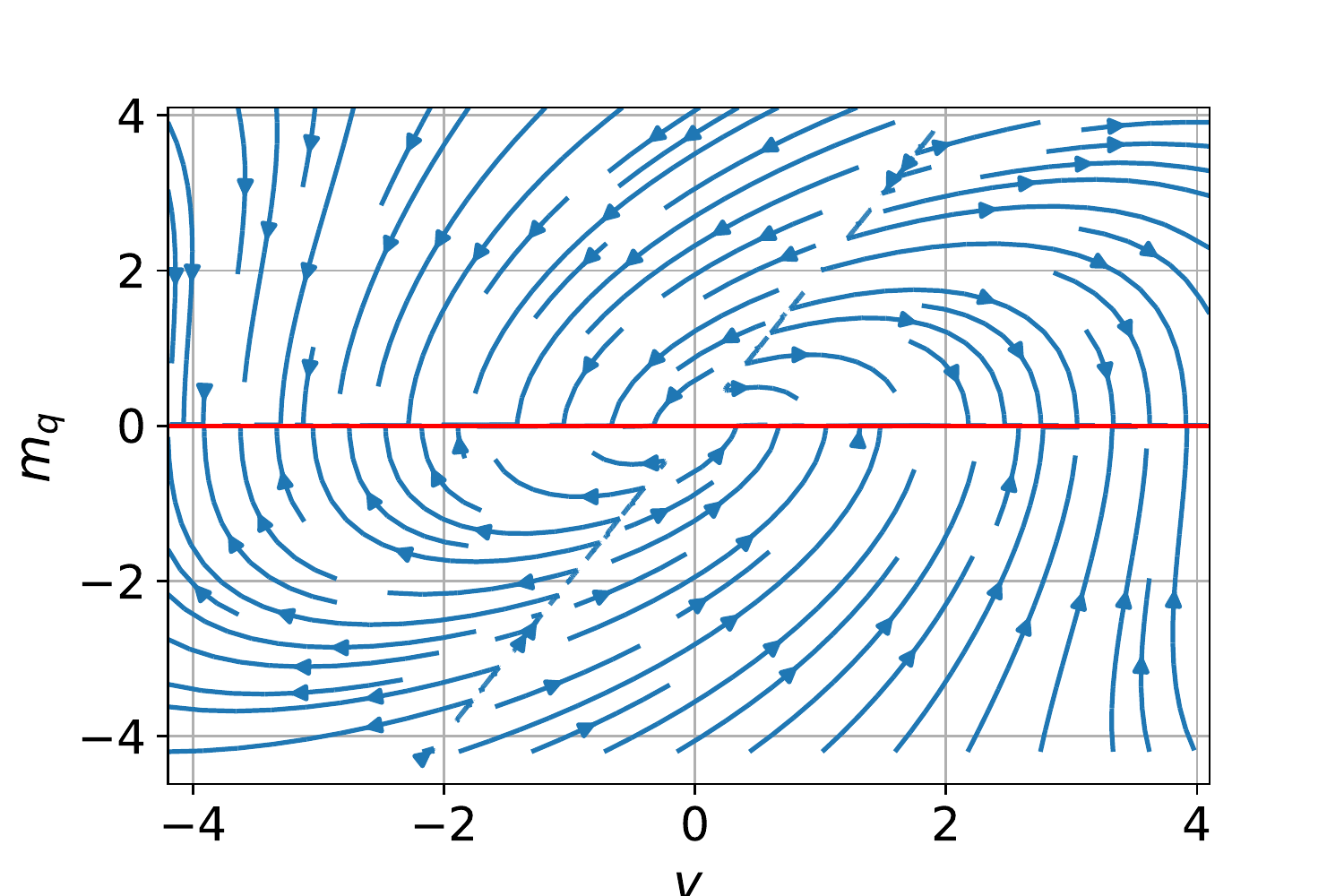}
            \caption{\scriptsize $\sigma$ =  $5$}
            \label{fig:phase_portrait_sigma_5}
        \end{subfigure}
        \caption{(a, b) Dynamics of training for the mean of a single Gaussian (c, d) Phase portrait of the dynamics of training for the mean of a single Gaussian. Color codes are: dotted blue $\rightarrow$ target, red $\rightarrow$ trainable model, magenta $\rightarrow$ witness point.}
        \vspace{-2mm}
\end{figure*}

\begin{figure*}[t!]
\centering
\begin{minipage}[t]{\textwidth}
    \centering
        \begin{subfigure}[c]{0.25\textwidth}
            \centering
            \includegraphics[width=\linewidth]{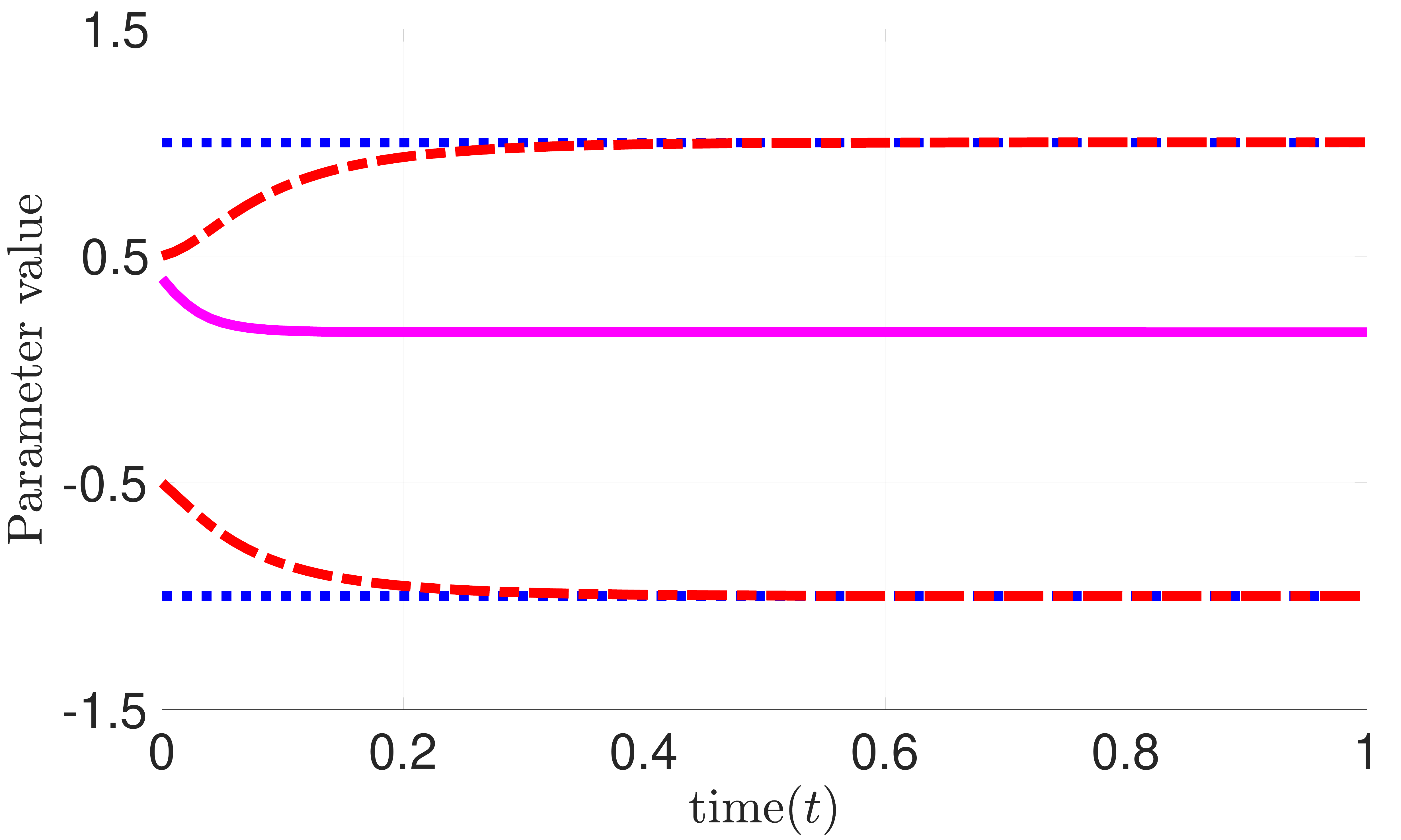}
            \caption{\scriptsize $\lambda=5$}
            \label{subfig:mog_a}
        \end{subfigure}
        \begin{subfigure}[c]{0.25\textwidth}
            \centering
            \includegraphics[width=\linewidth]{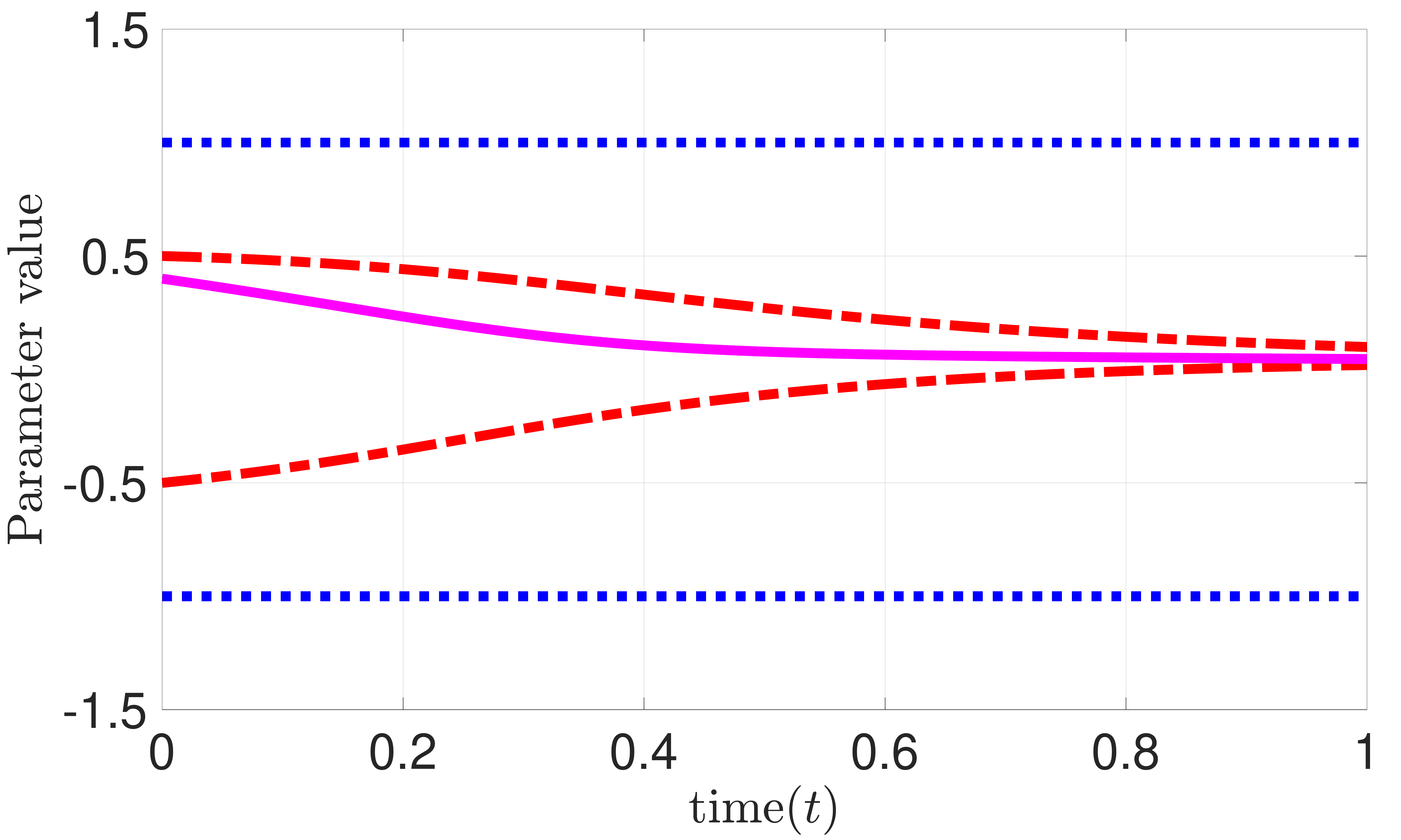}
            \caption{\scriptsize $\lambda=\infty$}
            \label{subfig:mog_b}
        \end{subfigure}
        \begin{subfigure}[c]{0.24\textwidth}
            \centering
            \includegraphics[width=\linewidth]{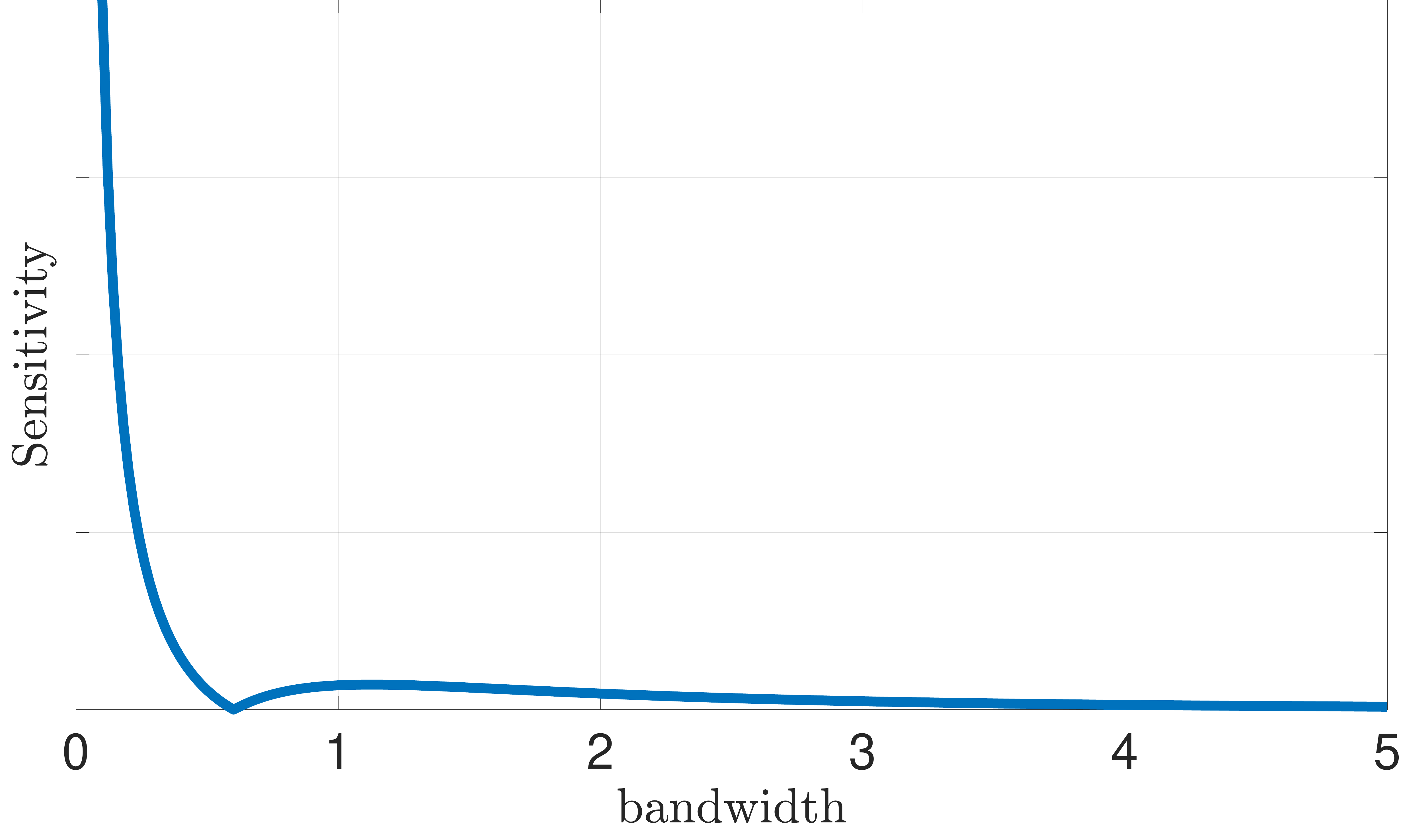}
            \caption{\scriptsize MMD}
            \label{subfig:sens_a}
        \end{subfigure}
        \begin{subfigure}[c]{0.24\textwidth}
            \centering
            \includegraphics[width=\linewidth]{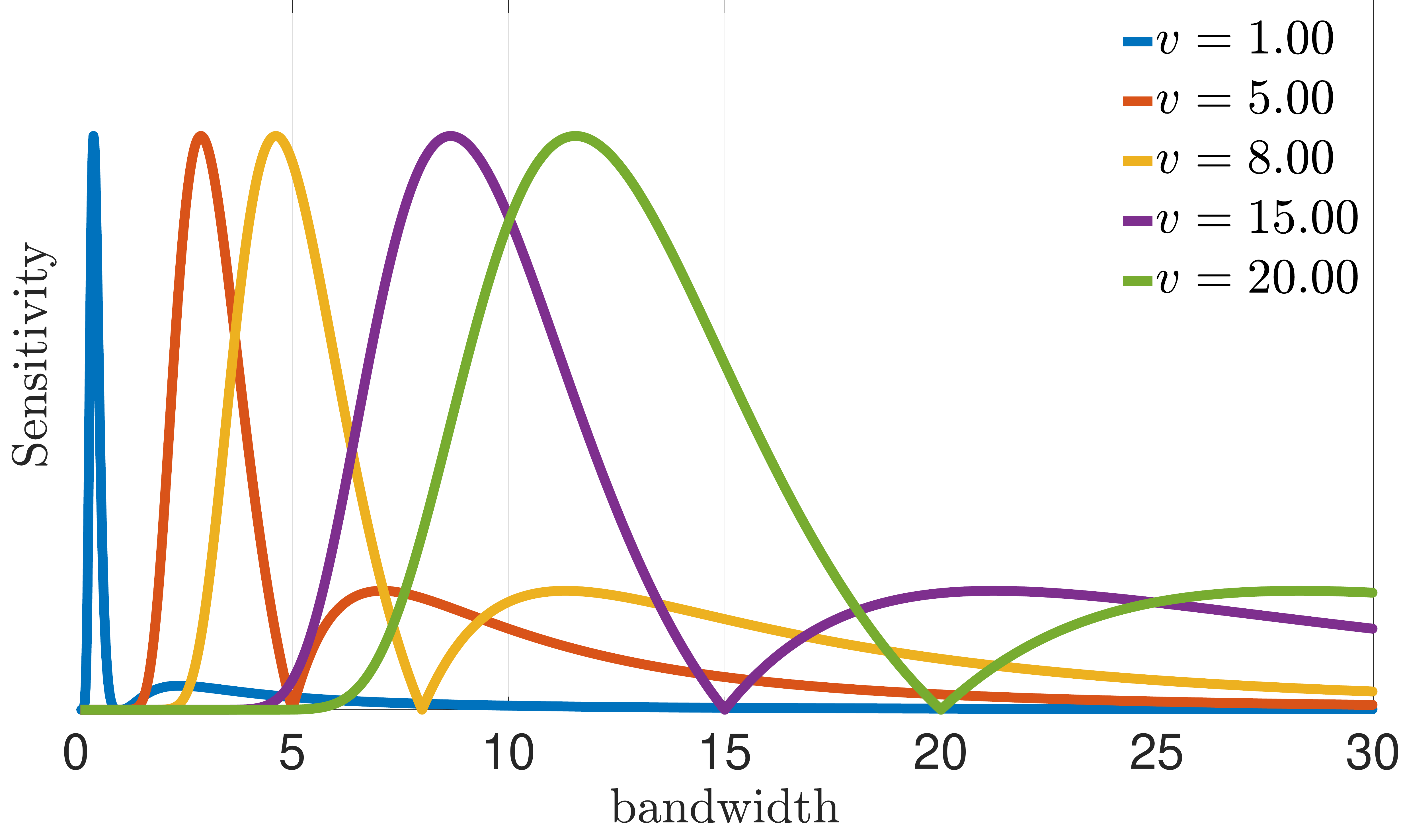}
            \caption{\scriptsize UME}
            \label{subfig:sens_b}
        \end{subfigure}
        \caption{(a, b) Numerical solution of the differential equations governing the GD training of two means of a mixture of Gaussians when only one witness point is provided. The results are plotted for various values of the regularization weight $\lambda$. $\lambda=\infty$ means that only the UME term exists. Color codes are the same as~\Cref{fig:single_gaussian_dynamics_lamda_5,fig:single_gaussian_dynamics_lamda_inf}. (c, d) While MMD is indifferent to the presence of different frequencies in the target distribution (See (c)), the UME can use its witness point to modulate its sensitivity to certain frequencies in the target distribution which are not yet captured by the model (See (b)).\\}
        \label{fig:low_dims_sim_mog}
\end{minipage}%
\vspace{-5mm}
\end{figure*}

\section{Experiments}
\label{sec:experiments}

\subsection{Numerical Simulation}
\label{sec:experiments_simulation}

First, we analyze the gradient descent (GD) dynamics of \ours on a one-dimensional problem with the analytic solution. The GD is considered as an update rule with infinitesimal step size resulting in a continuous dynamical system.


\paragraph{Single Gaussian.}
In this simulation, the model $Q_{\theta_q}=\mathcal{N}(m_q, \sigma_q)$ and
target distributions $P_\Xb=\mathcal{N}(0, 1)$ are both Gaussian and one
witness point is available to the model. Training dynamics  are derived by
computing the time derivative of parameters as $d\theta_q / dt =
\shortminus\nabla_{\theta_q}(\mathcal{L})$ and $dv / dt
=\nabla_v(\mathcal{L})$.~\Cref{fig:single_gaussian_dynamics_lamda_5} shows how
training dynamics of parameters change by varying the value of the witness
point. The simulation gives us two insights: First, as it is also clear from
the theory when the model captures the target distribution, the dynamics of the
witness point stops evolving which means that the value of witness points is
meaningful only when there is some mismatch between the model and the target.
Second, the rate of convergence of the model parameters is affected by the
initial value of the witness points which is also expected due to their tangled
dynamics. We observed that either too large or too small values of the witness
point both result in slow convergence suggesting that a clever choice of its
initial value could lead to faster convergence of the model to data
(See~\Cref{sec:appendix_single_gaussian}).

To gain a more holistic view of the dynamics, the \emph{phase portrait} of this dynamical system in a region around its equilibrium ($m_q=0$) is plotted for different values of the kernel bandwidth. It is seen in~\Cref{fig:phase_portrait_sigma_0.1,fig:phase_portrait_sigma_5} that the system posses a continuum of equilibria (depicted by the red line) rather than an isolated one. This is proved for the general case in Section~\ref{sec:theory} and Appendix~\ref{sec_appendix:local_stability_proof}. Moreover, notice that the value of the kernel bandwidth distorts the trajectories but does not lead to a qualitative change in the dynamical landscape of the parameters. 

\begin{figure*}[t!]
\begin{minipage}[t]{\textwidth}
    \centering
        \begin{subfigure}[c]{0.24\textwidth}
            \centering
            \includegraphics[width=\linewidth]{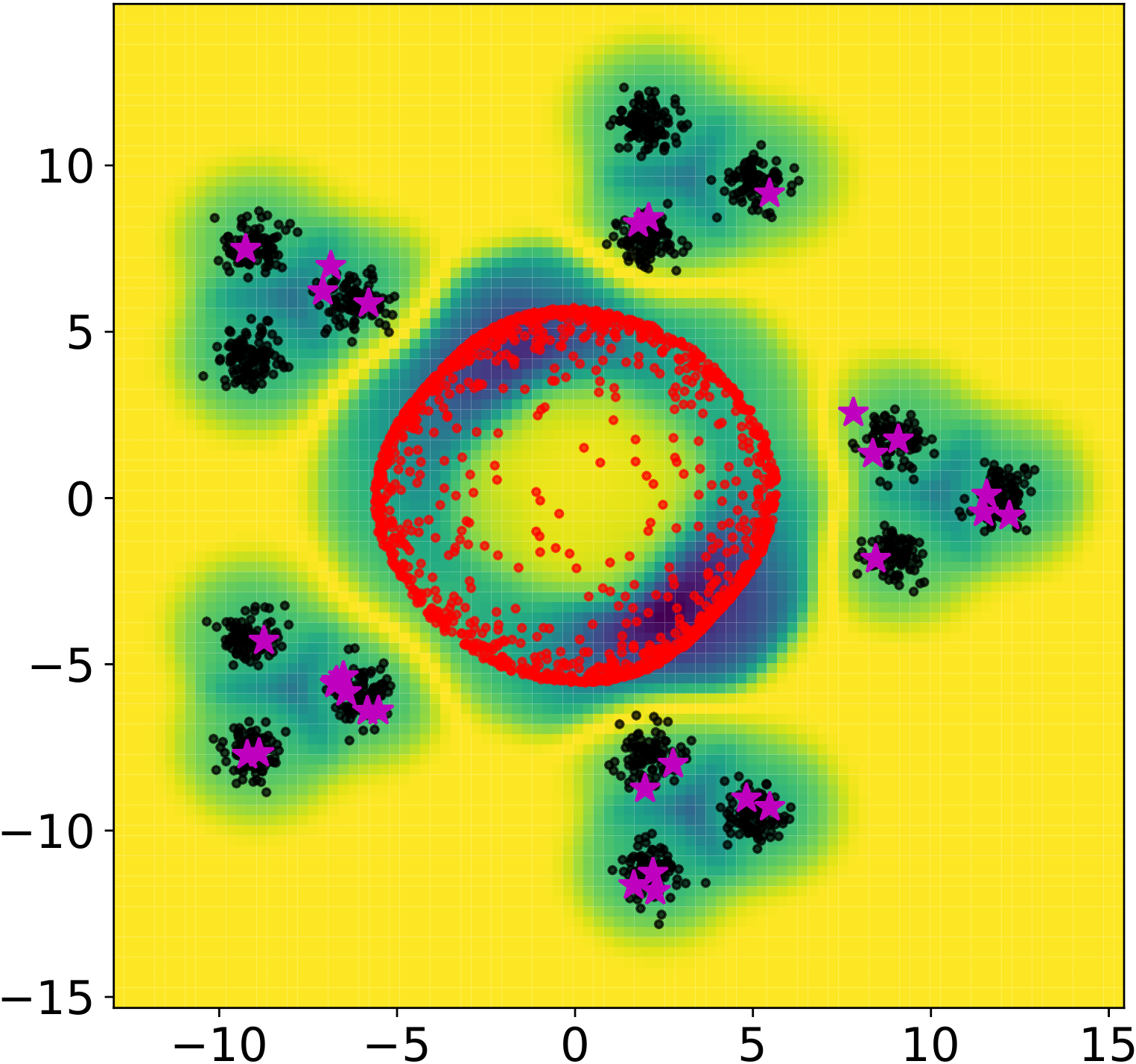}
            \caption{\scriptsize iter =  $10$}
            \label{subfig:momog_heatmap_a}
        \end{subfigure}
        \begin{subfigure}[c]{0.24\textwidth}
            \centering
            \includegraphics[width=\linewidth]{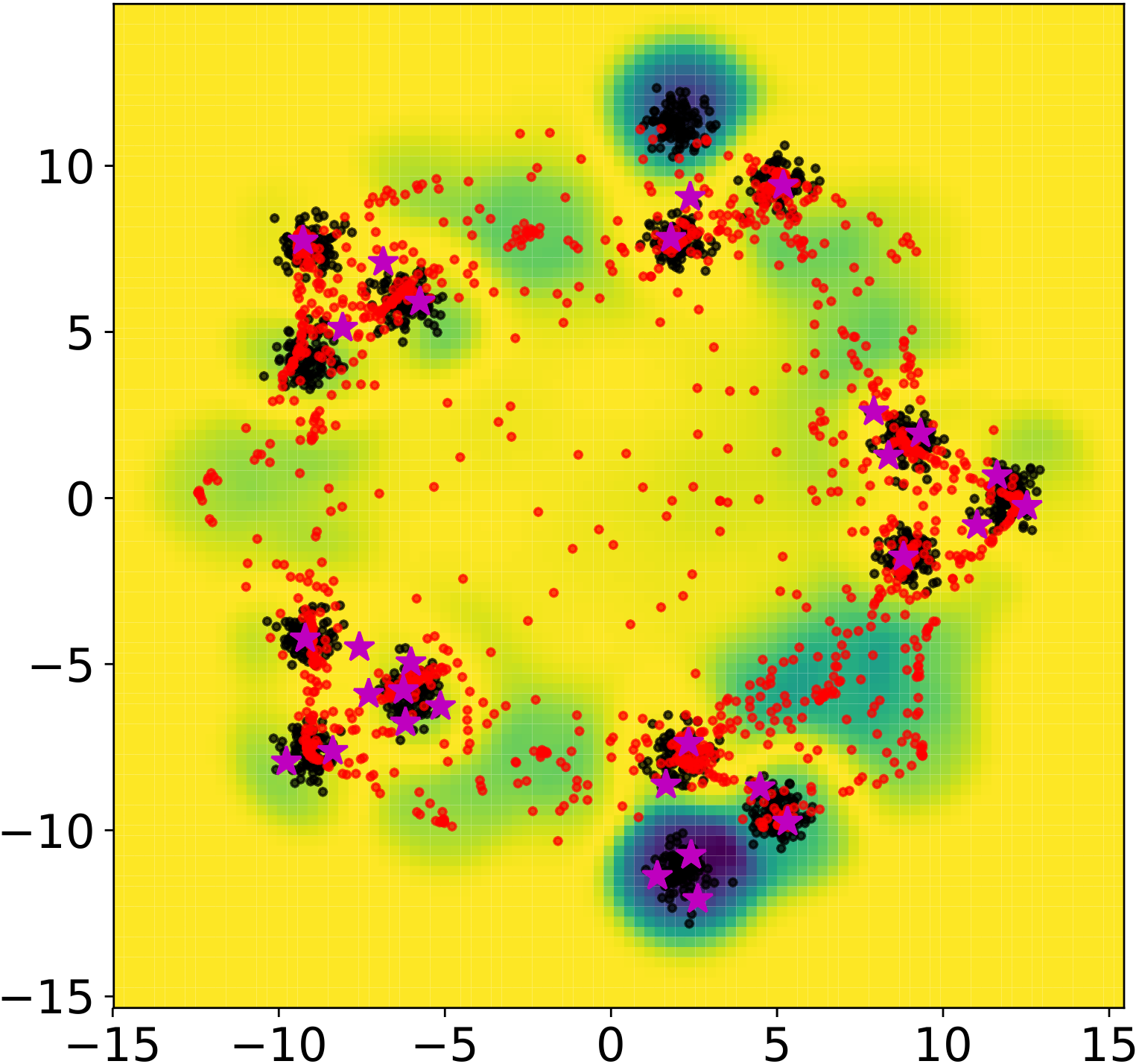}
            \caption{\scriptsize iter =  $200$}
            \label{subfig:momog_heatmap_b}
        \end{subfigure}
        \begin{subfigure}[c]{0.24\textwidth}
            \centering
            \includegraphics[width=\linewidth]{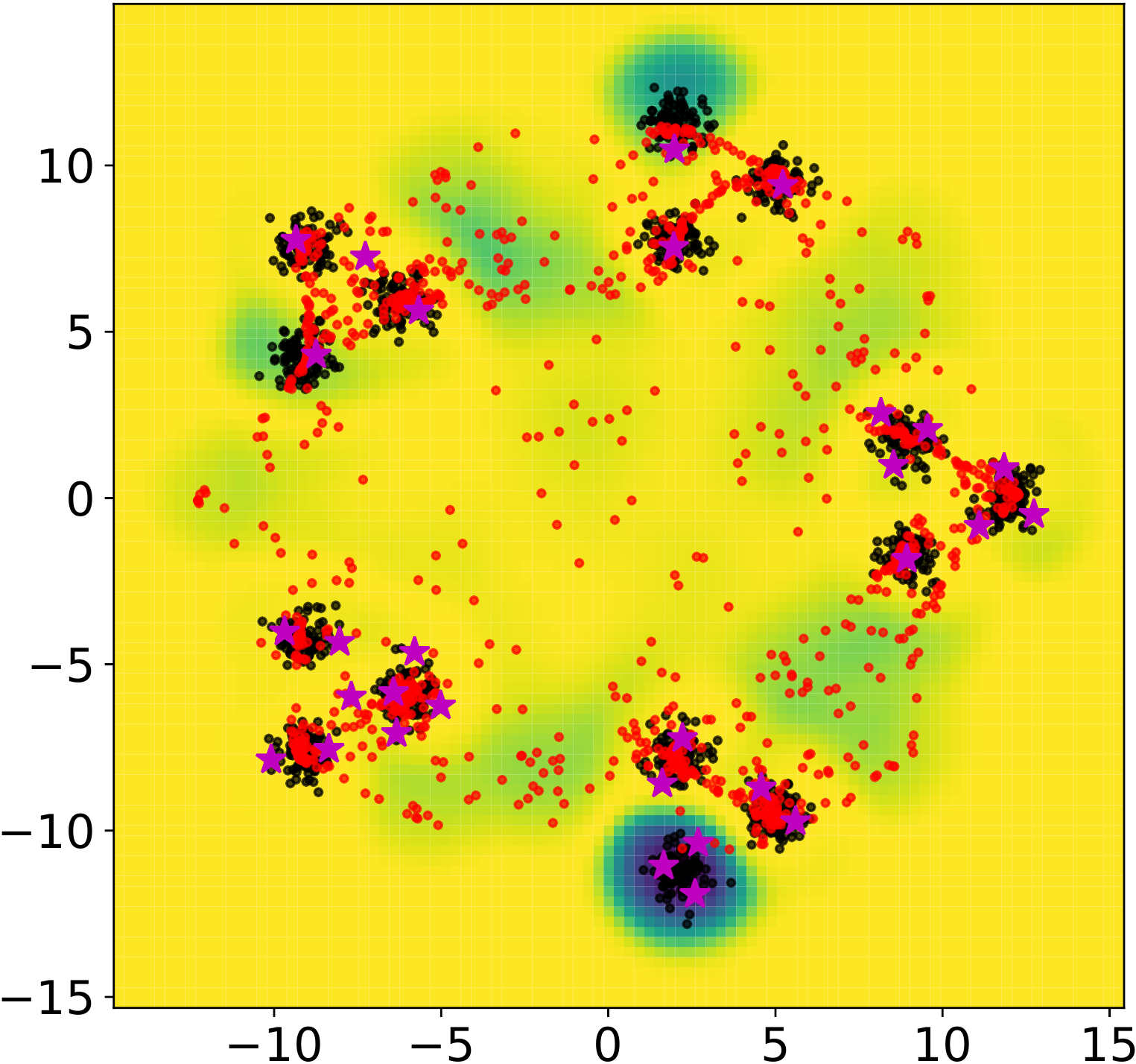}
            \caption{\scriptsize iter =  $400$}
            \label{subfig:momog_heatmap_c}
        \end{subfigure}
        \begin{subfigure}[c]{0.24\textwidth}
            \centering
            \includegraphics[width=\linewidth]{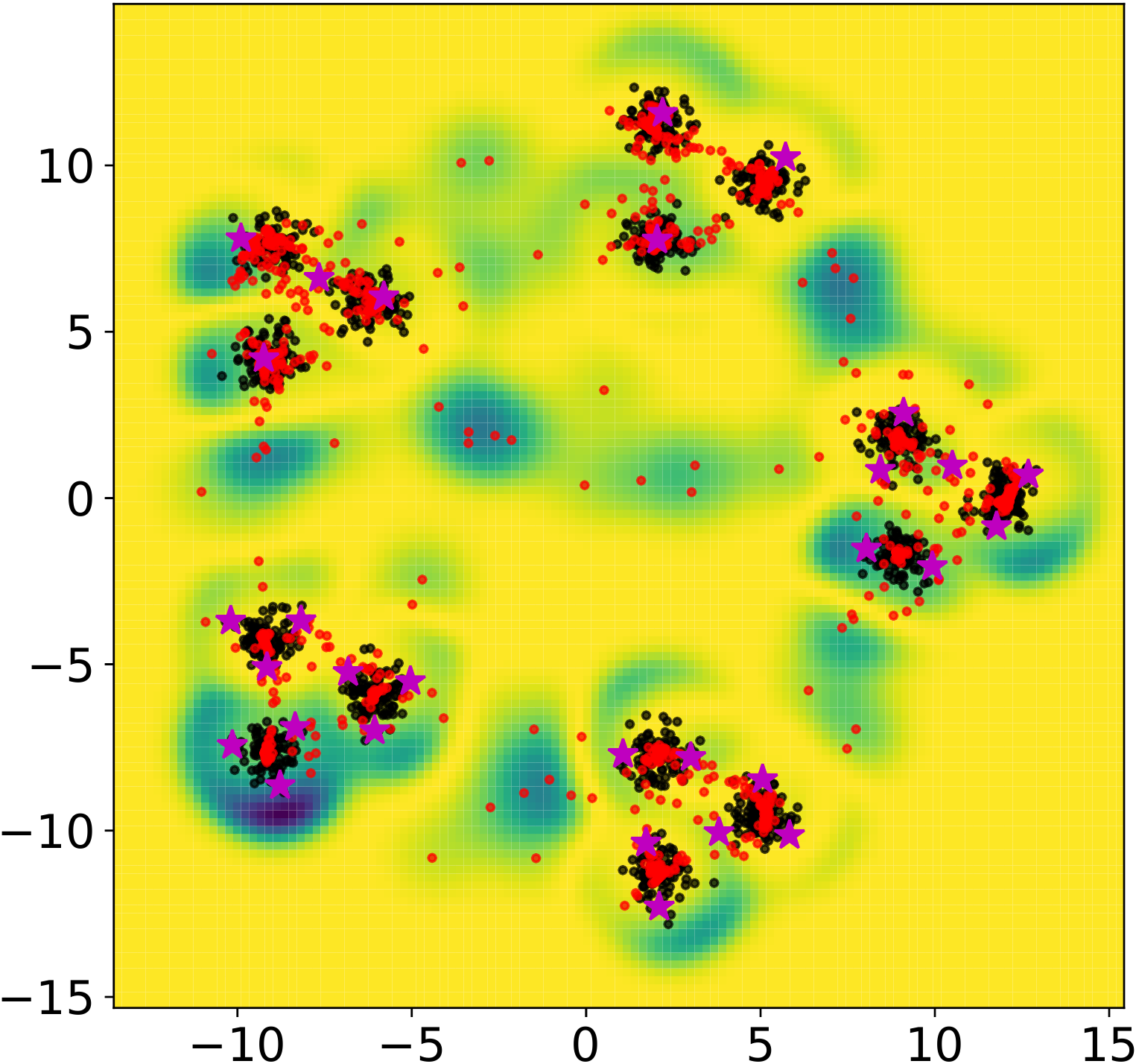}
            \caption{\scriptsize iter =  $1000$}
            \label{subfig:momog_heatmap_d}
        \end{subfigure}
        \caption{Two-dimensional mixture of Gaussian mixture models. Red dots: generated samples, black dots: target distribution, magenta stars: witness points, blue shade: the heatmap of the loss function of $\Vset$. The heatmap shows areas of the space that tend to absorb witness points during the optimization.}
         \label{fig:low_dims_2d_mog}
\end{minipage}%
\vspace{-5mm}
\end{figure*}

\begin{figure}[t!]
    \includegraphics[width=0.95\columnwidth]{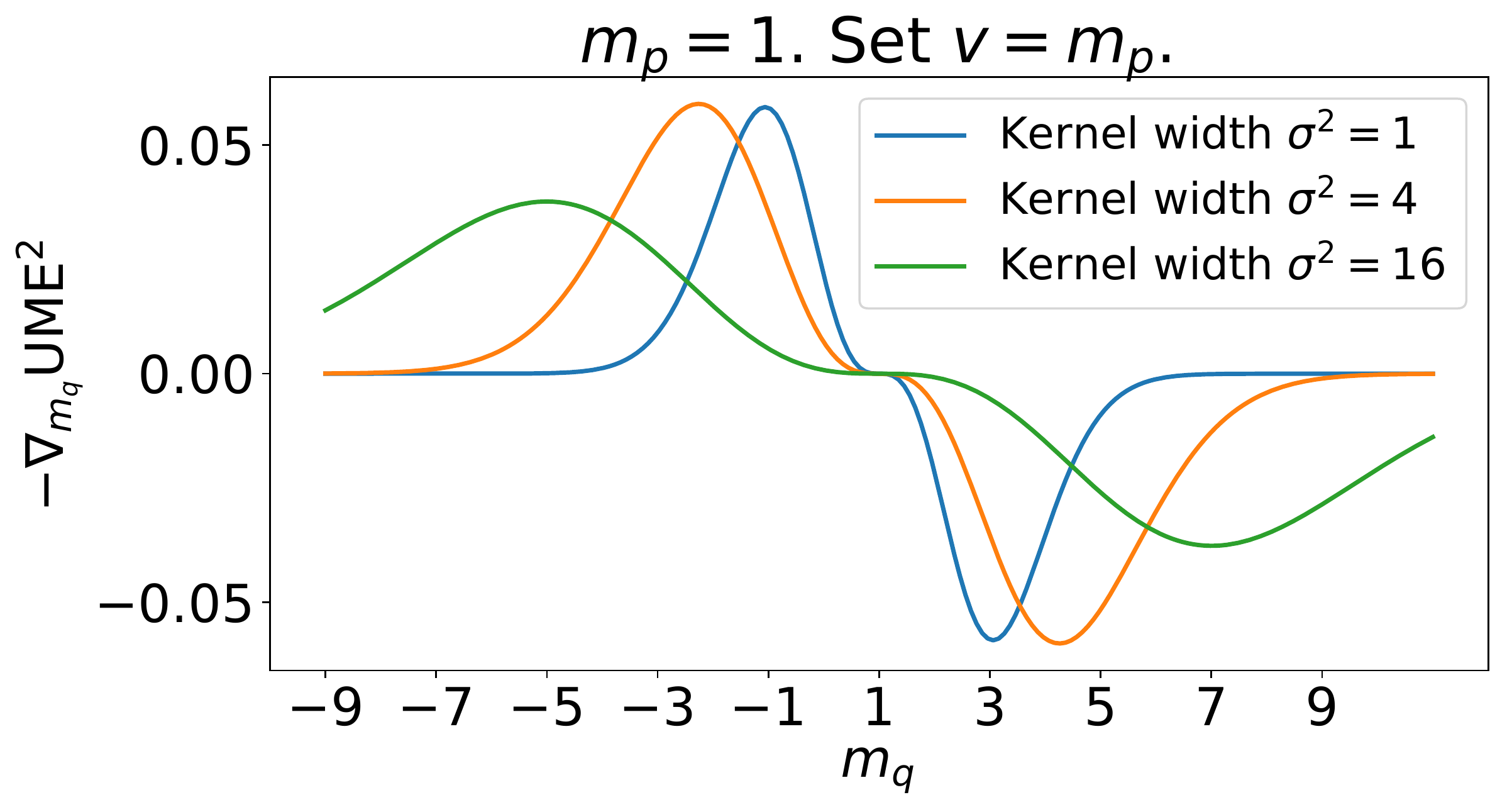}
    \caption{$-\nabla_{m_q}\mathrm{UME}^2$ evaluated at various values of $m_q$
Witness points in the UME guide the generator to learn to capture unknown data modes. 
        Gradient from the UME where the true parameter to learn is at $m_p=1$. 
    }
    \label{fig:mo2g_grad_ume}
\end{figure}

\paragraph{Mixture of Gaussians.}
Next, we investigate the dynamics of the parameters $\bm{\theta}_q = \{m_1,m_2\}$ of the model $Q_{\bm{\theta}_q} = 0.5\cdot\mathcal{N}(m_1,1) + 0.5\cdot\mathcal{N}(m_2,1)$ that aims to capture a target distribution when the model is given only a single witness point $v$. This experiment suggests that the number of witness points does not need to be equal to the modes of data. Since witness points are themselves trainable, their dynamics allows learning several modes by having a single witness point. The simulated dynamics of the parameters over the course of GD training is shown in~\Cref{subfig:mog_a,subfig:mog_b} . It was observed that the fastest convergence occurs for a middle-valued $\lambda$ that emphasizes the positive role of having UME along with the global term, which in this case is MMD (See~\Cref{fig:appendix_low_dims_sim_mog} in the appendix for further details).

\paragraph{Spiky Gaussian.} 
This experiment aims to show the ability of the \ours to detect local differences.
We use a mixture of two Gaussians as
the target distribution $P_\Xb$. Specifically, we consider
$P_1\coloneqq \mathcal{N}(0,1)$, $P_2\coloneqq \mathcal{N}(0,\sigma_{q}^{2})$ where variance $\sigma_{q}^{2} > 0$ and $P_\Xb\coloneqq wP_1+(1-w)P_2$ is defined on $\mathbb{R}$
for some weight $w\in[0,1]$.
We are interested in the case where $\sigma_q^2$ is small, and $w$ is close to 1. Consider a moment where the primary difference is local (at the origin), e.g., $Q=\mathcal{N}(0,1)$. Because the
second component ($P_2$) visually appears as a spike that rides on the wider Gaussian, we refer to $P_\Xb$ as a \emph{spiky} Gaussian. Now we aim to approximate this
distribution by a single Gaussian $Q$ as the model. 
%
In this setting, the population $\mathrm{MMD}$ and the population $\mathrm{UME}$ can be derived in closed form: 
\begin{align*}
\mathrm{MMD}^{2} &= (1-w)^{2}\left[c(2)+c(2\sigma_{q}^{2})-2c(1+\sigma_{q}^{2})\right], \\
\mathrm{UME}^{2}
 &
 =(1-w)^{2}\left[c(1)e^{\frac{-v^{2}}{2(\sigma^{2}+1)}}-c(\sigma_{q}^{2})
 e^{\frac{-v^{2}}{2(\sigma^{2}+\sigma_{q}^{2})}}\right]^{2},
\end{align*}
where $c(z)\coloneqq \sqrt{\frac{\sigma^{2}}{\sigma^{2}+z}}$ for $z>0$.
To show the usability of $\mathrm{MMD}$ and $\mathrm{UME}$ in detecting
the spiky component, we compare $\nabla_{\sigma_q}\mathrm{MMD}^2$ and $\nabla_{\sigma_q}\mathrm{UME}^2$ in~\Cref{subfig:sens_a} and~\Cref{subfig:sens_b}. As can be seen in~\Cref{subfig:sens_a}, MMD has no freedom to change its
sensitivity to different frequencies in the input. On the other hand,~\Cref{subfig:sens_b} shows that the sensitivity function of
UME peaks at different frequencies (bandwidths) by having the freedom to displace the witness points. This suggests
that UME enjoys this extra degree of freedom provided by witness points to focus
its attention on different frequencies in the input which is
analogous to what adaptive filters carry out~\citep{haykin2008adaptive}. 
Full derivation can be found in Section \ref{sec:spiky_gauss_details} in the appendix.


\paragraph{The Gradient $-\nabla \mathrm{UME}^2$ as a Training Guide.}
The added UME term to the main objective (see \eqref{eq:objective_kernel}) acts as a training
guide which directs the generator to capture an unknown data mode
as indicated by the witness points. The goal of this section is to
make this statement explicit with the following illustrative example.

Define $S=\mathcal{N}(0,1)$. Let the true data distribution be $P:=\omega P_{1}+(1-\omega)S$
where $P_{1}:=\mathcal{N}(m_{p},1)$ for some mean $m_{p}\neq0$ and
mixing proportion $\omega\in[0,1]$. Let the model be $Q:=\omega Q_{1}+(1-\omega)S$
where $Q_{1}:=\mathcal{N}(m_{q},1)$ and $m_{q}$ is the parameter
to learn. This problem illustrates a case where the data generating
distribution $P$ is bi-modal, and one of its two modes is already
captured by our model $Q$, i.e., the second component $S$. Assume
that one witness point $v$ is used, and it places at the mode of
$P_{1}$, i.e., $v=m_{p}$. Assume a Gaussian kernel with bandwidth
$\sigma^{2}$. The gradient $-\nabla_{m_{q}}\mathrm{UME}^{2}$ 
is
{\small
\begin{align*}
-\frac{2\sigma^{2}}{\sigma^{2}+1}\omega^{2}
\left[1-e^{\left(-\frac{(m_{p}-m_{q})^{2}}{2(\sigma^{2}+1)}\right)}
\right]e^{\left(-\frac{(m_{p}-m_{q})^{2}}{2(\sigma^{2}+1)}\right)}
\frac{m_{p}-m_{q}}{\sigma^{2}+1}.
\end{align*}
}%
and 
is shown in Figure \ref{fig:mo2g_grad_ume} (full derivations in \Cref{sec:ume_grad})
where we assume that the true parameter $m_{p}=1$ and $\omega=1/2$.

We observe that at $m_{q}=m_{p}$ (i.e., $P=Q$), the gradient is
zero, providing no signal to change $m_{q}$ further. When $m_{q}>m_{p}$,
the gradient is negative (pulling back), which will decrease the value
of $m_{q}$ toward $m_{p}$. Likewise, when $m_{q}<m_{p}$, the gradient
is positive (pushing forward), leading to a positive change of $m_{q}$
toward $m_{p}$. Since $\mathrm{UME}$ measures local differences
around the witness point $v$, the gradient signal is strong in the
neighborhood of $v$, and much weaker in other regions that are far
from $v$ (theoretically non-zero). This observation holds true for
any kernel bandwidth $\sigma^{2}>0$. It can be seen that a higher
value of the kernel bandwidth makes UME less local (i.e., larger coverage
with dispersed gradient signal) at the expense of lower magnitudes
(pointwise).

\begin{figure*}[t]
\begin{minipage}[t]{\textwidth}
    \newcommand{\eachw}{0.18\textwidth}
    \centering
        \begin{subfigure}[c]{\eachw}
            \centering
            \includegraphics[width=\linewidth]{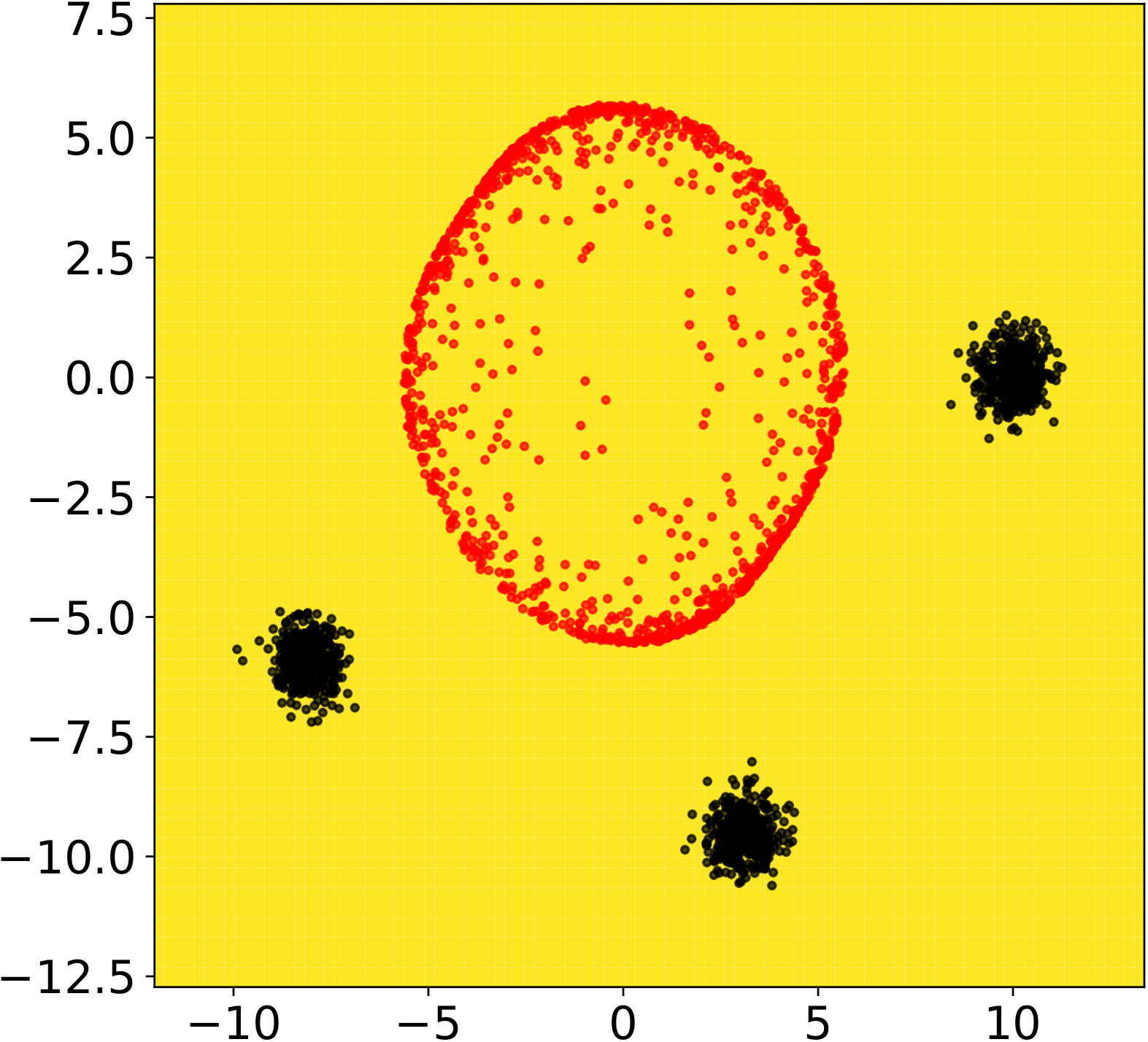}
            \caption{\scriptsize Phase $1$, iter =  $10$}
            \label{subfig:continual_learning_mmd_a}
        \end{subfigure}
        \begin{subfigure}[c]{\eachw}
            \centering
            \includegraphics[width=\linewidth]{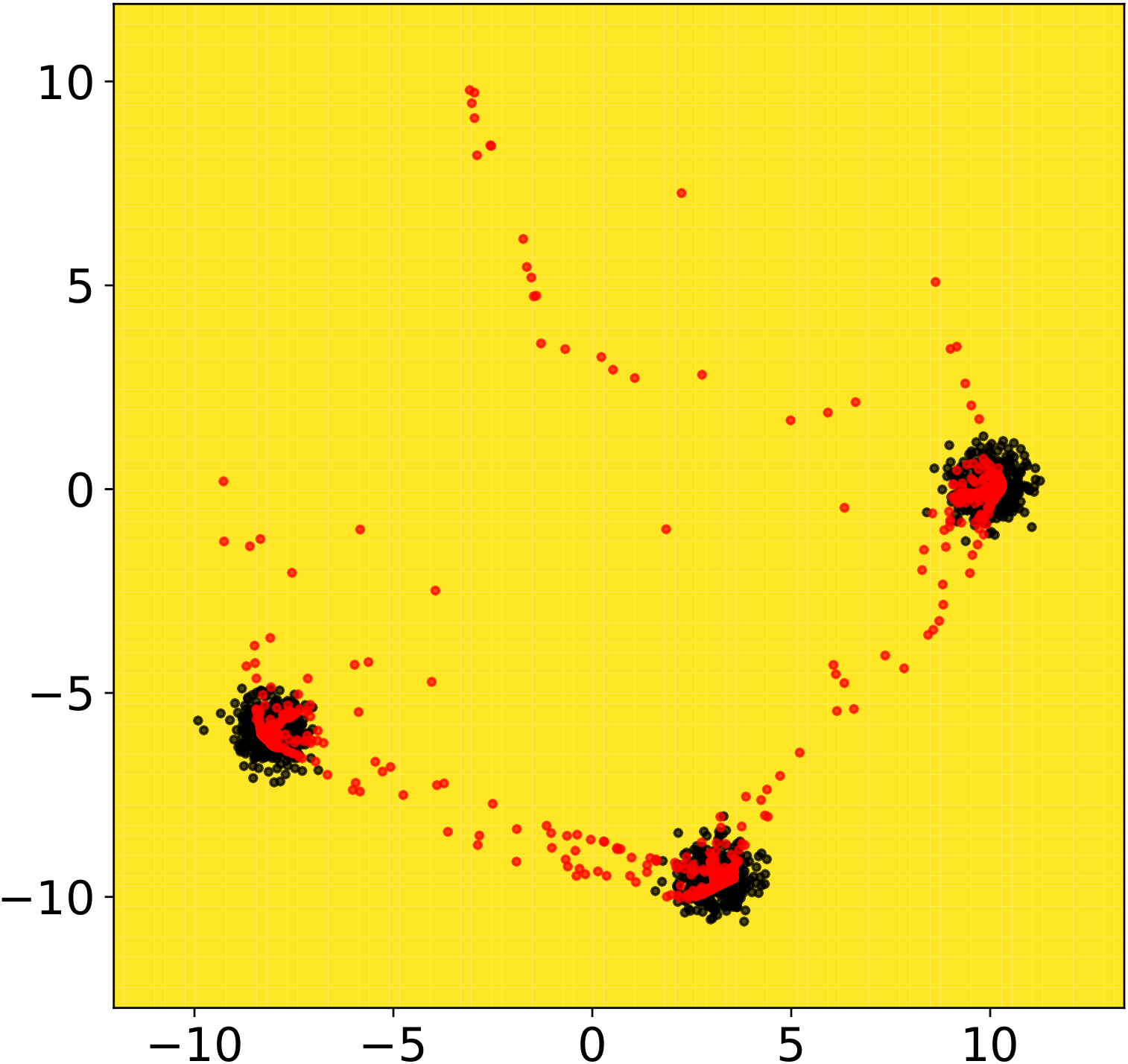}
            \caption{\scriptsize Phase $1$, iter =  $50$}
            \label{subfig:continual_learning_mmd_b}
        \end{subfigure}
        \begin{subfigure}[c]{\eachw}
            \centering
            \includegraphics[width=\linewidth]{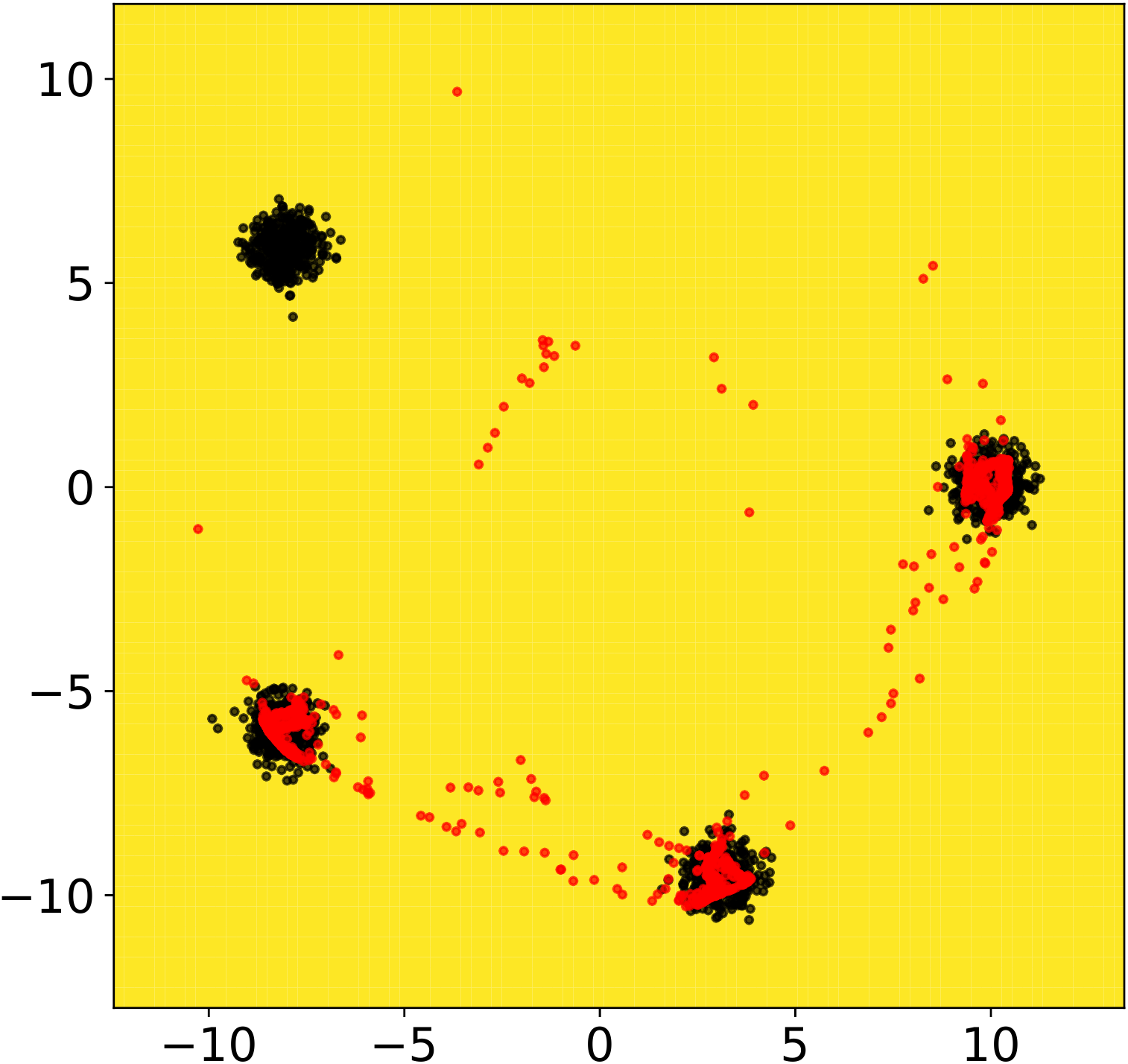}
            \caption{\scriptsize Phase $2$, iter =  $10$}
            \label{subfig:continual_learning_mmd_c}
        \end{subfigure}
        \begin{subfigure}[c]{\eachw}
            \centering
            \includegraphics[width=\linewidth]{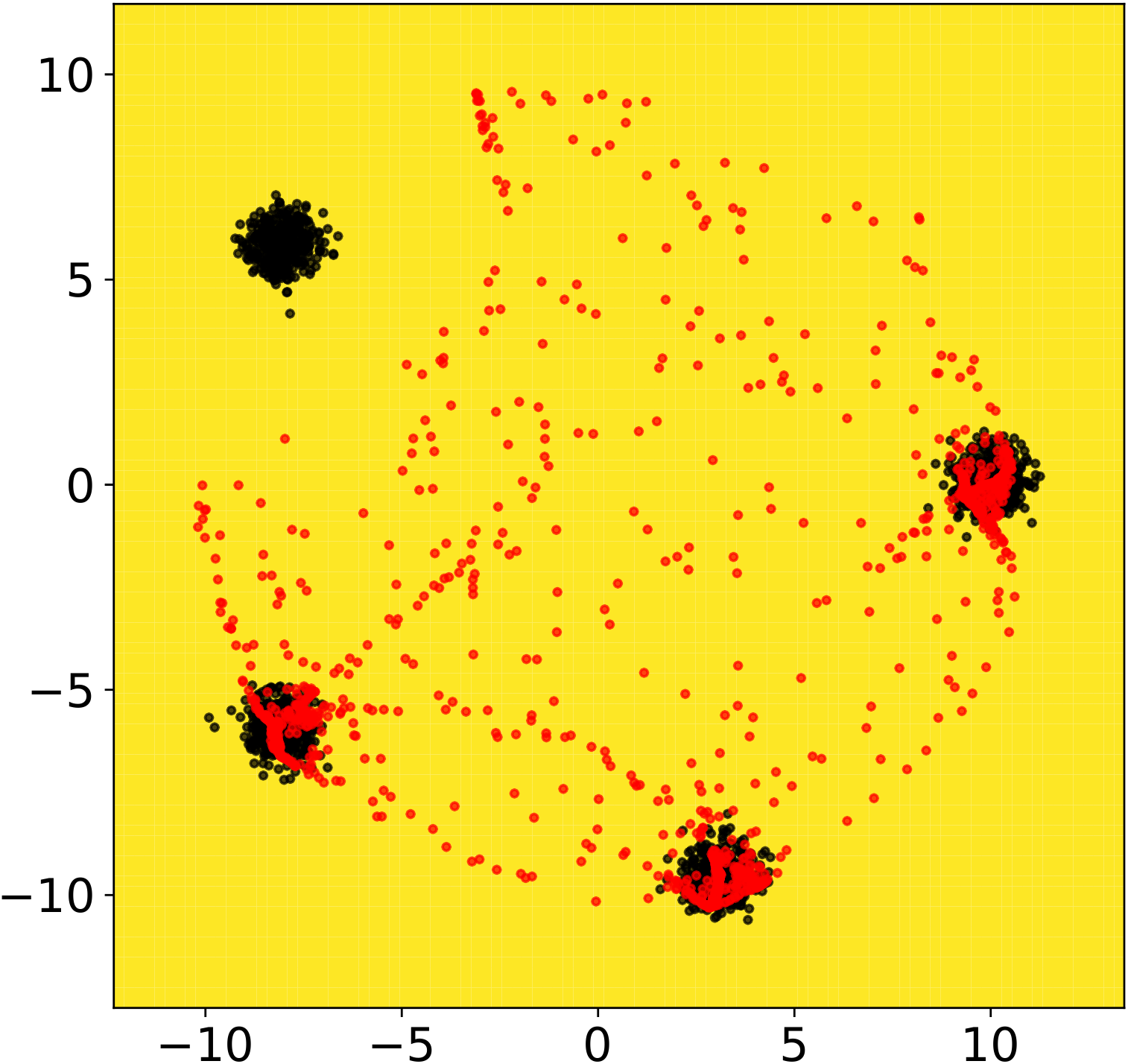}
            \caption{\scriptsize Phase $2$, iter =  $50$}
            \label{subfig:continual_learning_mmd_d}
        \end{subfigure}
        \begin{subfigure}[c]{\eachw}
            \centering
            \includegraphics[width=\linewidth]{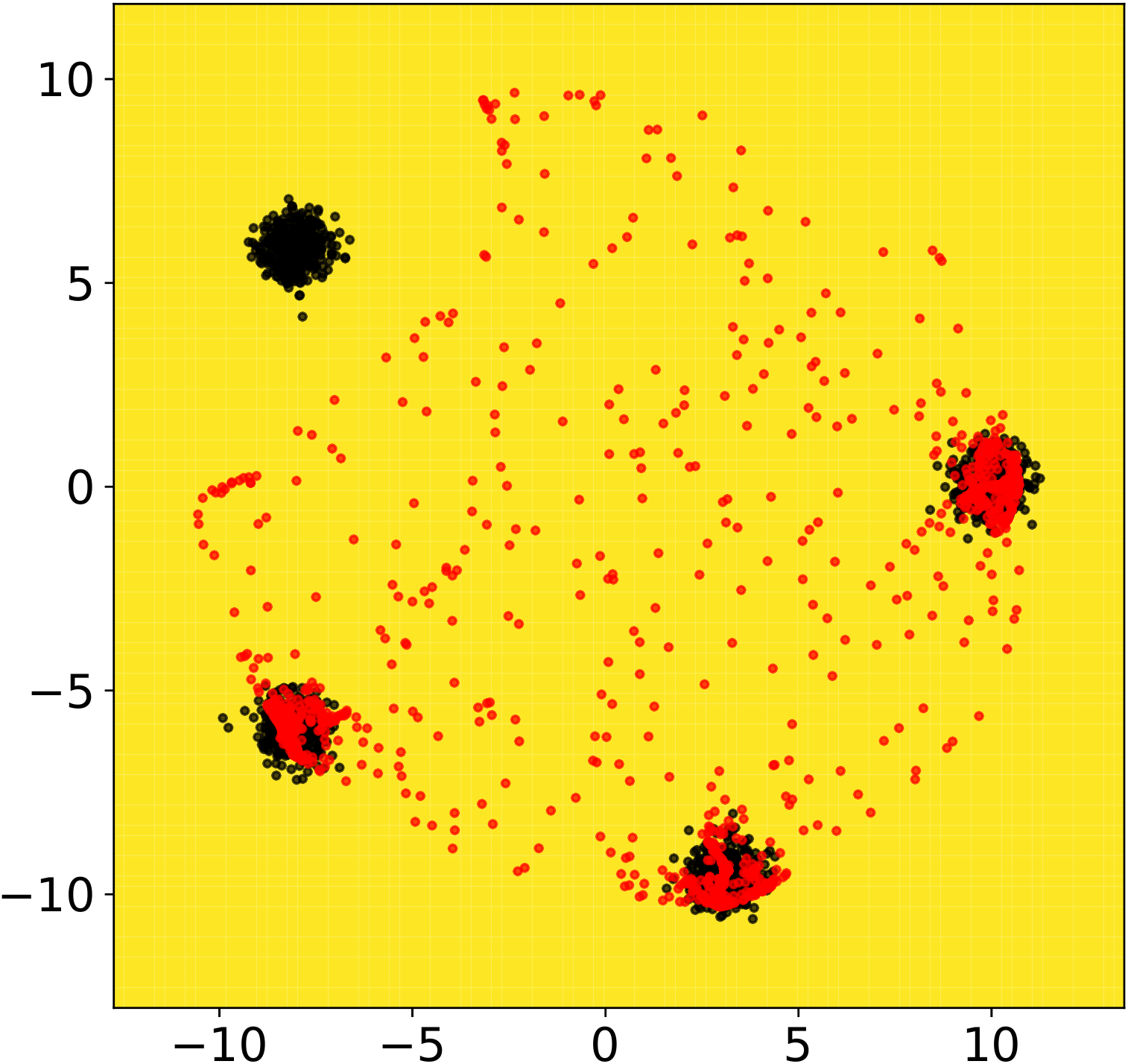}
            \caption{\scriptsize Phase $2$, iter =  $100$}
            \label{fsubfig:continual_learning_mmd_e}
        \end{subfigure}
        \vspace{2mm}
        
        \begin{subfigure}[c]{\eachw}
            \centering
            \includegraphics[width=\linewidth]{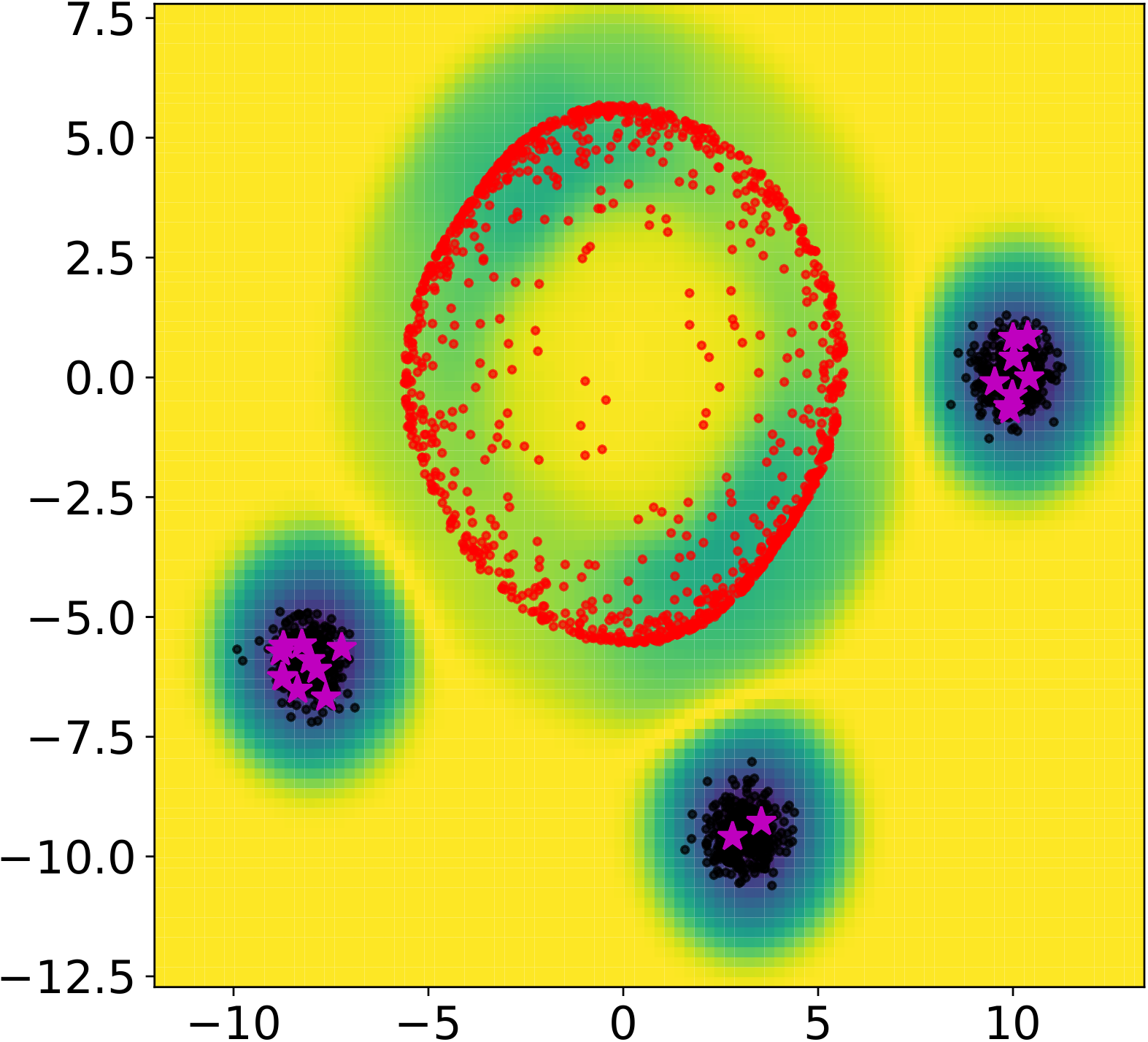}
            \caption{\scriptsize Phase $1$, iter =  $10$}
            \label{subfig:continual_learning_ume_a}
        \end{subfigure}
        \begin{subfigure}[c]{\eachw}
            \centering
            \includegraphics[width=\linewidth]{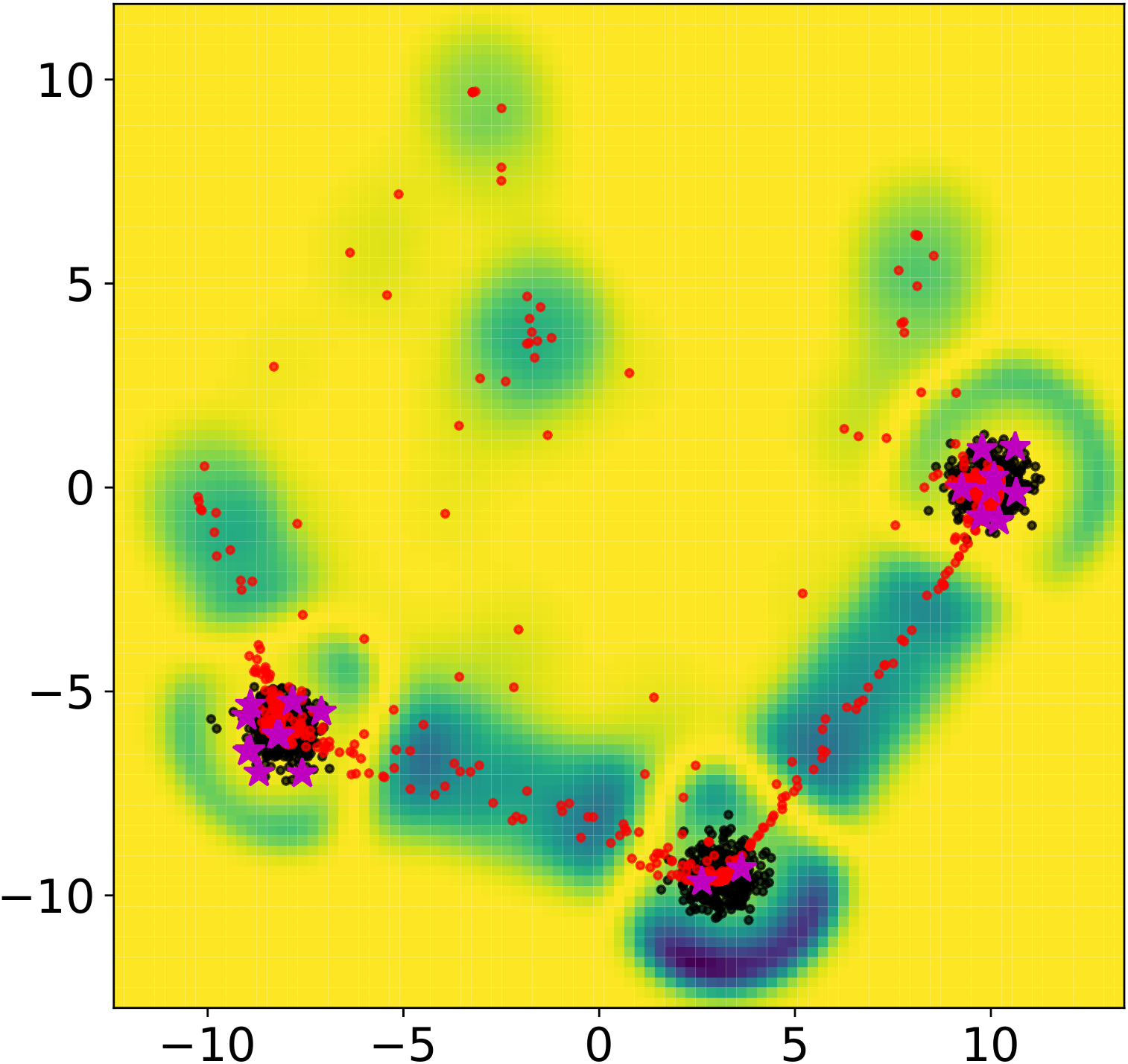}
            \caption{\scriptsize Phase $1$, iter =  $50$}
            \label{subfig:continual_learning_ume_b}
        \end{subfigure}
        \begin{subfigure}[c]{\eachw}
            \centering
            \includegraphics[width=\linewidth]{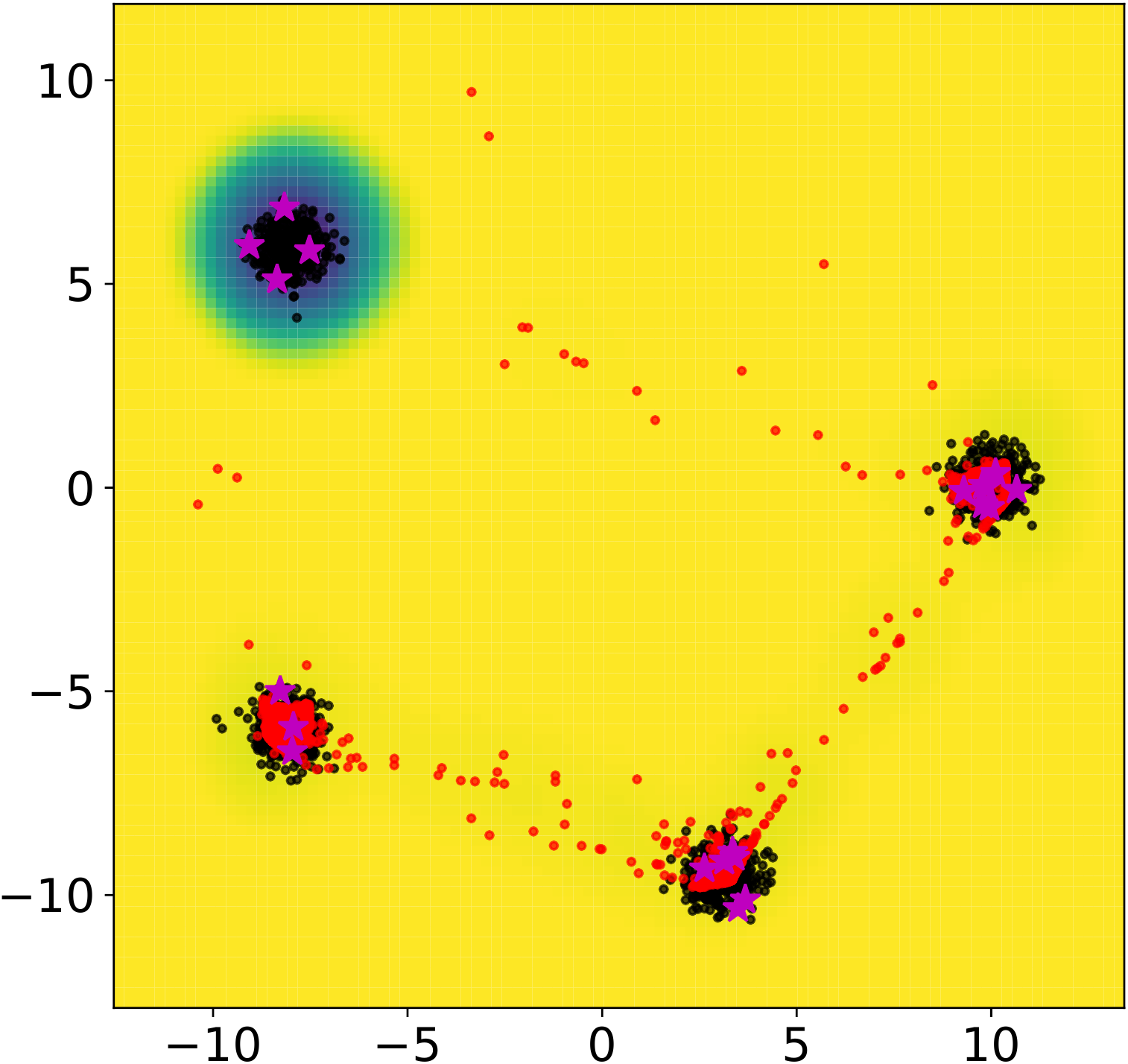}
            \caption{\scriptsize Phase $2$, iter =  $10$}
            \label{subfig:continual_learning_ume_c}
        \end{subfigure}        
        \begin{subfigure}[c]{\eachw}
            \centering
            \includegraphics[width=\linewidth]{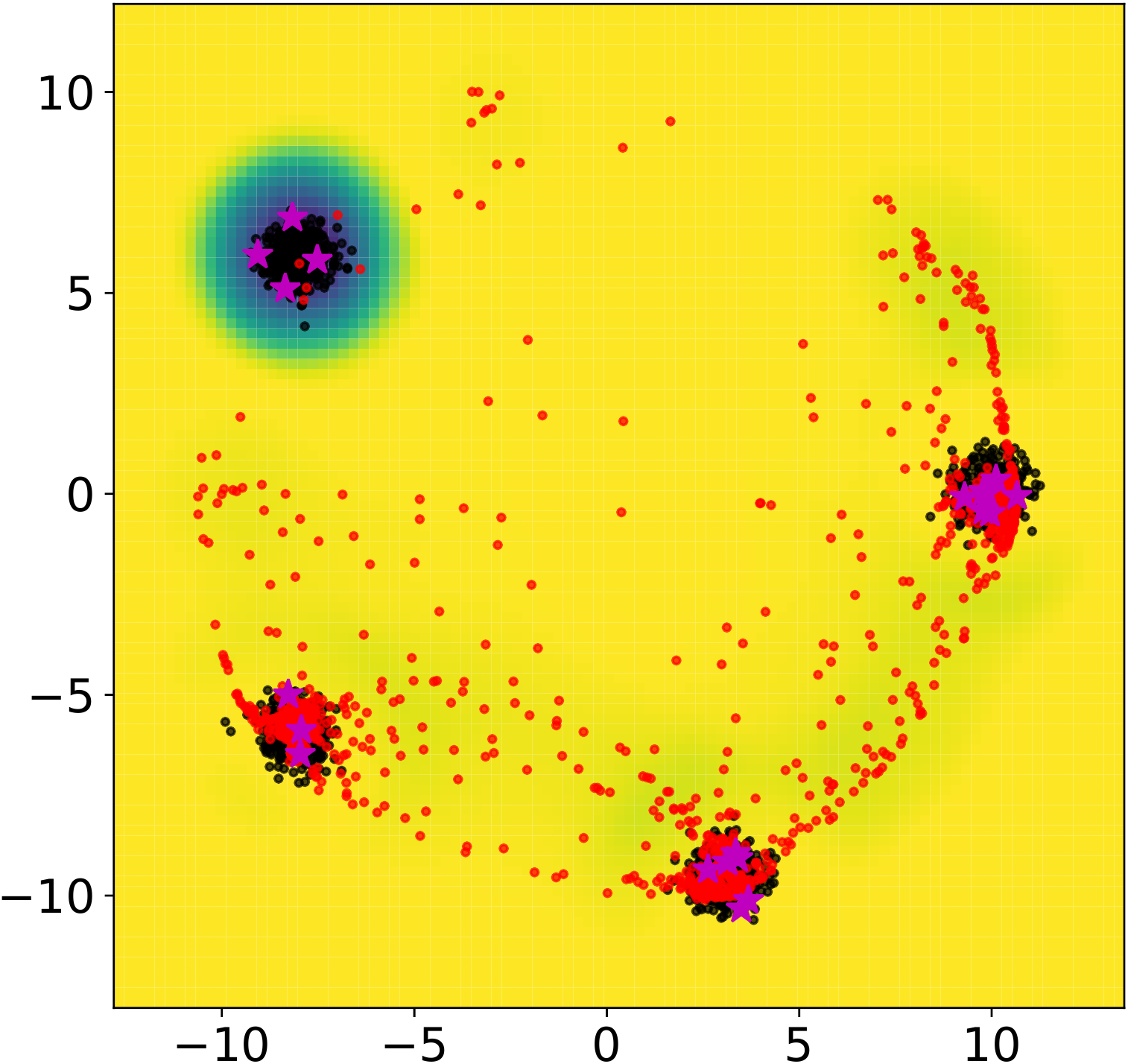}
            \caption{\scriptsize Phase $2$, iter =  $50$}
            \label{subfig:continual_learning_ume_d}
        \end{subfigure}
        \begin{subfigure}[c]{\eachw}
            \centering
            \includegraphics[width=\linewidth]{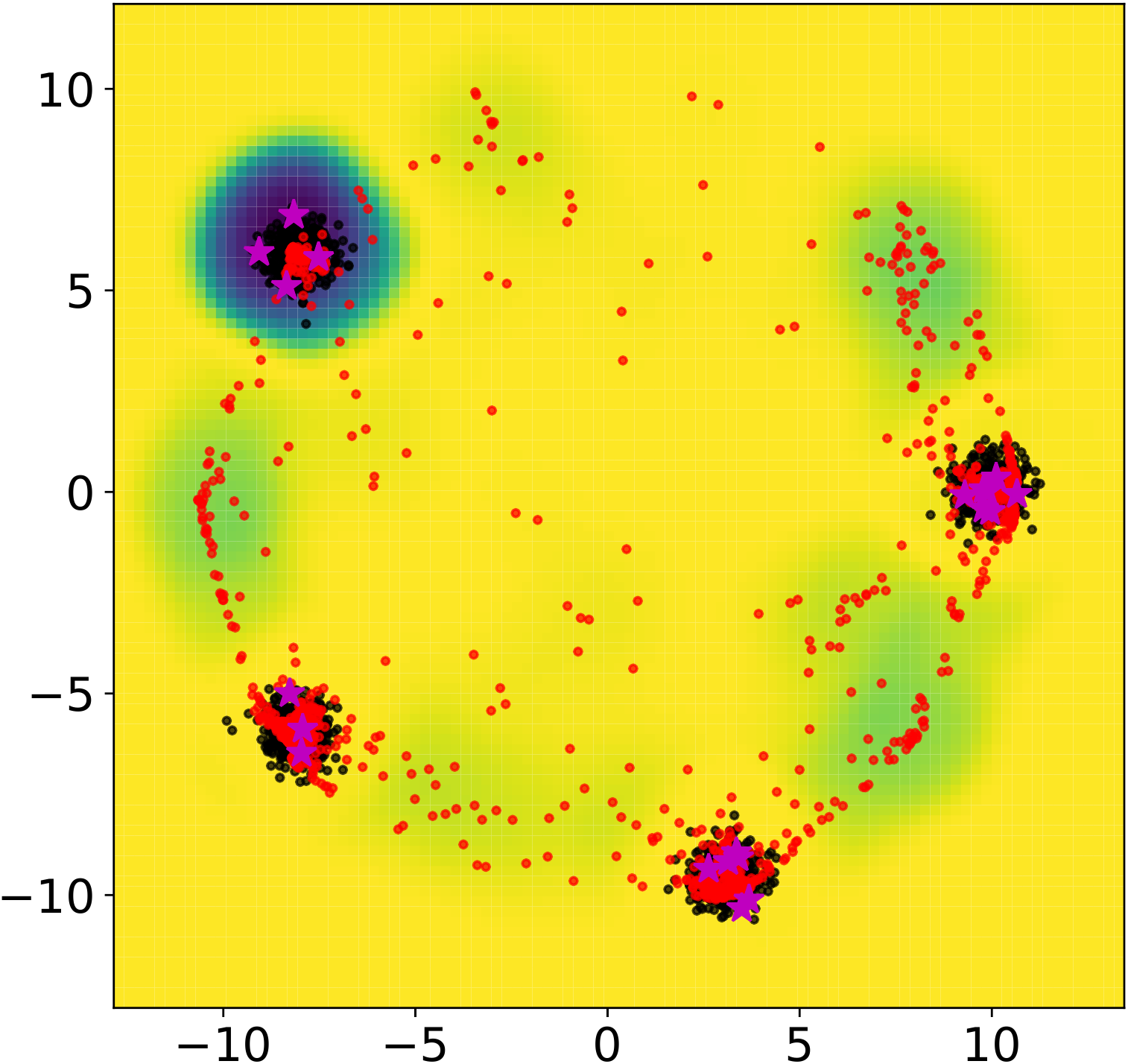}
            \caption{\scriptsize Phase $2$, iter =  $100$}
            \label{subfig:continual_learning_ume_e}
        \end{subfigure}
         \caption{The training progress of two generative models (Top row) MMD-GAN (Bottom row) MMD-GAN + witness points}
         \label{fig:continual_learning_mog}
\end{minipage}%
\end{figure*}

\subsection{Synthetic Data}
\label{sec:synthetic}

\textbf{Objective Function Augmented by UME.} In this section, we analyze our method using synthetic two-dimensional dataset. 
We aim to investigate the role of witness points in capturing the local structure of $P_\Xb$. We use a neural network to parameterize the generator. Witness points are parameterized as two-dimensional vectors in the input space.
The dataset used in this experiment is a mixture of Gaussian mixture models. 
The \emph{outer} mixture places $5$ components on a ring around the origin. 
Each of these components contains $3$ Gaussian components which are also located on a smaller ring (see~\Cref{subfig:momog_heatmap_a}). 
The evolution of the generator (red dots) and witness points (magenta stars) in~\Cref{fig:low_dims_2d_mog} shows that the global structure is initially captured by the generator; subsequently, the local structure of each component is captured by the witness points that act as guides for the generator to capture finer details of the distribution. Any missed mode is detected by the witness points which subsequently attract the probability mass of the generator. We observed that MMD-GAN (without UME term) failed to capture all the modes unless the learning rate is set to about $100$ times smaller which consequently makes the training much slower to converge.

Learning witness points in the latent space 
was introduced in~\Cref{subsec:latent_space} to allow presenting the
theoretical results for a more general setting of~\eqref{eq:dynamical_system}.
Our initial experiments confirm that learning witness points in the
latent space is feasible in practice and their interpretations are aligned with
our expectation (detecting regions of mismatch). See~\Cref{sec:mnist} for
details.

\paragraph{Introduce Prior Knowledge via Witness Points.} This section concerns a scenario where a generative model is already trained on a dataset $\Dcal_1$. Then, another dataset $\Dcal_2$ comes and we want the generative model to capture the new dataset too without forgetting the previous dataset. This can be seen as life-long learning in generative models~\cite{ramapuram2017lifelong}. ~\Cref{fig:continual_learning_mog} illustrates a similar scenario. The prior information here is the fact that $\Dcal_2$ comes after $\Dcal_1$ and we have access to samples from $\Dcal_2$. Choosing a few witness points from $\Dcal_2$ seems to be an effective way to incorporate this information in the training process. As can be seen in~\Cref{subfig:continual_learning_mmd_c}, it is hard for a generative model without witness points (MMD) to move some probability mass from $\Dcal_1$ to $\Dcal_2$. However, it is clear from~\Cref{subfig:continual_learning_mmd_c} that having witness points from $\Dcal_2$ in MMD+UME explicitly encourages this mass transfer. In addition, previously placed witness points in $\Dcal_1$ act as anchors and ensure that previously learned distribution won't be forgotten.

\begin{figure*}[!t]
\begin{minipage}[t]{\textwidth}
    \centering
        \begin{subfigure}[c]{0.85\textwidth}
            \centering
            \includegraphics[width=\linewidth]{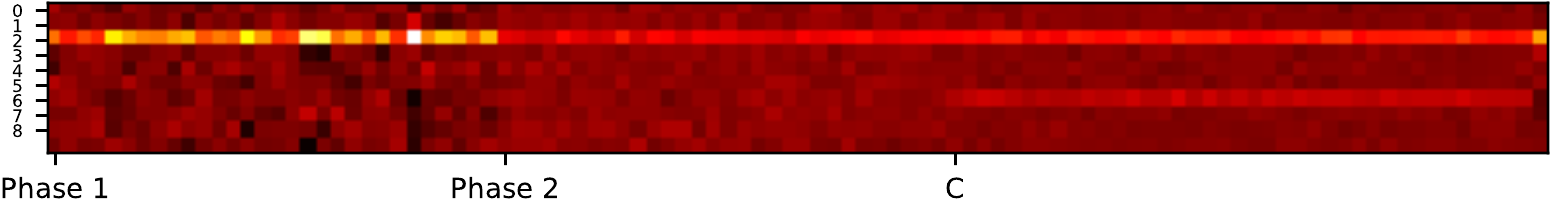}
            \caption{\scriptsize MMD}
            \label{fig:mnist_mmd_heatmap}
        \end{subfigure}
        \begin{subfigure}[c]{0.85\textwidth}
            \centering
            \includegraphics[width=\linewidth]{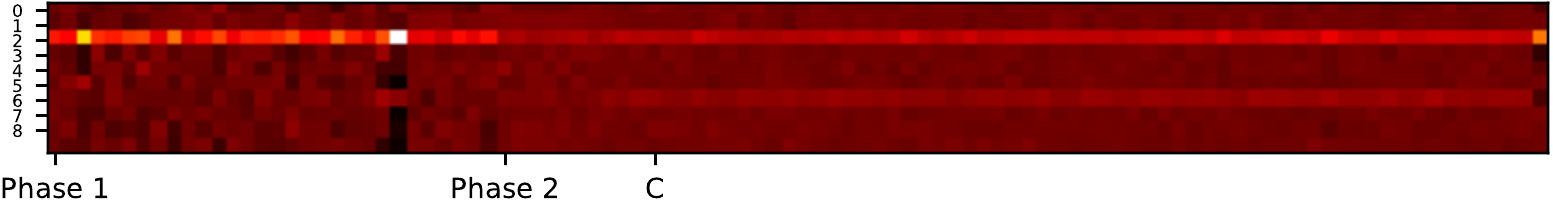}
            \caption{\scriptsize MMD+UME}
            \label{fig:nme_mnist_heatmap}
        \end{subfigure}
        \caption{The course of training and transition from Phase $1$ where
        only digit $2$ exists to Phase $2$ where the digits $2$ and $6$ both
    are present in the training set. The heatmap shows the normalized logits of
the classifier on the generated examples (the darker colors are closer to
zero). Phase $2$ in the $x$ axis shows the starting point of the second phase.
The point C in the same axis shows the moment in which the second mode is
captured by the generator. This is a point where the number of hits for the
second mode passes a threshold of $10\%$ of all samples. The corresponding
iterations to every moment on the $x$ axis: Phase 1: iter 1k (very initial
steps are removed due to lack of space), Phase 2: iter 5k, C(a): iter 9.7k,
C(b):6.1k.}
        \label{fig:mnist_heatmap}
\end{minipage}
\end{figure*}

\subsection{High Dimensional Data}
\label{sec:high_dimensional}
\begin{figure}[t]
    \centering
        %
            \includegraphics[width=\linewidth]{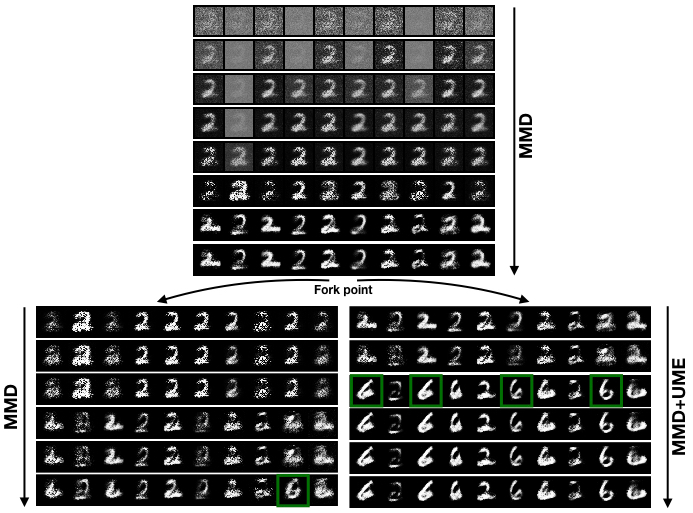}
        \caption{Witness points in the UME guide the generator to learn to capture unknown data modes. 
            \label{fig:mnist_fork}  
        The generated images by MMD and MMD+UME after both models are initially trained by MMD objective until the fork and a new mode is presented to the model after the fork. The rows from top to bottom shows the chronological sequence of generated images in the course of training.}
\end{figure}

In this section, we show the efficacy of witness points as a method to guide training in a high-dimensional case. Similar to the experiment of~\Cref{fig:continual_learning_mog}, we take $\Dcal$ as the entire MNIST dataset~\citep{lecun1998gradient}. We take $\Dcal_1$ as the dataset of the images corresponding to a random subset of labels. $\Dcal_2$ is the images corresponding to the complement set of digits. Let $\tau$ shows the training steps. We train a moment matching generative model~\citep{li2015generative} on $\Dcal_1$ from $\tau=0$ until $\tau=\tau_{\rm fork}$ with MMD objective. After $\tau_{\rm fork}$, we continue training in two branches (See~\Cref{fig:mnist_fork}). In one branch, witness points from $\Dcal_2$ are introduced and the MMD+UME objective is optimized on $\Dcal_1 \cup \Dcal_2$. In another branch, the training is continued as before the fork by optimizing the MMD objective on $\Dcal_1 \cup \Dcal_2$. We use a pre-trained MNIST classifier to monitor the state of the generators after the fork. We observe that the presence of witness points directs the generator towards the incoming mode faster compared with when only MMD distance is optimized in the absence of witness points. As can be seen in~\Cref{fig:mnist_heatmap}, both models perform equally before $\tau_{\rm fork}$, but the guiding signal from the witness points enables the MMD+UME model to capture the second mode faster. The details of this experiment are provided in~\Cref{sec_appendix:experiments}. We run the same experiment with a different $\Dcal_1$ and $\Dcal_2$ and constantly observed that the second mode is captured faster. This implies that the learning signal from witness points is strong enough to guide the training at the initial steps of transition between phases.


\section{Conclusion}
\label{sec:related-work}

We introduced witness points as a dynamic component with theoretical
justification to guide the training trajectories of implicit generative models.
We showed its efficacy using multiple experiments and provided stability
guarantees ensuring that incorporating these points does not harm the stability
of current algorithms. Applications of the witness points in areas such as
monitoring the course of training, and interpreting the missing modes can
be ideas for future work.

\bibliography{refs}

\begin{thebibliography}{10}

\bibitem{Arbel:2018}
Michael Arbel, Dougal Sutherland, Mikolaj Binkowski, and Arthur Gretton.
\newblock On gradient regularizers for mmd gans.
\newblock {\em NeurIPS}, 2018.

\bibitem{arjovsky2017wasserstein}
M.~Arjovsky, S.~Chintala, and L.~Bottou.
\newblock {W}asserstein generative adversarial networks.
\newblock In {\em ICML}, 2017.

\bibitem{arjovsky2017towards}
Martin Arjovsky and L{\'e}on Bottou.
\newblock Towards principled methods for training generative adversarial
  networks.
\newblock In {\em International Conference on Learning Representations}, 2017.

\bibitem{arora2017gans}
Sanjeev Arora, Andrej Risteski, and Yi~Zhang.
\newblock Do {GAN}s learn the distribution? some theory and empirics.
\newblock In {\em International Conference on Learning Representations}, 2018.

\bibitem{BinSutArbGre2018}
Mikolaj Bi\'{n}kowski, Dougal Sutherland, Michael Arbel, and Arthur Gretton.
\newblock Demystifying {MMD} {GAN}s.
\newblock {\em International Conference on Learning Representations}, 2018.

\bibitem{brock2018large}
Andrew Brock, Jeff Donahue, and Karen Simonyan.
\newblock Large scale gan training for high fidelity natural image synthesis.
\newblock {\em arXiv preprint arXiv:1809.11096}, 2018.

\bibitem{chen2016infogan}
Xi~Chen, Yan Duan, Rein Houthooft, John Schulman, Ilya Sutskever, and Pieter
  Abbeel.
\newblock Infogan: Interpretable representation learning by information
  maximizing generative adversarial nets.
\newblock In {\em Advances in neural information processing systems}, pages
  2172--2180, 2016.

\bibitem{ChwRamSejGre2015}
Kacper~P Chwialkowski, Aaditya Ramdas, Dino Sejdinovic, and Arthur Gretton.
\newblock Fast two-sample testing with analytic representations of probability
  measures.
\newblock In {\em Advances in Neural Information Processing Systems 28}, pages
  1981--1989. Neural Information Processing Systems 2015, 2015.

\bibitem{dai2017towards}
Bo~Dai, Sanja Fidler, Raquel Urtasun, and Dahua Lin.
\newblock Towards diverse and natural image descriptions via a conditional gan.
\newblock In {\em Proceedings of the IEEE International Conference on Computer
  Vision}, pages 2970--2979, 2017.

\bibitem{daskalakis2018training}
Constantinos Daskalakis, Andrew Ilyas, Vasilis Syrgkanis, and Haoyang Zeng.
\newblock Training {GAN}s with optimism.
\newblock In {\em International Conference on Learning Representations}, 2018.

\bibitem{denton2015deep}
Emily~L Denton, Soumith Chintala, Rob Fergus, et~al.
\newblock Deep generative image models using laplacian pyramid of adversarial
  networks.
\newblock In {\em Advances in neural information processing systems}, pages
  1486--1494, 2015.

\bibitem{dziugaite2015training}
Gintare~Karolina Dziugaite, Daniel~M Roy, and Zoubin Ghahramani.
\newblock Training generative neural networks via maximum mean discrepancy
  optimization.
\newblock In {\em UAI}, 2015.

\bibitem{fukumizu}
Kenji Fukumizu, Arthur Gretton, Xiaohai Sun, and Bernhard Sch\"{o}lkopf.
\newblock Kernel measures of conditional dependence.
\newblock In J.~C. Platt, D.~Koller, Y.~Singer, and S.~T. Roweis, editors, {\em
  Advances in Neural Information Processing Systems 20}, pages 489--496. Curran
  Associates, Inc., 2008.

\bibitem{GarJitKan2017}
Damien {Garreau}, Wittawat {Jitkrittum}, and Motonobu {Kanagawa}.
\newblock {Large sample analysis of the median heuristic}.
\newblock {\em arXiv e-prints}, July 2017.

\bibitem{goodfellow_gan_2014}
Ian Goodfellow, Jean Pouget-Abadie, Mehdi Mirza, Bing Xu, David Warde-Farley,
  Sherjil Ozair, Aaron Courville, and Yoshua Bengio.
\newblock Generative adversarial networks.
\newblock In {\em Advances in Neural Information Processing Systems}, pages
  2672--2680, 2014.

\bibitem{GreBorRasSchSmo2012}
Arthur Gretton, Karsten~M. Borgwardt, Malte~J. Rasch, Bernhard Sch\"{o}lkopf,
  and Alexander Smola.
\newblock A kernel two-sample test.
\newblock {\em Journal of Machine Learning Research}, 13(Mar):723--773, 2012.

\bibitem{haykin2008adaptive}
Simon~S Haykin.
\newblock {\em Adaptive filter theory}.
\newblock Pearson Education India, 2008.

\bibitem{hinton2006reducing}
Geoffrey~E Hinton and Ruslan~R Salakhutdinov.
\newblock Reducing the dimensionality of data with neural networks.
\newblock {\em science}, 313(5786):504--507, 2006.

\bibitem{hu2018deep}
Zhiting Hu, Zichao Yang, Ruslan~R Salakhutdinov, LIANHUI Qin, Xiaodan Liang,
  Haoye Dong, and Eric~P Xing.
\newblock Deep generative models with learnable knowledge constraints.
\newblock In {\em Advances in Neural Information Processing Systems}, pages
  10501--10512, 2018.

\bibitem{JitSzaChwGre2016}
Wittawat Jitkrittum, Zolt{\'a}n Szab{\'o}, Kacper~P Chwialkowski, and Arthur
  Gretton.
\newblock Interpretable distribution features with maximum testing power.
\newblock In {\em Advances in Neural Information Processing Systems}, pages
  181--189, 2016.

\bibitem{karras2018progressive}
Tero Karras, Timo Aila, Samuli Laine, and Jaakko Lehtinen.
\newblock Progressive growing of {GAN}s for improved quality, stability, and
  variation.
\newblock In {\em International Conference on Learning Representations}, 2018.

\bibitem{lecun1998gradient}
Yann LeCun, L{\'e}on Bottou, Yoshua Bengio, and Patrick Haffner.
\newblock Gradient-based learning applied to document recognition.
\newblock {\em Proceedings of the IEEE}, 86(11):2278--2324, 1998.

\bibitem{li2017mmd}
Chun-Liang Li, Wei-Cheng Chang, Yu~Cheng, Yiming Yang, and Barnab{\'a}s
  P{\'o}czos.
\newblock {MMD GAN}: Towards deeper understanding of moment matching network.
\newblock In {\em Advances in Neural Information Processing Systems}, pages
  2203--2213, 2017.

\bibitem{li2017adversarial}
Jiwei Li, Will Monroe, Tianlin Shi, S{\.{e}}bastien Jean, Alan Ritter, and Dan
  Jurafsky.
\newblock Adversarial learning for neural dialogue generation.
\newblock In {\em Proceedings of the 2017 Conference on Empirical Methods in
  Natural Language Processing}, pages 2157--2169. Association for Computational
  Linguistics, 2017.

\bibitem{li2015generative}
Yujia Li, Kevin Swersky, and Rich Zemel.
\newblock Generative moment matching networks.
\newblock In {\em International Conference on Machine Learning}, pages
  1718--1727, 2015.

\bibitem{loomis2013introduction}
Lynn~H Loomis.
\newblock {\em Introduction to abstract harmonic analysis}.
\newblock Courier Corporation, 2013.

\bibitem{MehSch18}
A.~Mehrjou and B.~Sch{\"o}lkopf.
\newblock Nonstationary {GANs}: Analysis as nonautonomous dynamical systems.
\newblock In {\em Workshop on Theoretical Foundations and Applications of Deep
  Generative Models at ICML}, July 2018.

\bibitem{mirza2014conditional}
Mehdi Mirza and Simon Osindero.
\newblock Conditional generative adversarial nets.
\newblock {\em arXiv preprint arXiv:1411.1784}, 2014.

\bibitem{MuaFukSriSch17}
K.~Muandet, K.~Fukumizu, B.~Sriperumbudur, and B.~Sch{\"o}lkopf.
\newblock Kernel mean embedding of distributions: A review and beyond.
\newblock {\em Foundations and Trends in Machine Learning}, 10(1-2):1--141,
  2017.

\bibitem{Mueller97:IPM}
Alfred M\"uller.
\newblock Integral probability metrics and their generating classes of
  functions.
\newblock {\em Advances in Applied Probability}, 29(2):429--443, 1997.

\bibitem{nagarajan2017gradient}
Vaishnavh Nagarajan and J~Zico Kolter.
\newblock Gradient descent {GAN} optimization is locally stable.
\newblock In {\em Advances in Neural Information Processing Systems}, pages
  5585--5595, 2017.

\bibitem{NeyBhoCha2017}
B.~{Neyshabur}, S.~{Bhojanapalli}, and A.~{Chakrabarti}.
\newblock {Stabilizing GAN Training with Multiple Random Projections}.
\newblock {\em ArXiv e-prints}, May 2017.

\bibitem{NowCseTom2016}
Sebastian Nowozin, Botond Cseke, and Ryota Tomioka.
\newblock {f-GAN}: Training generative neural samplers using variational
  divergence minimization.
\newblock In {\em Advances in Neural Information Processing Systems}, pages
  271--279, 2016.

\bibitem{ramapuram2017lifelong}
Jason Ramapuram, Magda Gregorova, and Alexandros Kalousis.
\newblock Lifelong generative modeling.
\newblock {\em arXiv preprint arXiv:1705.09847}, 2017.

\bibitem{sajjadi2018tempered}
Mehdi~SM Sajjadi, Giambattista Parascandolo, Arash Mehrjou, and Bernhard
  Sch{\"o}lkopf.
\newblock Tempered adversarial networks.
\newblock In {\em ICLR Workshop}, 2018.

\bibitem{strang1993introduction}
Gilbert Strang.
\newblock {\em Introduction to linear algebra}, volume~3.
\newblock Wellesley-Cambridge Press Wellesley, MA, 1993.

\bibitem{sutherland2017generative}
Dougal~J Sutherland, Hsiao-Yu Tung, Heiko Strathmann, Soumyajit De, Aaditya
  Ramdas, Alex Smola, and Arthur Gretton.
\newblock Generative models and model criticism via optimized maximum mean
  discrepancy.
\newblock In {\em International Conference on Learning Representations}, 2017.

\bibitem{wang2018improving}
Yuan~Sun Wei~Wang and Saman Halgamuge.
\newblock Improving {MMD}-{GAN} training with repulsive loss function.
\newblock In {\em International Conference on Learning Representations}, 2019.

\bibitem{zhao2016energy}
Junbo Zhao, Michael Mathieu, and Yann LeCun.
\newblock Energy-based generative adversarial network.
\newblock {\em arXiv preprint arXiv:1609.03126}, 2016.

\bibitem{zhu2017unpaired}
Jun-Yan Zhu, Taesung Park, Phillip Isola, and Alexei~A Efros.
\newblock Unpaired image-to-image translation using cycle-consistent
  adversarial networks.
\newblock In {\em Computer Vision (ICCV), 2017 IEEE International Conference
  on}, 2017.

\end{thebibliography}
\bibliographystyle{plain}


\onecolumn
\appendix

\begin{center}
{\LARGE{}{}{}{}{}{}\ourtitle{}} 
\par\end{center}

\begin{center}
\textcolor{black}{\Large{}{}{}{}{}{}Supplementary}{\Large{}{}{}{}{}
} 
\par\end{center}
\section{\autoours Algorithm}
\begin{algorithm}[h!] 
    \caption{\autoours}
    \label{alg:autoours}
     
    \SetKwInOut{Input}{input}\SetKwInOut{Output}{output}
    \Input{$J$, $\lambda$, $\gamma$, $c$, $B$, $n_g, n_v$ as Alg.~\ref{alg:ours} and $n_e, n_d$ the training iterations for encoder and decoder.}
    do initialization as Alg.~\ref{alg:ours};
     initialize encoder and decoder parameters $\thetade, \thetadd$;
     define the convergence criterion;
     
    
    \Output{$G_{\thetagen}(\zbm)$, $\Vset$ as Alg.~\ref{alg:ours} and the decoder $D_\thetad(\cdot)$ .}

     \While{convergence criterion is not met}{
     
        \For{$t=1,\ldots,n_d$}{
        Sample minibatches $\Xset=\{\xbm_i\}_{i=1}^B \sim P_\Xb$ and $\Zset=\{\zbm_j\}_{j=1}^B \sim P_\Zb$ \\
            $\thetade \leftarrow \thetade - \gamma \cdot \nabla_\thetade \Lcal^{r}_{(\thetade,\thetadd)}(\Xset, \Zset)$\\ 
            $\thetadd \leftarrow \thetadd - \gamma \cdot \nabla_\thetadd \Lcal^{r}_{(\thetade,\thetadd)}(\Xset, \Zset)$\\ 
        }
        \For{$t=1,\ldots,n_e$}{
        Sample minibatches $\Xset$ and $\Zset$\\
            $\thetade \leftarrow \thetade + \gamma \cdot \nabla_\thetade \Lcal_{(\thetade,\thetagen,\Vset)}(\Xset, \Zset)$\\ 
        }
        \For{$t=1,\ldots,n_g$}{
        Sample minibatches $\Xset$ and $\Zset$ \\
            $\thetagen \leftarrow \thetagen - \gamma \cdot \nabla_\thetagen \Lcal_{(\thetade,\thetagen,\Vset)}(\Xset, \Zset)$\\ 
        }
        \For{$t=1,\ldots,n_v$}{
        Sample minibatches $\Xset$ and $\Zset$ \\
            \For{$j=1,\ldots,J$}{
                $\vj \leftarrow \vj + \gamma \cdot \nabla_\vj \Lcal_{(\thetade,\thetagen,\Vset)}(\Xset, \Zset)$ \\ 
            }
        }
     }
\end{algorithm}

\subsection{Experiment: Witness Points in a Latent Space}
\label{sec:mnist}

\begin{figure}[t!]
\centering
\begin{minipage}[t]{.48\textwidth}
     \centering
    \begin{flushleft}
    \includegraphics[width=0.95\textwidth]{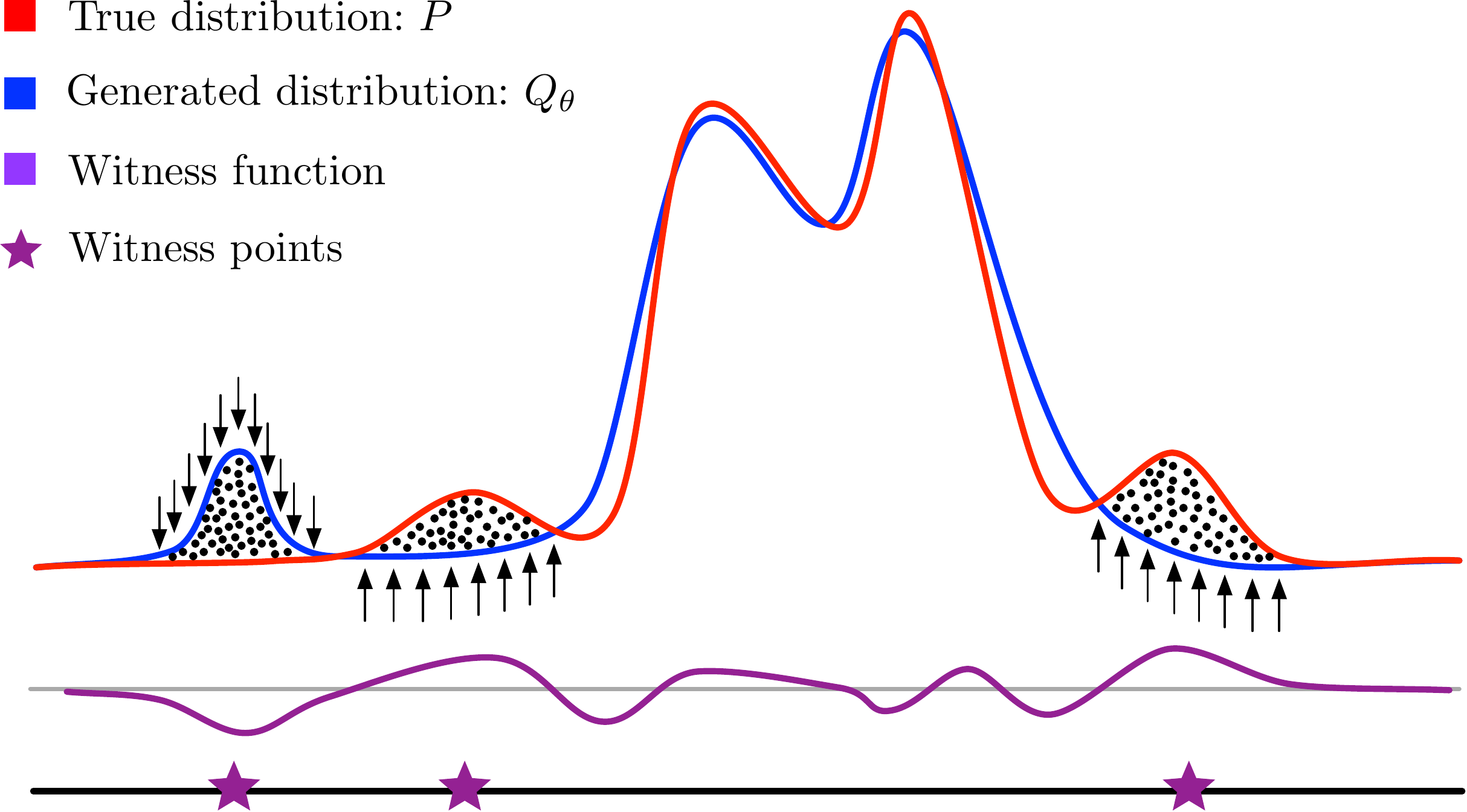}
    \caption{\label{fig:global-local}The witness points help locate the \emph{local} differences between the true and synthetic distributions, which are then used in guiding the model to capture the missed regions (as shown by arrows).
    }
    \end{flushleft}
\end{minipage}%
\hspace{4mm}
\begin{minipage}[t]{.48\textwidth}
    \includegraphics[width=0.95\textwidth]{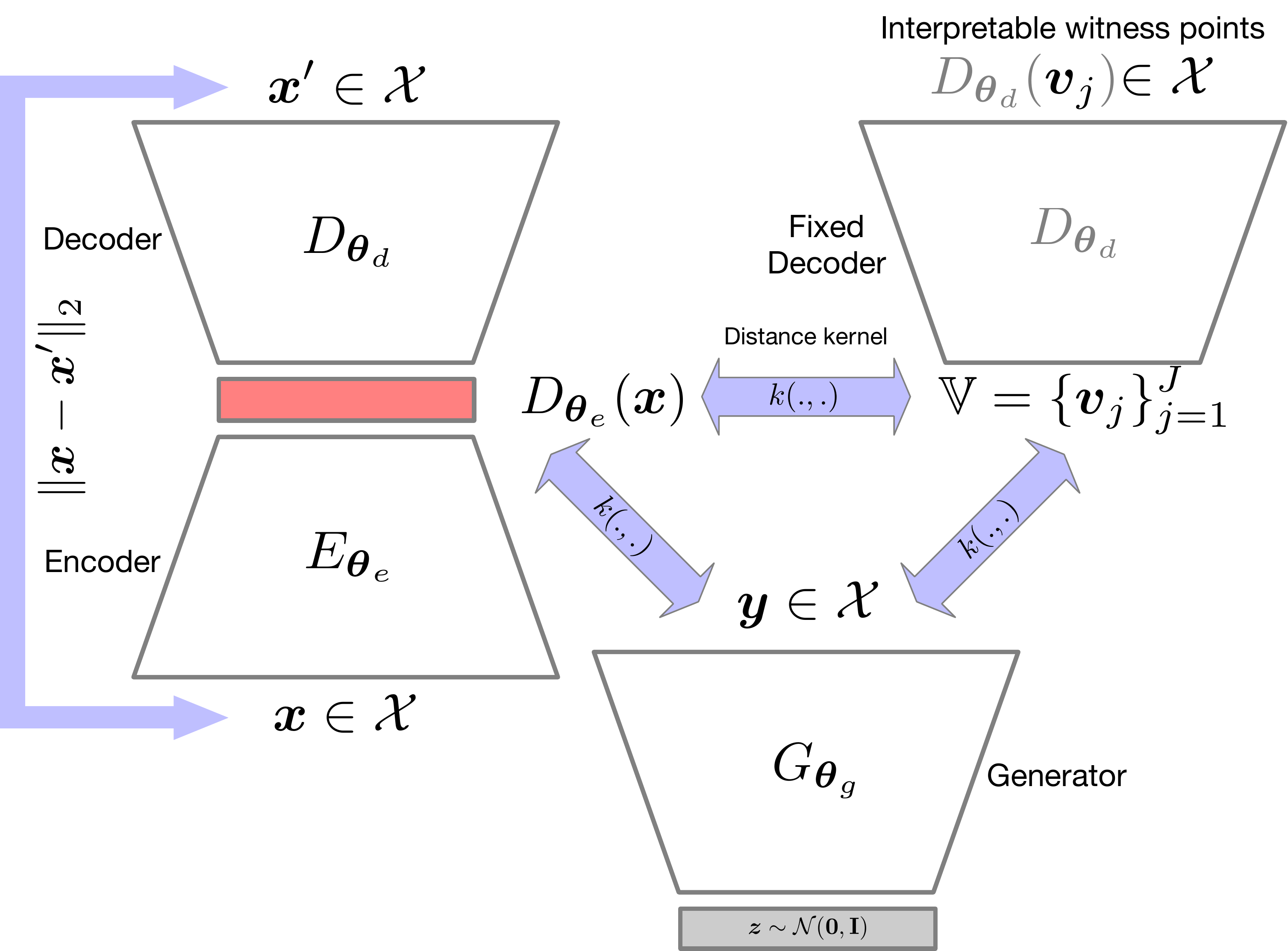}
    \caption{The schematic of \autoours where the witness points $\vj$ live in
        the latent space. The trained decoder is used to visualize the witness
        points.
    }
   \label{fig:algorithm_schematics}

\end{minipage}%
\end{figure}

\begin{figure*}
\begin{minipage}[t]{\textwidth}
    \centering
        \begin{subfigure}[c]{0.24\textwidth}
            \centering
            \includegraphics[width=\linewidth]{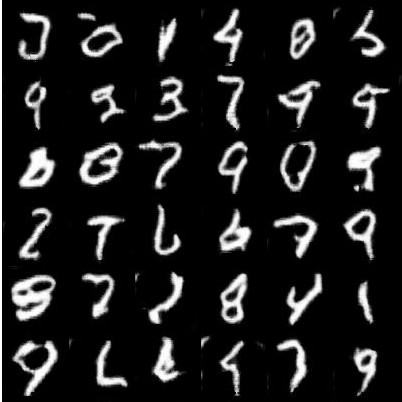}
            \caption{\scriptsize early witness}
            \label{fig: autoglocad_mnist1}
        \end{subfigure}
        \begin{subfigure}[c]{0.24\textwidth}
            \centering
            \includegraphics[width=\linewidth]{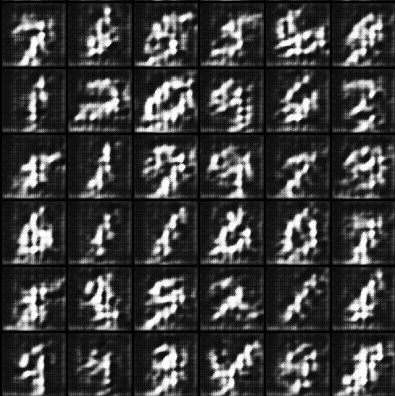}
            \caption{\scriptsize Phase $2$, iter =  $10$}
            \label{fig: autoglocad_mnist2}
        \end{subfigure}
        \begin{subfigure}[c]{0.24\textwidth}
            \centering
            \includegraphics[width=\linewidth]{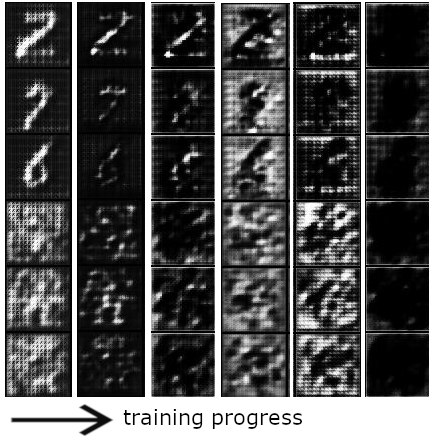}
            \caption{\scriptsize late witness}
            \label{fig: autoglocad_mnist3}
        \end{subfigure}
        \begin{subfigure}[c]{0.24\textwidth}
            \centering
            \includegraphics[width=\linewidth]{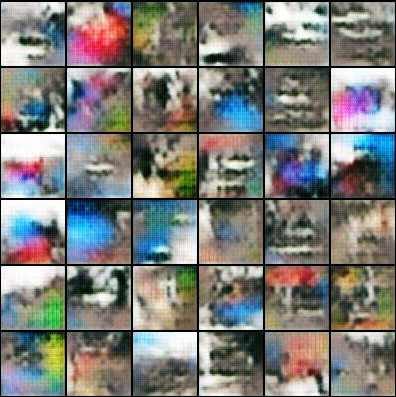}
            \caption{\scriptsize witness points on CIFAR10, iter=30k}
            \label{fig: autoglocad_cifar}
        \end{subfigure}
        \caption{\footnotesize(a) Generated samples in a middle stage of the training process (b) The decoded learned witness points at early stages of \autoours (c) The progress of the witness points. Upper three rows are initialized to samples from the training set ($\xbm\in\Xcal$) and the lower three rows are initialized to ($G(\zbm)\in\Ycal$). Initial states show the complete distinctions (almost clean images from $P_\Xb$ and unstructured noise from $Q_\Yb$. In later stages of training, this clear distinction vanishes as the witness points become more and more similar. (d) Some witness points learned for CIFAR10 dataset at intermediate stages of training.}
        \label{fig:mnist}
\end{minipage}
\end{figure*}

To showcase the performance of \autoours and also the flexibility of the algorithm with respect to the first term of~\eqref{eq:total_loss} that is an arbitrary loss of an implicit generative model, we tested our method on a high-dimensional dataset, MNIST~\citep{lecun1998gradient} and Energy-Based GAN~\citep{zhao2016energy} as the employed generative model. Energy-Based GAN conceptually fits well in our framework since its discriminator is an autoencoder. Hence, we do not need to employ an extra autoencoder to encode the witness points as it is required in~\Cref{alg:autoours} and will be detailed in the following. According to~\Cref{alg:autoours}, the witness points are defined and optimized in the latent space of this autoencoder. The results can be seen in~\Cref{fig:mnist}. As depicted in~\Cref{fig: autoglocad_mnist3}, the progress of the witness points is informative about the training procedure. The algorithm is given $20$ witness points, half of which initialized to $E_\thetade(\xbm)$ (for $\xbm\in\mathrm{training\ set}$) and the other half initialized to $E_\thetade(\ybm)$ (for $\ybm=G_\thetagen(\zbm)$). The upper three rows of~\Cref{fig: autoglocad_mnist3} show the progress of the first set and the lower three rows show the progress of the second set. In the beginning, when $P_\Xb$ and $Q_\Yb$ are completely distinct, the witness points converge to the support of the distributions (almost clear digits belonging to $P_\Xb$ in the upper rows and noisy images belonging to $Q_\Yb$ in the lower rows). As the training proceeds, two sets of witness points become more and more similar implying that the distinctions between two distributions are getting smaller. Whenever a mismatch occurs between $P_\Xb$ and $Q_\Yb$, a structured image appears in the witness points suggesting that the distinction is more than noise. When the distributions match, we expect to see unstructured images in the witness points. We did the same experiment on CIFAR10 dataset and observed the same condition. The learned witness points for intermediate stages of training when there still exists some local differences between true and generated distributions is shown in~\Cref{fig: autoglocad_cifar}. Even though the images are not meaningful, we conjecture the witness points (when decoded from the latent space by the decoder of the autoencoder) can be used to find the regions of mismatch between the true and generated distributions as a diagnostic tool to probe the missed modes of the target distribution. We postpone more exploration in this direction to future work.

\newpage
\section{Stability Analysis}
\label{sec_appendix:stability_analysis}
 In the following, we study the equilibria and the dynamical behaviour of the
 parameters of \ours at the equilibrium. Our analysis is similar to that
 presented in \cite{wang2018improving} and \cite{nagarajan2017gradient}.
 Recall the following loss functions which are considered in \ours:
\begin{subequations}\label{eq:overall-loss}
   \begin{eqnarray}
    L^{\text{mmd}} &=&\set{E}_{P_\Xb}[k_D(\xbm,\xbm^{\prime})]-2\set{E}_{P_\Xb,Q_\Yb}[k_D(\ybm,\ybm)]+\set{E}_{Q_\Yb}[k_D(\ybm,\ybm^{\prime})] \label{eq:loss-mmd} \\
    L^{\text{ume}} &=& \frac{1}{J}\sum_{j=1}^{J} \left(\mathbb{E}_{P_\Xb}[ k_D(\xbm,\vbm_j)] - \mathbb{E}_{Q_\Yb}[ k_D(\ybm,\vbm_j)]\right)^2 \label{eq:loss-ume} \\
    \mathcal{L} &=& L^{\text{mmd}}+\lambda L^{\text{ume}}\label{eq_appendix:total_loss}
   \end{eqnarray}
\end{subequations}
where $k_D(\xbm,\ybm)=k(D(\xbm),D(\ybm))$ and $D:\Xcal\to\Kcal$ is the feature extractor which we will also call the discriminator interchangeably. Let $\thetagen\in\Theta_G$ and $\thetad\in\Theta_D$ be the parameters of the generator and the discriminator respectively. Let $\Vset=\vjs$ and $\vbm_j\in \Vcal$ for $\Vcal\subseteq\Kcal$. The trainable parameters of \ours are $\{\thetagen,\thetad,\Vset\}$. Using gradient based optimization methods to find the optima of these losses gives us the following dynamical system when the learning rate of the optimization algorithm tends to zero ($\gamma \to 0$):

\begin{equation}
\label{eq_appendix:dynamical_system}
\left\{
    \begin{array}{ll}
        \dot{\bm{\theta}}_g=\shortminus \nabla_{\thetagen}\mathcal{L}(\thetagen,\thetad,\Vset)\\
        \dot{\bm{\theta}}_d=\nabla_{\thetad}\mathcal{L}(\thetagen,\thetad,\Vset)\\
        \dot{\Vset}_\vbm=\nabla_{\vbm}\mathcal{L}(\thetagen,\thetad,\Vset)\\
    \end{array}
\right.
\end{equation}
We proceed by making the following assumption about the equilibria of \ours.
\begin{assumption_s}{\label{assumption_appendix:equilibrium}}
At equilibrium $(\thetagopt, \thetad, \Vset)$, we have $P_{\thetagopt}=P_\Xb$ and $D_{\thetadopt}(\xbm)$ is injective on $\supp(P_\Xb)\bigcup\supp(P_{\thetagopt})$.
\end{assumption_s}

The above assumption implies that, at equilibrium, the set of witness points $\Vset=\vjs$ can be arbitrary, that is, $L^{\text{ume}} = 0$ for any non-empty set $\Vset$ while being at equilibrium. In what follows, we provide the detailed proof of Proposition \ref{prop:equilibrium}.

\subsection{Existence of Equilibria}
\label{sec_appendix:equilibrium_existence_proof}

\begin{proposition_s}
If the generator is realizable, i.e., there exists a $\thetagopt\in\Theta_G$ such that $Q_\Yb=P_\Xb$, then the dynamical system of~\eqref{eq_appendix:dynamical_system} has equilibria $(\thetagopt,\thetad, \Vset)$ for any $\thetad\in\Theta_D$ and $\Vset\in\Vcal^J$.
\end{proposition_s}
\begin{proof}
Here, we expand the gradient of the loss~\eqref{eq:overall-loss} w.r.t. each set of trainable parameters to obtain the rhs of~\eqref{eq_appendix:dynamical_system}. Let $\xbm\sim P_\Xb$, $\zbm\sim P_\Zb$, $\Yb=G_{\thetagen}(\Zb)\sim Q_{\Yb}$. Suppose $\xbm$, $\zbm$ and $\ybm$ be realized samples from these distributions. Let denote the output of the $D(\cdot)$ by $\bm{d}$, i.e., $\bm{d}_\xbm = D(\xbm)$ and $\bm{d}_\ybm = D(\ybm)$. To ease the derivations, we assume that the kernel is isotropic stationary, i.e.,  $k(\xbm,\ybm)=\psi(\lVert \xbm - \ybm\rVert)$. The first term of~\eqref{eq_appendix:total_loss} is independent of $\Vset$ and was proved to be stable in~\citep{wang2018improving}. 

The second term $L^{\text{ume}}$ is a function of all three set of trainable variables $\{\thetagen, \thetad, \Vset\}$ and can be written as
\begin{align*}
 L^{\text{ume}}(\thetagen, \thetad, \Vset) &= \frac{1}{J}\sum_{j=1}^{J} \left(\expectation_{\xbm\sim P_{\Xb}}[ k(D_\thetad(\xbm),\vbm_j)] - \expectation_{\ybm\sim Q_{\Yb}}[ k(D_\thetad(\ybm),\vbm_j)]\right)^2 \\
&= \frac{1}{J}\sum_{j=1}^{J} \Big(\mathbb{E}^2_{\xbm}[ k(D_\thetad(\xbm),\vbm_j)] -2\mathbb{E}_{\xbm}[ k(D_\thetad(\xbm),\vbm_j)]\mathbb{E}_{\ybm}[ k(D_\thetad(\ybm),\vbm_j)] \\ 
&+ \mathbb{E}^2_{\ybm}[ k(D_\thetad(\ybm),\vbm_j)]\Big) ,
\end{align*}
\noindent where $\mathbb{E}_{\xbm}[\cdot] \coloneqq  \mathbb{E}_{\xbm\sim P_{\Xb}}[\cdot]$ and  $\mathbb{E}_{\ybm}[\cdot] \coloneqq  \mathbb{E}_{\ybm\sim Q_{\Yb}}[\cdot]$. To simplify the notation, let $\bm{e}_{\bm{xv}}^j=D(\xbm)-\vbm_j$ and $\bm{e}_{\bm{yv}}^j=D(\ybm)-\vbm_j$, emphasizing our assumption that the kernel is isotropic. With slight abuse of notation, we simply use $k$ instead of $\psi$ where $k(\xbm,\ybm)=\psi(\xbm-\ybm)$. Because $\ybm=G_\thetagen(\zbm)$, we rewrite $\mathbb{E}_\ybm[\cdot]$ as $\mathbb{E}_\zbm[\cdot]$. This re-parameterization is done since computing $\partial\mathbb{E}_\zbm[\cdot]/\partial \thetagen$ is much more convenient than $\partial\mathbb{E}_\ybm[\cdot]/\partial \thetagen$ due to the fact that the distribution with respect to which the expectation is taken is no longer a function of $\thetagen$ and we can swap the expectation with the derivation operators. Then, the dynamic of the system w.r.t. the parameters of the generator can be written as

\begin{eqnarray}
        \dot{\bm{\theta}}_g &=&
        -\frac{1}{J}\sum_{j=1}^{J} \Bigg(2\mathbb{E}_\xbm\left[k_D(\xbm,\vj)\right]\mathbb{E}_\xbm\left[\pder[k_D]{\exvj}\pder[\exvj]{\thetagen}\right]
        -2\mathbb{E}_\xbm\left[k_D(\xbm, \vj)\right]\mathbb{E}_\zbm\left[\pder[k_D]{\eyvj}\pder[\eyvj]{\thetagen}\right] \nonumber \\
        &&-2\mathbb{E}z\ybm\left[k_D(\ybm, \vj)\right]\mathbb{E}_\xbm\left[\pder[k_D]{\exvj}\pder[\exvj]{\thetagen}\right] 
        + 2\mathbb{E}z\ybm\left[k_D(\ybm,\vj)\right]\mathbb{E}_\zbm\left[\pder[k_D]{\eyvj}\pder[\eyvj]{\thetagen}\right]\label{eq:appendix_dynamics_g_firstline} \Bigg)\\
        &=&-\frac{1}{J}\sum_{j=1}^{J} \Bigg( 2\left(\mathbb{E}_\zbm\left[k_D(\ybm,\vj)\right]-\mathbb{E}_\xbm\left[k_D(\xbm, \vj)\right]\right)\mathbb{E}_\zbm\left[\pder[k_D]{\eyvj}\pder[\eyvj]{\thetagen}\right]\label{eq:appendix_dynamics_g_secondline} \Bigg)\\
        &=& 0 ,
\end{eqnarray}
\noindent where we used the fact that $\pder[\exvj]{\thetagen} = 0$ to derive \eqref{eq:appendix_dynamics_g_secondline}. The last equality \eqref{eq:appendix_dynamics_g_secondline} follows from the fact that $\expectation_\zbm[k_D(\ybm,\vj)]-\expectation_\xbm[k_D(\xbm, \vj)]$ vanishes at the equilibrium (when $P_\Xb=Q_\Yb$), see Assumption ~\ref{assumption_appendix:equilibrium}.

Similarly, the dynamics of the parameters of the feature extractor $\thetad$ can be written as
\begin{eqnarray}
    \dot{\bm{\theta}}_d &=&
        \frac{1}{J}\sum_{j=1}^{J}\Bigg(2\mathbb{E}_\xbm[k_D(\xbm,\vj)]\mathbb{E}_\xbm\left[\pder[k_D]{\exvj}\pder[\exvj]{\thetad}\right]
        -2\mathbb{E}_\xbm[k_D(\xbm, \vj)]\mathbb{E}_\zbm\left[\pder[k_D]{\eyvj}\pder[\eyvj]{\thetad}\right]\nonumber \\
        &&     \qquad\qquad -2\mathbb{E}_\zbm[k_D(\ybm,\vj)]\mathbb{E}_\xbm\left[\pder[k_D]{\exvj}\pder[\exvj]{\thetad}\right]
        +2\mathbb{E}_\zbm[k_D(\ybm,\vj)]\mathbb{E}_\zbm\left[\pder[k_D]{\eyvj}\pder[\eyvj]{\thetad}\right]  \Bigg) \nonumber \\
        &=&\frac{1}{J}\sum_{j=1}^{J} \Bigg( 2\left(\mathbb{E}_\xbm[k_D(\xbm,\vj)]-\mathbb{E}_\zbm[k_D(\ybm, \vj)]\right)\mathbb{E}_\xbm\left[\pder[k_D]{\exvj}\pder[\exvj]{\thetad}\right] \nonumber \\ 
        && \qquad\qquad + 2\left(\mathbb{E}_\zbm[k_D(\ybm,\vj)]-\mathbb{E}_\xbm[k_D(\xbm, \vj)]\right)\mathbb{E}_\xbm\left[\pder[k_D]{\eyvj}\pder[\eyvj]{\thetad}\right]\Bigg) \nonumber \\
        &=&\ 0,
\end{eqnarray}
where we used the fact that  $\expectation_\zbm[k_D(\ybm,\vj)]-\expectation_\xbm[k_D(\xbm, \vj)]$ vanishes at the equilibrium.

Finally, the dynamics of the witness points can be written as
\begin{eqnarray}
    \dot{\bm{v}}_j &=&
        \frac{1}{J}\sum_{j=1}^{J}\Bigg(2\mathbb{E}_\xbm[k_D(\xbm,\vj)]\mathbb{E}_\xbm\left[\pder[k_D]{\exvj}\pder[\exvj]{\vj}\right]
        -2\mathbb{E}_\xbm[k_D(\xbm, \vj)]\mathbb{E}_\zbm\left[\pder[k_D]{\eyvj}\pder[\eyvj]{\vj}\right]
        \label{eq:appendix_dynamics_d_firstline}\\
        &&\qquad \qquad -2\mathbb{E}_\zbm[k_D(\ybm,\vj)]\mathbb{E}_\xbm\left[\pder[k_D]{\exvj}\pder[\exvj]{\vj}\right] 
        + 2\mathbb{E}_\zbm[k_D(\ybm,\vj)]\mathbb{E}_\zbm\left[\pder[k_D]{\eyvj}\pder[\eyvj]{\vj}\right] \nonumber \Bigg)\\
        &=&\frac{1}{J}\sum_{j=1}^{J} \Bigg( 2\left(\mathbb{E}_\xbm[k_D(\xbm,\vj)]-\mathbb{E}_\zbm[k_D(\ybm, \vj)]\right)\mathbb{E}_\xbm\left[\pder[k_D]{\exvj}\pder[\exvj]{\vj}\right] \\
        && \qquad\qquad + 2\left(\mathbb{E}_\ybm[k_D(\ybm,\vj)]-\mathbb{E}_\xbm[k_D(\xbm, \vj)]\right)\mathbb{E}_\xbm\left[\pder[k_D]{\eyvj}\pder[\eyvj]{\vj}\right] \Bigg)\label{eq:appendix_dynamics_d_secondline}\\
        &=&\ 0,
\end{eqnarray}
\noindent where we used the same argument as above to conclude that $\dot{\bm{v}}_j$ vanishes at the equilibrium.

Above analysis show that $(\dot{\bm{\theta}}_g, \dot{\bm{\theta}}_g, \Vset)$ vanishes at an equilibrium where  $P_{\Xb}=Q_{\Yb=G_{\thetagopt}(\Zb)}$. Notice that the equilibrium is not an isolated point in the space of parameters. In fact, $\Ecal=\{(\thetagopt,\thetad, \vbm)\ \text{for all}\ \thetad\in\Theta_D\}$ is the set of equilibria that is an equilibrium subspace $\mathcal{E}\subset\Theta_G\times\Theta_D\times\mathcal{V}^J$ instead of an isolated equilibrium point. In other words, we have continuum of equilibria rather than isolated equilibria.
\end{proof}

\subsection{Local Stability}
\label{sec_appendix:local_stability_proof}

Given that a desirable equilibrium exists as shown in Proposition \ref{sec_appendix:equilibrium_existence_proof}, we are now in a position to show the local stability of this equilibrium subspace. 

\begin{lemma_s}[\cite{nagarajan2017gradient,wang2018improving}]\label{lemma:exp_stable}
    Consider a non-linear system of parameters \((\bm{\theta}, \bm{\gamma})\): \(\dot{\bm{\theta}}=h_1(\bm{\theta}, \bm{\gamma})\), \(\dot{\bm{\gamma}}=h_2(\bm{\theta}, \bm{\gamma})\) with an equilibrium point at \((\bm{0},\bm{0})\). Let there exist \(\epsilon\) such that \(\forall\gamma\in\set{B}_\epsilon(\bm{0})\), \((\bm{0},\bm{\gamma})\) is an equilibrium. If \(\bm{J}=\frac{\partial h_1(\bm{\theta}, \bm{\gamma})}{\partial\bm{\theta}}\big\rvert_{(\bm{0},\bm{0})}\) is a Hurwitz matrix, the non-linear system is exponentially stable.
\end{lemma_s}

\begin{proposition_s}
The non-linear dynamical system with states $\bm{\Phi}=(\thetagen,\thetad, \Vset)$ is exponentially stable at its equilibrium $(\bm{\theta}^*_G,\thetad, \Vset)$ for any $\thetad\in\Theta_d$ and $\Vset\in\Vcal^J$.
\end{proposition_s}
\begin{proof}

We start off with deriving the Jacobian of the system~\eqref{eq_appendix:dynamical_system} as
\begin{equation}
    \bm{J}
    \triangleq
    \begin{bmatrix}
    \partial\dt{\thetagen}^T/\partial\thetagen & \partial\dt{\thetagen}^T/\partial\thetad & \partial\dt{\thetad}^T/\partial\bm{\Vset}\\ \partial\dt{\thetad}^T/\partial\thetagen & \partial\dt{\thetad}^T/\partial\thetad & \partial\dt{\thetad}^T/\partial{\Vset}\\
    \partial\dt{\Vset}^T/\partial\thetagen & \partial\dt{\Vset}^T/\partial\thetad & \partial\dt{\Vset}^T/\partial\Vset
    \end{bmatrix}.
\end{equation}

For simplicity, we assume that the effect of the feature extractor is absorbed in the kernel and $k_D(\xbm,\ybm)=k(D(\xbm),D(\ybm))=k(\xbm,\ybm)=k_\xybm$ and the kernel is radial $k(\xbm,\ybm)=k(\lVert \xbm-\ybm \rVert)=k(\xybm)$ where $\xybm$ is a scalar. Notice that $\xybm$ is not a multiplication. It only denotes a scalar which is a function of both $\xbm$ and $\ybm$. This simplifies the derivatives to a great extent.
To further simplify the notations of partial derivatives, let
\(\Delta_{\bm{b}}^{\bm{a}}=\frac{\partial^2\bm{a}}{\partial\bm{b}^2}\) and
\(\Delta_{\bm{b}\bm{c}}^{\bm{a}}=\frac{\partial^2\bm{a}}{\partial\bm{b}\partial\bm{c}}\).
In addition, we only consider a single witness point ($\vbm$), i.e., $J=1$. The
loss function is then simplified as 
\begin{equation}
    \Lcal^{\text{ume}}=\mathbb{E}^2_{\xbm}[k(\xvbm)]+\mathbb{E}^2_{\zbm}[k(\yvbm)]-2\mathbb{E}_{\xbm}[k(\xvbm)]\mathbb{E}_{\zbm}[k(\yvbm)] .
\end{equation}
Assume the only trainable parameters are the parameters of the generator $\thetabm=\bm{\theta}_g$ and the witness point $\vbm$. The dynamical system is then simplified as 
\begin{align}
    \dot{\thetabm}&=-
    2\mathbb{E}_\zbm[k(\yvbm)]\mathbb{E}_\zbm[k'_\yvbm\nabla^\yvbm_\thetabm]+2\mathbb{E}_\xbm[k(\xvbm)]\mathbb{E}_\zbm[k'_\yvbm\nabla^\yvbm_\thetabm]\\
    \dot{\vbm}&=
    2\mathbb{E}_\xbm[k(\xvbm)]\mathbb{E}_\xbm[k'_\xvbm\nabla^\xvbm_\vbm]
    +2\mathbb{E}_\zbm[k(\yvbm)]\mathbb{E}_\zbm[k'_\yvbm\nabla^\yvbm_\vbm]
    -2\mathbb{E}_\xbm[k(\xvbm)]\mathbb{E}_\zbm[k'_\yvbm\nabla^\yvbm_\vbm].
\end{align}

The Jacobian matrix 
\begin{equation}
    \bm{J}
    \triangleq
    \begin{bmatrix}
    \Jb_{GG} & \Jb_{GV}\\
    \Jb_{VG} & \Jb_{VV}
    \end{bmatrix}
    =\begin{bmatrix}
    \partial\dt{\thetabm}^T/\partial\thetabm & \partial\dt{\thetabm}^T/\partial\vbm\\
    \partial\dt{\vbm}^T/\partial\thetabm & \partial\dt{\vbm}^T/\partial\vbm\\
    \end{bmatrix}
\end{equation}
then describes the behaviour of the system around the equilibrium point. Moreover, we have
\begin{align}
    \frac{1}{2}\Jb_{GG} =& 
    -\mathbb{E}_{\zbm}[k_\yvbm^{\prime^2}\Delta_\thetabm^\yvbm]
    \mathbb{E}_{\zbm}[k_\yvbm^{\prime^2}\Delta_\thetabm^{\yvbm\tran}]
    +\mathbb{E}_{\zbm}[k(\yvbm)]\mathbb{E}_{\zbm}[k''_\yvbm\Delta_\thetabm^\yvbm+k'_\yvbm\Delta_\thetabm^\yvbm]\\
    &-\mathbb{E}_{\xbm}[k(\xvbm)]\mathbb{E}_{\zbm}[k''_\yvbm\Delta_\thetabm^\yvbm+k'_\yvbm\Delta_\thetabm^\yvbm]\nonumber\\
    =& -\mathbb{E}_{\zbm}[k_\yvbm^{\prime^2}\Delta_\thetabm^\yvbm]\mathbb{E}_{\zbm}[k_\yvbm^{\prime^2}\Delta_\thetabm^{\yvbm\tran}]+(\mathbb{E}_{\zbm}[k(\yvbm)] - \mathbb{E}_{\xbm}[k(\xvbm)])
    \mathbb{E}_{\zbm}[k''_\yvbm\Delta_\thetabm^\yvbm+k'_\yvbm\Delta_\thetabm^\yvbm]\label{eq_appendix:Jacobian_secondline}\\
    =&  -\mathbb{E}_{\zbm}[k_\yvbm^{\prime^2}\Delta_\thetabm^\yvbm]\mathbb{E}_{\zbm}[k_\yvbm^{\prime^2}\Delta_\thetabm^{\yvbm\tran}] \preceq 0.
\end{align}
Since $(\mathbb{E}_{\zbm}[k(\yvbm)] - \mathbb{E}_{\xbm}[k(\xvbm)]) = 0$ at the equilibrium  when $P_\Xb=Q_\Yb$ (See Assumption \ref{assumption_appendix:equilibrium}), the second term of~\eqref{eq_appendix:Jacobian_secondline} vanishes at the equilibrium. The last line follows from the fact that $\mathbf{M}\mathbf{M\tran}$ is positive semidefinite for every matrix $\mathbf{M} \in\set{R}^{m\times n}$ \citep{strang1993introduction}. Having proved that $\Jb_{GG}$ is negative definite makes it straightforward, inspired by~\cite{nagarajan2017gradient}(Lemma C.3), to take the last step and show the local exponential stability of the system. The idea is to expand the eigenvalue decomposition of $\Jb_{GG}$ and projecting the system to the subspace that is orthogonal to the equilibria. The resultant projected system has a Hurwitz Jacobian matrix ensuring the exponential stability of the system in the subspace orthogonal to the equilibria. The result then follows from Lemma \ref{lemma:exp_stable} (See the last step of the proof of the Proposition 1 of \citep{wang2018improving}).
\end{proof}

\section{Derivation of the Analytical Dynamics Function}
Here we derive the formulation of the dynamical system of~\eqref{eq:dynamical_system} for simple tractable cases which are used to simulate the dynamical systems of~\Cref{sec:experiments_simulation}.
We first reproduce a well-known lemma that will be useful for derivations to come. 
\begin{lemma_s}[Gaussian integral computation (\cite{GarJitKan2017}, Lemma E.1)]
\label{lem:gauss_integral} Let $a,b,c,d\in\mathbb{R}$ with $b,d>0$.
Then
\begin{align*}
\frac{1}{\sqrt{2\pi}}\int\exp\left(\frac{-(x-a)^{2}}{b}+\frac{-(x-c)^{2}}{d}\right)\thinspace\mathrm{d}x & =\sqrt{\frac{bd}{2(b+d)}}\exp\left(\frac{-(a-c)^{2}}{b+d}\right).
\end{align*}
\end{lemma_s}

\subsection{Single Gaussian}
\label{subsec_appendix:ume_pq_single_gaussian}
Assume
$P=\mathcal{N}(m_{p},\sigma_{p}^{2}),Q=\mathcal{N}(m_{q},\sigma_{q}^{2})$ and
$k(x,y) = \exp\left( -\frac{ (x-y)^2}{2\sigma^2}\right)$ be a Gaussian kernel with bandwidth $\sigma^{2}$.
\begin{align}
\mu_{P}(v) & =\int k(x,v)p(x)\mathrm{d}x\nonumber \\
 & =\frac{1}{\sigma_{p}}\frac{1}{\sqrt{2\pi}}\int\exp\left(-\frac{(x-v)^{2}}{2\sigma^{2}}-\frac{(x-m_{p})^{2}}{2\sigma_{p}^{2}}\right)\mathrm{d}x\nonumber \\
 & \stackrel{(a)}{=} \sqrt{\frac{\sigma^{2}}{\sigma^{2}+\sigma_{p}^{2}}}\exp\left(-\frac{(v-m_{p})^{2}}{2(\sigma^{2}+\sigma_{p}^{2})}\right), \label{memb_gaussian}
\end{align}
where at $(a)$ we use Lemma \ref{lem:gauss_integral}.
It follows that 
\begin{align*}
 & (\mu_{P}(v)-\mu_{\theta}(v))^{2}\\
 & =\mu_{P}^{2}(v)+\mu_{Q}^{2}(v)-2\mu_{P}(v)\mu_{Q}(v)\\
 & =\frac{\sigma^{2}}{\sigma^{2}+\sigma_{p}^{2}}\exp\left(-\frac{(v-m_{p})^{2}}{\sigma^{2}+\sigma_{p}^{2}}\right)+\frac{\sigma^{2}}{\sigma^{2}+\sigma_{q}^{2}}\exp\left(-\frac{(v-m_{q})^{2}}{\sigma^{2}+\sigma_{q}^{2}}\right)\\
 & \phantom{=}-2\sqrt{\frac{\sigma^{2}}{\sigma^{2}+\sigma_{p}^{2}}\frac{\sigma^{2}}{\sigma^{2}+\sigma_{q}^{2}}}\exp\left(-\frac{(v-m_{p})^{2}}{2(\sigma^{2}+\sigma_{p}^{2})}-\frac{(v-m_{q})^{2}}{2(\sigma^{2}+\sigma_{q}^{2})}\right).
\end{align*}

After further simplification $p=\mathcal{N}(0,1),q=\mathcal{N}(m_{q},1)$ and $\theta=m_{q}$, we have
\begin{align*}
(\mu_{P}(v)-\mu_{\theta}(v))^{2} & =\frac{\sigma^{2}}{\sigma^{2}+1}\exp\left(-\frac{v^{2}}{\sigma^{2}+1}\right)+\frac{\sigma^{2}}{\sigma^{2}+1}\exp\left(-\frac{(v-m_{q})^{2}}{\sigma^{2}+1}\right)\\
 & \phantom{=}-2\frac{\sigma^{2}}{\sigma^{2}+1}\exp\left(-\frac{v^{2}}{2(\sigma^{2}+1)}-\frac{(v-m_{q})^{2}}{2(\sigma^{2}+1)}\right).
\end{align*}
giving rise to the following gradient vector field
\begin{align*}
\frac{dv}{dt}=\nabla_{v}(\mu_{P}(v)-\mu_{\theta}(v))^{2} & =-\frac{2\sigma^{2}}{\left(\sigma^{2}+1\right)^{2}}v\exp\left(-\frac{v^{2}}{\sigma^{2}+1}\right)\\
& -\frac{2\sigma^{2}}{\left(\sigma^{2}+1\right)^{2}}(v-m_{q})\exp\left(-\frac{(v-m_{q})^{2}}{\sigma^{2}+1}\right)\\
 & \phantom{=}+\frac{2\sigma^{2}}{(\sigma^{2}+1)^{2}}(2v-m_{q})\exp\left(-\frac{v^{2}+(v-m_{q})^{2}}{2(\sigma^{2}+1)}\right).
\end{align*}
Similarly, we have 
\begin{align*}
\frac{dm_q}{dt}=-\nabla_{m_{q}}(\mu_{P}(v)-\mu_{\theta}(v))^{2} & =-\frac{2\sigma^{2}}{\left(\sigma^{2}+1\right)^{2}}(v-m_{q})\exp\left(-\frac{(v-m_{q})^{2}}{\sigma^{2}+1}\right)\\
 & \phantom{=}+\frac{2\sigma^{2}}{(\sigma^{2}+1)^{2}}(v-m_{q})\exp\left(-\frac{v^{2}+(v-m_{q})^{2}}{2(\sigma^{2}+1)}\right)\\
 & =-\frac{2\sigma^{2}}{\left(\sigma^{2}+1\right)^{2}}(v-m_{q})\\&\left[\exp\left(-\frac{(v-m_{q})^{2}}{\sigma^{2}+1}\right)-\exp\left(-\frac{v^{2}+(v-m_{q})^{2}}{2(\sigma^{2}+1)}\right)\right].
\end{align*}

\subsection{Spiky Gaussian}
\label{sec:spiky_gauss_details}

Let $P_1\coloneqq \mathcal{N}(0,1)$, $P_2\coloneqq \mathcal{N}(0,\sigma_{q}^{2})$ and $P\coloneqq wP_1+(1-w)P_{2}$ defined on $\mathbb{R}$
for some weight $w\in[0,1]$ and variance $\sigma_{q}^{2}$.
When $\sigma_{q}^{2}$ is small, the two distributions illustrate
a case where the primary difference is local (at the origin). Consider a moment when the trainable distribution (model) is $Q=\mathcal{N}(0,1)$. We refer
to $P$ as a spiky Gaussian. Assume a Gaussian kernel $k(x,y)=\exp\left(-\frac{(x-y)^{2}}{2\sigma^{2}}\right)$
for some bandwidth $\sigma^{2}>0$. In the following we derive the loss functions of both MMD and UME.

\paragraph{MMD:}

We have 
\begin{align*}
\mathrm{MMD}^{2}(P,Q) & =\mathbb{E}_{x,x'\sim P}k(x,x')+\mathbb{E}_{y,y'\sim Q}k(y,y')-2\mathbb{E}_{x\sim P}\mathbb{E}_{y\sim Q}k(x,y)\\
 & \stackrel{(a)}{=}(1-w)^{2}\left[c(2)+c(2\sigma_{q}^{2})-2c(1+\sigma_{q}^{2})\right],
\end{align*}
where we note that all the expectations in the first line are Gaussian
integrals and use Lemma \ref{lem:gauss_integral} to simplify all
the expressions at $(a)$, and $c(z)\coloneqq \sqrt{\frac{\sigma^{2}}{\sigma^{2}+z}}$
for $z>0$. Note that when $w=1$ or $\sigma_{q}^{2}=1$, $P=Q$ and
$\mathrm{MMD}^{2}(P,Q)=0$.

\paragraph{UME:}
With the help of Lemma \ref{lem:gauss_integral}, we have similarly
\begin{align*}
\mu_{Q}(v) & =\mathbb{E}_{y\sim Q}k(y,v)=c(1)\exp\left(\frac{-v^{2}}{2(\sigma^{2}+1)}\right),\\
\mu_{P}(v) & =w\mathbb{E}_{x\sim Q}k(x,v)+(1-w)\mathbb{E}_{x\sim P_{2}}k(x,v)\\
 & =w\mu_{Q}(v)+(1-w)c(\sigma_{q}^{2})\exp\left(\frac{-v^{2}}{2(\sigma^{2}+\sigma_{q}^{2})}\right).
\end{align*}
It follows that
\begin{align*}
\mathrm{UME}^{2}(P,Q) & =(\mu_{P}(v)-\mu_{Q}(v))^{2}\\
 & =(1-w)^{2}\left[c(1)\exp\left(\frac{-v^{2}}{2(\sigma^{2}+1)}\right)-c(\sigma_{q}^{2})\exp\left(\frac{-v^{2}}{2(\sigma^{2}+\sigma_{q}^{2})}\right)\right]^{2}.
\end{align*}

The derivatives of these loss functions with respect to $\sigma_q$ are taken for the sensitivity analysis.
\begin{align}
\label{eq_appendix:sensitivity}
    \pder[\mathrm{(MMD^2)}]{\sigma_q} &= \frac{\sigma_q \sigma^2}{2\sqrt{\frac{\sigma^2}{\sigma^2+\sigma_q^2+1}}(\sigma^2+\sigma_q^2+1)^2}-\frac{\sigma^2}{4(\sigma^2+2\sigma_q)^2\sqrt{\frac{\sigma^2}{\sigma^2+2\sigma_q}}}\\
    \pder[\mathrm{(UME^2)}]{\sigma_q}&=  -2(1-w)^2\left(\frac{\exp(-\frac{v^2}{2\sigma^2+2})\sqrt{\frac{\sigma^2}{\sigma^2+1}}}{2}-\frac{\exp{(-\frac{v^2}{2\sigma^2+2\sigma_q^2}})\exp(\frac{\sigma^2}{\sigma^2+\sigma_q^2})}{2}\right)\\
    &\times\left(\frac{2\sigma_q v^2 \exp(-\frac{v^2}{2\sigma^2+2\sigma_q^2})\sqrt{\frac{\sigma^2}{\sigma^2+\sigma_q^2}}}{(2\sigma^2+2\sigma_q^2)^2}-
    \frac{\sigma^2 \sigma_q \exp(-\frac{v^2}{2\sigma^2+2\sigma_q^2})}{2(\sigma^2+\sigma_q^2)^2\sqrt{\frac{\sigma^2}{\sigma^2+\sigma_q^2}}}\nonumber
    \right).
\end{align}

\subsection{Mixture of Gaussians}

\label{subsec_appendix:ume_pq_mixtureofgaussians}
Assume that $p(x)\coloneqq \sum_{j=1}^{c_{p}}\omega_{p,j}\mathcal{N}(x\mid m_{p,j},\sigma_{p,j}^{2})\coloneqq \sum_{j=1}^{c_{p}}\omega_{p,j}p_{j}(x)$
where we define the j\textsuperscript{th} component $p_{j}(x)\coloneqq \mathcal{N}(x\mid m_{p,j},\sigma_{p,j}^{2})$,
$\boldsymbol{\omega}_{p}\coloneqq (\omega_{p,1},\ldots,\omega_{p,c_{p}})^{\top}$
is a vector of non-negative mixing weights such that $\sum_{j=1}^{c_{p}}\omega_{p,j}=1$,
$\{m_{p,j}\}_{j=1}^{c_{p}}$ are the means of the $c_{p}>0$ components,
and $\{\sigma_{p,j}^{2}\}_{j=1}^{c_{p}}$ are the variances. Similarly,
assume that $q(y)\coloneqq \sum_{j=1}^{c_{q}}\omega_{q,j}\mathcal{N}(y\mid m_{q,j},\sigma_{q,j}^{2})$.
It follows from~\eqref{memb_gaussian} that the mean embedding $\mu_{p}$
of $p$ w.r.t a Gaussian kernel with bandwidth $\sigma^{2}$ is 
\begin{align*}
\mu_{p}(v) & =\sum_{j=1}^{c_{p}}\omega_{p,j}\mu_{p_{j}}(v)\\
 & =\sum_{j=1}^{c_{p}}\omega_{p,j}\sqrt{\frac{\sigma^{2}}{\sigma^{2}+\sigma_{p,j}^{2}}}\exp\left(-\frac{(v-m_{p,j})^{2}}{2(\sigma^{2}+\sigma_{p,j}^{2})}\right).
\end{align*}


\section{UME Provides An Extra Signal to the Gradient}
\label{sec:ume_grad}
\textbf{Guiding the Generator to Capture a Missing Mode}
The added UME term to the main objective (see \eqref{eq:objective_kernel}) acts as a training
guide which directs the generator to capture the unknown data modes
as indicated by the witness points. The goal of this section is to
make this statement explicit with the following illustrative example.

Define $S=\mathcal{N}(0,1)$. Let the true data distribution be $P:=\omega P_{1}+(1-\omega)S$
where $P_{1}:=\mathcal{N}(m_{p},1)$ for some mean $m_{p}\neq0$ and
mixing proportion $\omega\in[0,1]$. Let the model be $Q:=\omega Q_{1}+(1-\omega)S$
where $Q_{1}:=\mathcal{N}(m_{q},1)$ and $m_{q}$ is the parameter
to learn. This problem illustrates a case where the data generating
distribution $P$ is bi-modal, and one of its two modes is already
captured by our model $Q$ , i.e., the second component $S$. We will
show that by placing a witness point $v$ in a high density region
of $P_{1}$, the gradient $-\nabla_{m_{q}}\mathrm{UME}^{2}$ leads
$m_{q}$ toward $m_{p}$.

Assume a Gaussian kernel $k(x,y)=\exp\left(-\frac{(x-y)^{2}}{2\sigma^{2}}\right)$
for some bandwidth $\sigma^{2}>0$. By \eqref{memb_gaussian}, we have
\begin{align*}
\mu_{P}(v) & =\omega\mathbb{E}_{x\sim P_{1}}k(x,v)+(1-\omega)\mathbb{E}_{x\sim\mathcal{N}(0,1)}k(x,v)\\
 & =\sqrt{\frac{\sigma^{2}}{\sigma^{2}+1}}\left[\omega\exp\left(-\frac{(v-m_{p})^{2}}{2(\sigma^{2}+1)}\right)+(1-\omega)\exp\left(-\frac{v^{2}}{2(\sigma^{2}+1)}\right)\right].
\end{align*}
Likewise, 
\begin{align*}
\mu_{Q}(v) & \text{=}\sqrt{\frac{\sigma^{2}}{\sigma^{2}+1}}\left[\omega\exp\left(-\frac{(v-m_{q})^{2}}{2(\sigma^{2}+1)}\right)+(1-\omega)\exp\left(-\frac{v^{2}}{2(\sigma^{2}+1)}\right)\right].
\end{align*}
It follows that 
\begin{align*}
\mathrm{UME}^{2}(P,Q) & =\left(\mu_{P}(v)-\mu_{Q}(v)\right)^{2}\\
 & =\frac{\sigma^{2}}{\sigma^{2}+1}\omega^{2}\left[\exp\left(-\frac{(v-m_{p})^{2}}{2(\sigma^{2}+1)}\right)-\exp\left(-\frac{(v-m_{q})^{2}}{2(\sigma^{2}+1)}\right)\right]^{2},
\end{align*}
where we use one witness point $v$ for $\mathrm{UME}^{2}(P,Q)=\mathrm{UME}^{2}$.
To make the dependency on $v$ explicit, we write $\mathrm{UME}_{v}^{2}$.
To provide an explicit training guide which directs $Q$ to capture
the mode (i.e., $P_{1}$), we set $v=m_{p}$, giving
\begin{align*}
\mathrm{UME}_{v=m_{p}}^{2} & =\frac{\sigma^{2}}{\sigma^{2}+1}\omega^{2}\left[1-\exp\left(-\frac{(m_{p}-m_{q})^{2}}{2(\sigma^{2}+1)}\right)\right]^{2}.
\end{align*}
The parameter $m_{q}$ is updated with a first-order optimization
algorithm which relies on the gradient
\begin{align*}
-\nabla_{m_{q}}\mathrm{UME}_{v=m_{p}}^{2} & =-\frac{2\sigma^{2}}{\sigma^{2}+1}\omega^{2}\left[1-e^{\left(-\frac{(m_{p}-m_{q})^{2}}{2(\sigma^{2}+1)}\right)}\right]e^{\left(-\frac{(m_{p}-m_{q})^{2}}{2(\sigma^{2}+1)}\right)}\frac{m_{p}-m_{q}}{\sigma^{2}+1}.
\end{align*}

To illustrate, we assume that the true parameter $m_{p}=1$ and $\omega=1/2$.
The gradient $-\nabla_{m_{q}}\mathrm{UME}_{v=m_{p}}^{2}$ evaluated
at various values of $m_{q}$ is shown in Figure \ref{fig:mo2g_grad_ume}.

We observe that at $m_{q}=m_{p}$ (i.e., $P=Q$), the gradient is
zero, providing no signal to change $m_{q}$ further. When $m_{q}>m_{p}$,
the gradient is negative (pulling back), which will decrease the value
of $m_{q}$ toward $m_{p}$. Likewise, when $m_{q}<m_{p}$, the gradient
is positive (pushing forward), leading to a positive change of $m_{q}$
toward $m_{p}$. Since $\mathrm{UME}$ measures local differences
around the witness point $v$, the gradient signal is strong in the
neighborhood of $v$, and much weaker in other regions that are far
from $v$ (theoretically non-zero). This observation holds true for
any kernel bandwidth $\sigma^{2}>0$. It can be seen that a higher
value of the kernel bandwidth makes UME less local (i.e., larger coverage
with dispersed gradient signal) at the expense of lower magnitudes
(pointwise).
\section{Other Interpretations}
\label{sec:appendix_other_interpretations}
Witness points can be theoretically interpreted in different ways. Two such interpretations are adaptive filters and change of coordinates as will be detailed in the following.

\subsection{Interpretation as an Adaptive Filter}
\label{sec:interpretations}

To elaborate more on the trade-off between the global and local terms in~\Cref{ex:mmd-ume} in the main text, we consider the shift-invariant kernel $k(\xbm,\xbm') = \psi(\xbm-\xbm')$ where $\xbm,\xbm'\in \mathbb{R}^{d_\xbm}$ and $\psi$ is a positive definite function. Let $\varphi_{P_\Xb}$ and $\varphi_{Q_\Yb}$ be characteristic functions of $P_\Xb$ and $Q_\Yb$, respectively. Using Bochner's theorem~\citep{loomis2013introduction} and the Euler's formula, we can rewrite the objective in \eqref{ex:mmd-ume} in the Fourier domain as
\begin{align*}
    & \int_{\mathbb{R}^{d_\xbm}}\left|\varphi_{P_\Xb}(\omegabm) - \varphi_{Q_\Yb}(\omegabm)\right|^2\,d\Lambda(\omegabm) \nonumber 
    +\frac{\lambda}{J}\sum_{j=1}^J\left(\int_{\mathbb{R}^d} [\varphi_{P_\Xb}(\omegabm) - \varphi_{Q_\Yb}(\omegabm)]\cos\left(\omegabm^\top \vbm_j\right) \,d\Lambda(\omegabm)\right)^2,
\end{align*}
where $\Lambda$ is the inverse Fourier transform of the kernel $k$. As we can
see, $\vjs$ appear as the frequency modulators. The summands of the second term
are most sensitive to frequencies which are determined by witness points
$\vjs$. Since these witness points are trainable, they determine frequencies to
be learned at each iteration. For example, in the case of natural images, the
algorithm has the freedom to first learn the low frequency (long-range)
structure of the image and then shift its attention to learning the high
frequency fine details. This transition is adaptively controlled by witness
points $\vjs$. An experiment will be presented to empirically show this
interpretation in Section~\ref{sec:experiments}
and~\Cref{subfig:sens_a,subfig:sens_b}.


\subsection{Interpretation as Change of Coordinates}
\label{sec_appendix:chnage_of_coordinates_interpretation}

Let $\mu_{P}\coloneqq \mathbb{E}_{\boldsymbol{x}\sim P}[\phi(\boldsymbol{x})]$
be the mean embedding of $P$ in a Hilbert space $\mathcal{H}$. The
UME between $P$ and $Q$ is defined as
\begin{align*}
\mathrm{UME}^{2}(P_\Xb,Q_\Yb) & \coloneqq \frac{1}{J}\sum_{j=1}^{J}(\mu_{P}(\boldsymbol{v}_{j})-\mu_{Q}(\boldsymbol{v}_{j}))^{2}\\
 & =\left\langle (\mu_{P}-\mu_{Q}),\left(\frac{1}{J}\sum_{j=1}^{J}\phi(\boldsymbol{v}_{j})\otimes\phi(\boldsymbol{v}_{j})\right)(\mu_{P}-\mu_{Q})\right\rangle _{\mathcal{H}},
\end{align*}
where $\otimes$ denotes the outer product, and $\left\langle \cdot,\cdot\right\rangle _{\mathcal{H}}$
denotes the inner product on $\mathcal{H}$. Our regularized objective
is 
\begin{align*}
\mathrm{MMD}^{2}(P,Q)+\lambda\mathrm{UME}^{2}(P,Q) &
=\|\mu_{P}-\mu_{Q}\|^{2}+\lambda\frac{1}{J}\sum_{j=1}^{J}(\mu_{P}(\boldsymbol{v}_{j})-\mu_{Q}(\boldsymbol{v}_{j}))^{2}\\
 & =\left\langle
 (\mu_{P}-\mu_{Q}),\left(I+\frac{\lambda}{J}\sum_{j=1}^{J}\phi(\boldsymbol{v}_{j})\otimes\phi(\boldsymbol{v}_{j})\right)(\mu_{P}-\mu_{Q})\right\rangle
 _{\mathcal{H}},
\end{align*}
where $I\colon\mathcal{H}\to\mathcal{H}$ is the identity operator.

Consider $\phi(\boldsymbol{x})=\boldsymbol{x}$ i.e., a linear kernel.
Then the regularized objective becomes
\begin{align*}
(\mu_{P}-\mu_{Q})^{\top}\left(I+\frac{\lambda}{J}
\sum_{j=1}^{J}\boldsymbol{v}_{j}\boldsymbol{v}_{j}^{\top}\right)(\mu_{P}-\mu_{Q}),
\end{align*}
where $\mu_{P}$ is simply the first moment of $P$.

If we assume orthonormal witness points $\vjs$, we can then view $A = \sum_{j=1}^{J}\boldsymbol{v}_{j}\boldsymbol{v}_{j}^{\top}$ as the eigenvalue decomposition of $A$. This basically means that the term $\frac{\lambda}{J}\sum_{j=1}^{J}\boldsymbol{v}_{j}\boldsymbol{v}_{j}^{\top}$ performs a change of coordinates by a rotation matrix whose eigenvectors are $\vbm_j$. During the optimization of the following loss function
\begin{equation*}
    \Lcal=(\mu_{P}-\mu_{Q})^{T}\left(I+\frac{\lambda}{J}\sum_{j=1}^{J}\boldsymbol{v}_{j}\boldsymbol{v}_{j}^{\top}\right)(\mu_{P}-\mu_{Q}),
\end{equation*}
with respect to the generator, we have to compute 
\pder[\Lcal]{(\mu_{P}-\mu_{Q})} that becomes 
\begin{equation*}
    \pder[\Lcal]{(\mu_{P}-\mu_{Q})}=(I+2A)(\mu_{P}-\mu_{Q}),
\end{equation*}
instead of
\begin{equation*}
    \pder[\Lcal]{(\mu_{P}-\mu_{Q})}=I(\mu_{P}-\mu_{Q}),
\end{equation*}
meaning that the gradients are scaled by $2A$. Since $\vjs$ are trainable, we can say that at each iteration of the generator, the gradients are altered by the matrix $A$ whose eigenvectors are the directions in which $P$ and $Q$ are most different.

\section{More on the Issue of Vanishing Gradient}
\label{sec_appendix:vanishing_grad}

In this section, we elaborate more on the general idea of introducing a third component to the training process of generative adverarial networks in addition to generator and discriminator and locate~\ours in this line of works. For example in \citep{sajjadi2018tempered}, a third component called~\emph{lens} is added to balance the training dynamics and prevent mode collapse. They proposed to pass the data distribution through an autoencoder and use the reconstructed data as the target distribution for a generator. This might lead to instability in higher dimensions according to the following reasoning. 

It has been proved by~\cite[Theorems 2.1 and 2.2]{arjovsky2017towards} that if $P_\Xb$ (real data) and $Q_\Yb$ (fake data) have supports contained within two disjoint compact subsets, there is a smooth optimal discriminator that has perfect accuracy on the union of their supports. This results in a vanishing gradient issue for the generator that leads to instability ~\citep[Theorems 2.4 and 2.6]{arjovsky2017towards} because of two reasons: 
\begin{itemize}
    \item The data distribution naturally lives on a low-dimensional manifold.
    \item The generated distribution is the result of applying function $G:\set{R}^{d_\zbm}\to\set{R}^{d_\xbm}$ on random noise $\zbm\in\Zcal$.
\end{itemize}

\cite[Lemma 1]{arjovsky2017towards} also proved that the $\dimension(\mathrm{Img(G)})\leq \dimension(\Zcal)$ where $\dimension(.)$ denotes the manifold dimension. Now assume an autoencoder function with latent dimension $l$ as $f_d(f_e(.))$ where $f_e:\set{R}^{d_\xbm}\to\set{R}^{d_l}$ and $f_d:\set{R}^{d_l}\to\set{R}^{d_\xbm}$. Based on \citep[Lemma 1]{arjovsky2017towards}, $\dimension(\mathrm{Img}(f_d \circ f_e))\leq l$ where we normally have $l\ll d_\xbm$ in autoencoders. This makes the real and fake data manifolds even more distinguishable and exacerbates the vanishing gradient problem. Simply speaking, lower-dimensional manifolds are more easily distinguishable in a high-dimensional embedding space. As a mental experiment, in a $3$-dimensional embedding space, an imaginary discriminator has easier job when it discriminates two lines($1$D manifold) than when it discriminates two planes($2$D manifold).


\section{Experiments}
\label{sec_appendix:experiments}
In this section, we provide details for the experiments presented in the main text and also more results from the experiments that was eliminated from the main text due to the shortage of space.

\subsection{Experiment Details}
The experiment details for the $2$-dimensional example of Section.~\ref{sec:synthetic} is as follows:
\begin{itemize}
    \item Generator architecture: [Linear($512$):Tanh:Linear($2048$):Tanh:Linear($512$)\\:Tanh:Linear($256$):Tanh:Linear($256$)]
    with $10$-dimensional input noise dimension.
    \item Kernel: Gaussian kernel with trainable bandwidth.
    \item Batchsize: $64$
    \item Trade-off weight between MMD and UME term ($\lambda$): $0.1$
    \item Learning rate for the generator, and witness points: $0.001$
    \item The number of iterations each component is trained ($n_g$, $n_v$): $1$
    \item Number of witness points: $20$
    \item Convergence criterion: The magnitude of the gradients back propagated to witness points is less than a threshold ($1e-5$) or the number of epochs is less than $20$.
\end{itemize}

The experiment details for the MNIST and CIFAR10 examples are as follows
\begin{itemize}
    \item Generator architecture:
    
    [ConvTranspose2d(1024, 512, 4, 2,1):BatchNorm(512):ReLU\\
    :ConvTranspose2d(512, 256, 4, 2, 1):BatchNorm(256):ReLU\\
    :ConvTranspose2d(256, 128, 4, 2, 1):BatchNorm(128):ReLU\\
    :ConvTranspose2d(128, 1, 4, 2, 1):Tanh]
    with $100$-dimensional noise in the input.
    \item Autoencoder architecture:
    
    Encoder:[Conv2d(1, 64, 4, 2, 1):LeakyReLU(0.2)\\
    :Conv2d(64, 128, 4, 2, 1):BatchNorm(128):LeakyReLU(0.2)\\
    :Conv2d(128, 128, 4, 2, 1):BatchNorm(128):LeakyReLU(0.2,)]
    
    Decoder:[ConvTranspose2d(hidden_dim, 128, 4, 2, 1):BatchNorm(128):LeakyReLU(0.2)\\
    :ConvTranspose2d(128, 64, 4, 2, 1):BatchNorm(64):LeakyReLU(0.2)\\
    :ConvTranspose2d(64, 1, 4, 2, 1):BatchNorm(1):Tanh()]
    
    Notice that the first two arguments of Conv2d and ConvTranspose2d are the number of input and output layers. The last three arguments are kernel size, stride, and padding respectively.

    \item Batchsize: 64
    \item Trade-off weight between MMD and UME term, $\lambda=0.5$
    \item Number of witness points: $256$
    \item Learning rate for the generator, encoder, decoder and witness points: $0.001$
    \item The number of iterations each component is trained ($n_g, n_e, n_d, n_v$): $1$
    \item Kernel: inverse multi-quadratic $k(x,y) = (c^2 + ||x-y||^2)^b$ with $c=1.0$ and $b=-0.5$.
    \item Convergence criterion: The magnitude of the gradients back propagated to witness points is less than a threshold ($1e-5$) or the number of epochs is less than $20$.


The idea of this experiment was to see the effectiveness of witness points for guiding the training of the generator. To make our point clear, we designed a two-phase experiment. In the first phase, only one digit was presented to the generator. The generator architecture was borrowed from~\cite{li2015generative}. To keep things simple, rather than employing multiple kernels with different bandwidths similar to~\cite{li2015generative}, we used a single Gaussian kernel with a bandwidth that is heuristically set to $30$ chosen by the median criterion.
\end{itemize}

\begin{figure*}[t]
    \centering
    \begin{minipage}[t]{\textwidth}
        \centering
        \begin{subfigure}[c]{0.24\textwidth}
            \centering
            \includegraphics[width=\linewidth]{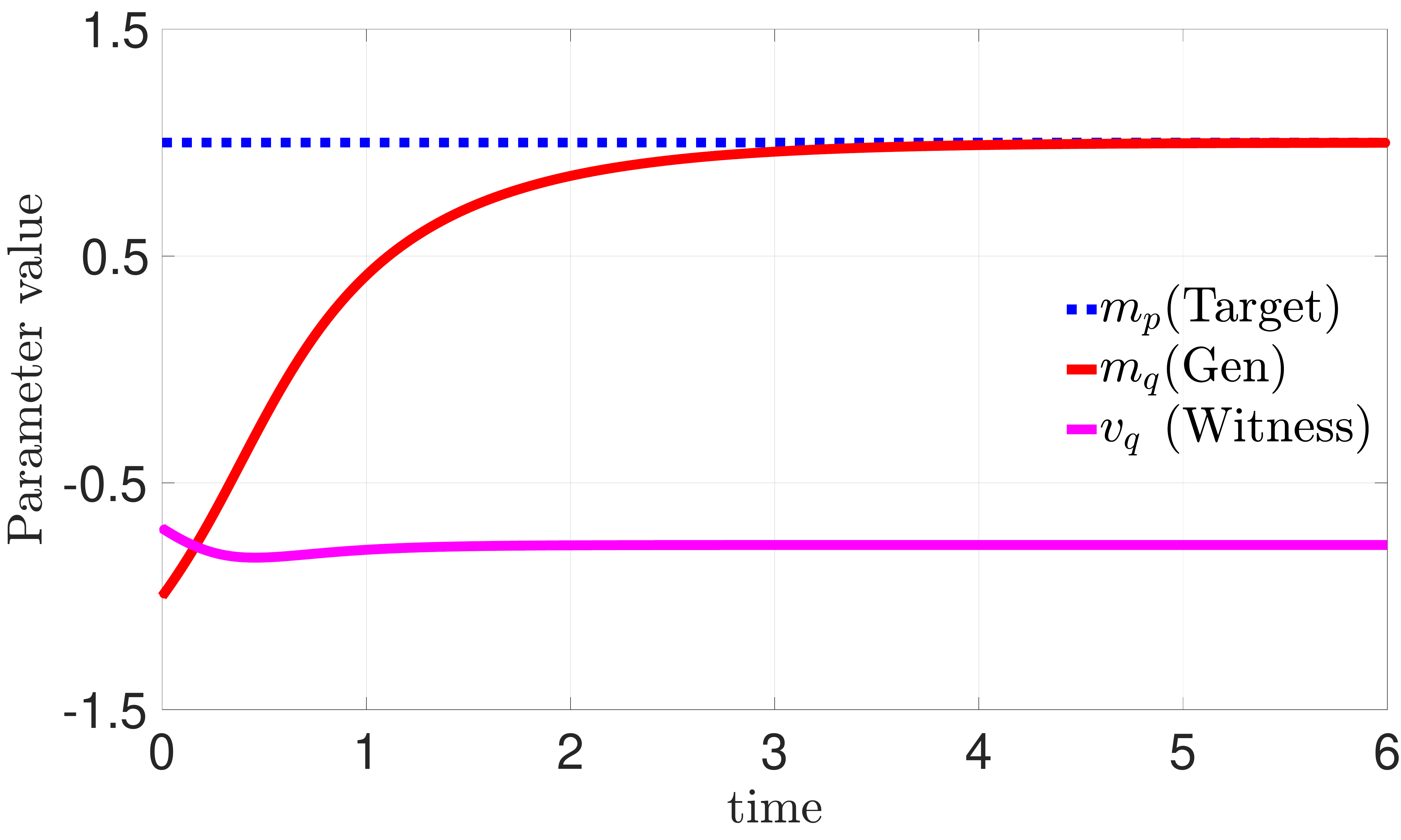}
            \caption{\scriptsize $\lambda=5$}
            \label{fig:appendix_single_gaussian_dynamics_lamda_5_a}
        \end{subfigure}
        \begin{subfigure}[c]{0.24\textwidth}
            \centering
            \includegraphics[width=\linewidth]{figs/low_dims/fixed_sigmas/sigma_2,00_lambda_5,00.pdf}
            \caption{\scriptsize $\lambda=\infty$}
            \label{fig:appendix_single_gaussian_dynamics_lamda_inf_a}
        \end{subfigure}
        \begin{subfigure}[c]{0.24\textwidth}
            \centering
            \includegraphics[width=\linewidth]{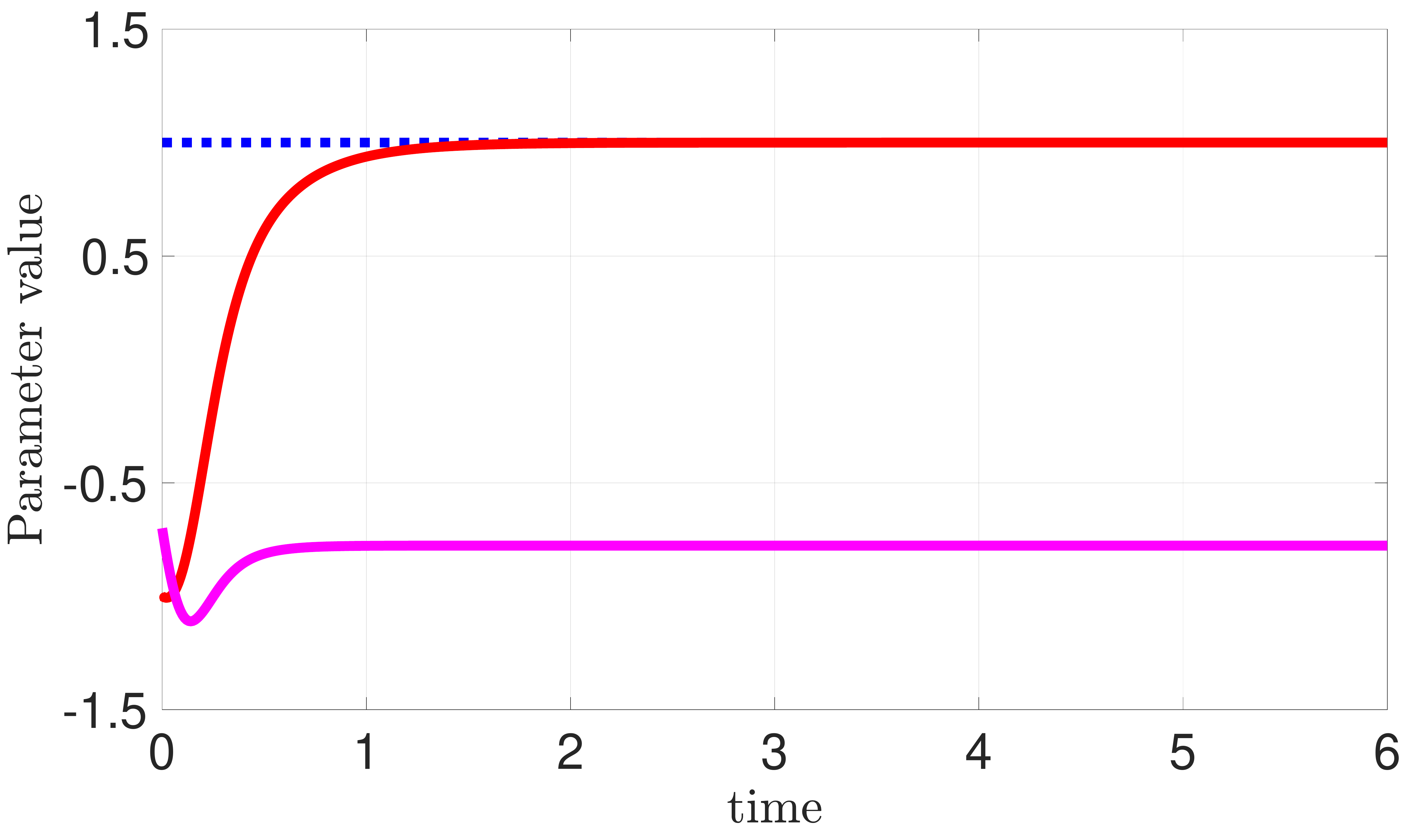}
            \caption{\scriptsize $\lambda=5$}
            \label{fig:appendix_single_gaussian_dynamics_lamda_5_b}
        \end{subfigure}
        \begin{subfigure}[c]{0.24\textwidth}
            \centering
            \includegraphics[width=\linewidth]{figs/low_dims/fixed_sigmas/sigma_2,00_lambda_Inf.pdf}
            \caption{\scriptsize $\lambda=\infty$}
            \label{fig:appendix_single_gaussian_dynamics_lamda_inf_b}
        \end{subfigure}
        \caption{Numerical solution of the differential equations governing the GD
    learning of the mean of a Gaussian distribution. Color codes: (dashed blue)
target value of the parameter, (red) learned value of the parameter, and (magenta)
witness point. $\lambda=\infty$ means that only the UME term exists.}
    \label{fig:appendix_low_dims_sim_fixed_sigmas}
    \end{minipage}
    \begin{minipage}[t]{\textwidth}
            \centering
                \begin{subfigure}[c]{0.25\textwidth}
                    \centering
                    \includegraphics[width=\linewidth]{./figs/low_dims/mog_1d/sigma_2,00_v0_0,40_lambda_5,00.pdf}
                    \caption{\scriptsize $\lambda=5$}
                    \label{subfig:appendix_mog_a}
                \end{subfigure}
                \begin{subfigure}[c]{0.25\textwidth}
                    \centering
                    \includegraphics[width=\linewidth]{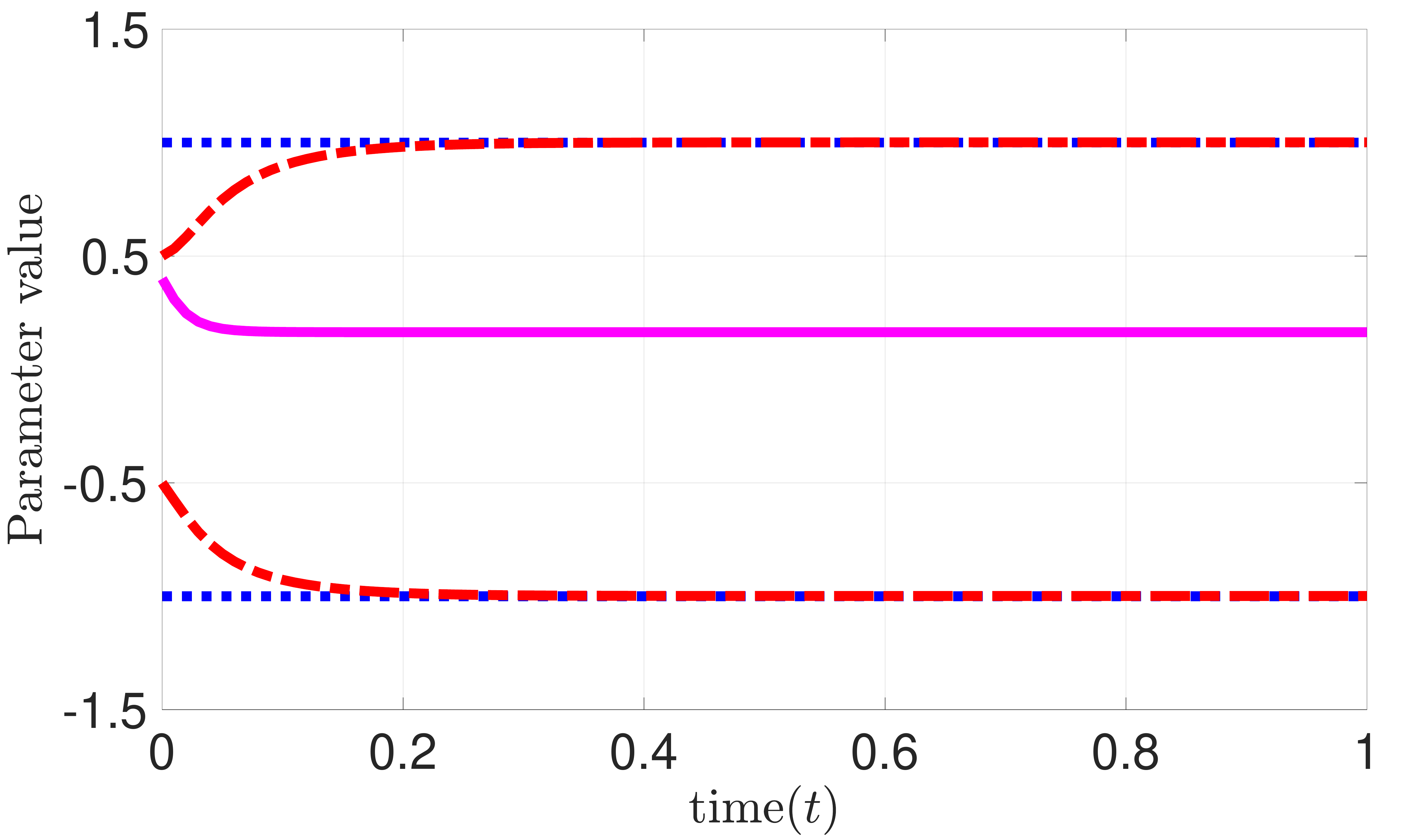}
                    \caption{\scriptsize $\lambda=\infty$}
                    \label{subfig:appendix_mog_b}
                \end{subfigure}
                \begin{subfigure}[c]{0.24\textwidth}
                    \centering
                    \includegraphics[width=\linewidth]{./figs/low_dims/mog_1d/sigma_2,00_v0_0,40_lambda_10,00.pdf}
                    \caption{\scriptsize MMD}
                    \label{subfig:appendix_sens_a}
                \end{subfigure}
                \begin{subfigure}[c]{0.24\textwidth}
                    \centering
                    \includegraphics[width=\linewidth]{./figs/low_dims/mog_1d/sigma_2,00_v0_0,40_lambda_Inf.pdf}
                    \caption{\scriptsize UME}
                    \label{subfig:appendix_sens_b}
                \end{subfigure}
                \caption{(a, b): Numerical solution of the differential equations governing the GD training of two means of a mixture of Gaussians when only one witness point is provided. The results are plotted for various values of the regularization weight $\lambda$. Color codes are the same as~\Cref{fig:appendix_low_dims_phase_portrait}}
                \label{fig:appendix_low_dims_sim_mog}
    \end{minipage}\\
    \begin{minipage}[t]{\textwidth}
        \centering
        \begin{subfigure}[c]{0.24\textwidth}
            \centering
            \includegraphics[width=\linewidth]{figs/low_dims/phase_portrait/phase_portrait_sigma_0,1.pdf}
            \caption{\scriptsize $\sigma$ =  $0.1$}
            \label{fig:appendix_phase_portrait_sigma_0.1_a}
        \end{subfigure}
        \begin{subfigure}[c]{0.24\textwidth}
            \centering
            \includegraphics[width=\linewidth]{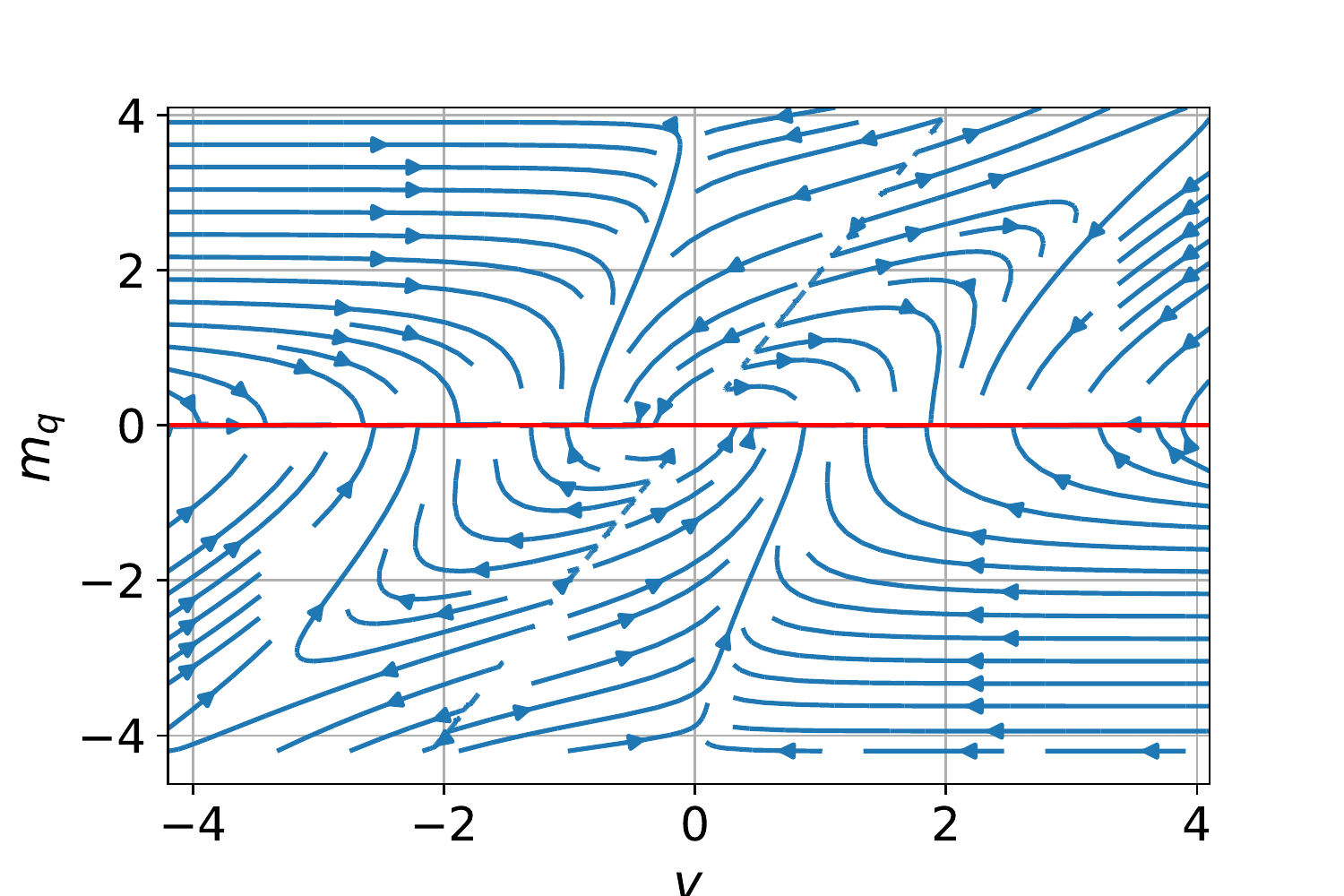}
            \caption{\scriptsize $\sigma$ =  $5$}
            \label{fig:appendix_phase_portrait_sigma_5_a}
        \end{subfigure}
        \begin{subfigure}[c]{0.24\textwidth}
            \centering
            \includegraphics[width=\linewidth]{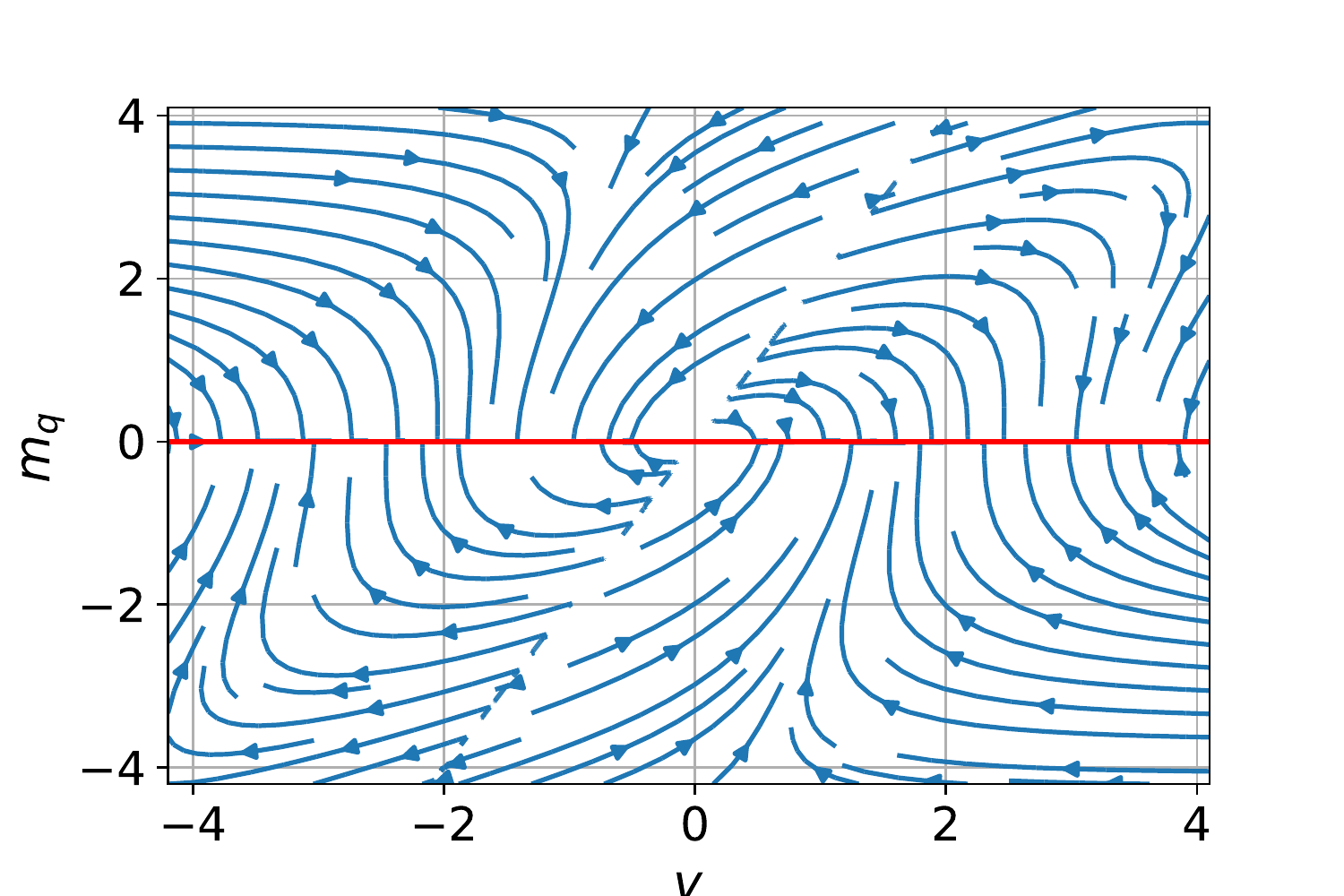}
            \caption{\scriptsize $\sigma$ =  $0.1$}
            \label{fig:appendix_phase_portrait_sigma_0.1_b}
        \end{subfigure}
        \begin{subfigure}[c]{0.24\textwidth}
            \centering
            \includegraphics[width=\linewidth]{figs/low_dims/phase_portrait/phase_portrait_sigma_5.pdf}
            \caption{\scriptsize $\sigma$ =  $5$}
            \label{fig:appendix_phase_portrait_sigma_5_b}
        \end{subfigure}
            \caption{Phase portrait of the 2-dimensional dynamical system of~\Cref{fig:appendix_low_dims_sim_fixed_sigmas} for various values of the kernel bandwidth $\sigma$. The plot shows two equilibria : stable (the horizontal red line) and unstable (the line oriented in $45^\circ$).}
    \label{fig:appendix_low_dims_phase_portrait}
    \end{minipage}    
\end{figure*}


\subsubsection{Single Gaussian}
\label{sec:appendix_single_gaussian}
Here, we provide a more comprehensive explanation of the single Gaussian experiment. In this simulation, we assume that $P_\Xb$ is a one-dimensional Gaussian $\mathcal{N}(0,
1)$ and the model $Q_{\theta_q}$
is also Gaussian $\mathcal{N}(m_q, \sigma_q)$ parameterised by
$\theta_q=\{m_q, \sigma_q\}$ where $\sigma_q
= 1$. 
Moreover, the algorithm has access to only a single witness point denoted by $v$.
To analyze the dynamics of the GD, we first compute the loss function $\mathcal{L}(\theta_q,v)$
and then derive the governing dynamics for each of these
parameters by calculating $d\theta_q / dt =
\shortminus\nabla_{\theta_q}(\mathcal{L})$ and $dv / dt =
\nabla_v(\mathcal{L})$.\footnote{See
Appendix.~\ref{subsec_appendix:ume_pq_single_gaussian} for the details.} 
Specifically, the loss function~\eqref{eq:objective_kernel} becomes an MMD distance between two
Gaussian distributions which is regularized by UME distance with $\lambda$ as the
regularization weight. 
Thus, the trainable parameters are $m_q$ and $v$
whose dynamics are governed by $\dot{v} =\nabla_{v}(\mu_{P}(v)-\mu_{\theta_q}(v))^{2}$ 
and
$\dot{m}_q
=-\nabla_{m_q}[\mathrm{MMD}^2(m_q, m_p=0)+\lambda(\mu_{P}(v)-\mu_{\theta_q}(v))^{2}]$. This dynamical system is then solved numerically. 
Since we are interested in the role of UME, we study the
effect of the governing dynamics on trajectories by changing the value of
$\lambda$.~\Cref{fig:appendix_low_dims_sim_fixed_sigmas} depicts the results of this analysis. As can be seen, the model parameter $m_q$ 
converges to the target value and the witness point settles on a
value different from the mean of the distributions. 
It is noteworthy that the final value of the witness point becomes irrelevant
when the two distributions match, i.e., $(\mu_{P}(v)-\mu_{\theta_q}(v))^{2}=0$
for all $v$ when $P=Q_{\theta_q}$.
Moreover, we can also see in~\Cref{fig:appendix_low_dims_sim_fixed_sigmas} that the speed of convergence is affected by the value of $\lambda$. 
The convergence gets slower for extremely low or large values of $\lambda$, suggesting that combining the global and local terms gives faster convergence than what could be achieved by either one alone.

\subsubsection{Mixture of Gaussians.}
Next, we investigate the dynamics of the parameters $\bm{\theta}_q = \{m_1,m_2\}$ of the model $Q_{\bm{\theta}_q} = 0.5\cdot\mathcal{N}(m_1,1) + 0.5\cdot\mathcal{N}(m_2,1)$ that aims to capture a target distribution when the model is given only a single witness point $v$. This experiment suggests that the number of witness points does not need to be equal to the modes of data. Since witness points are themselves trainable, their dynamics allows learning several modes by having a single witness point. The simulated dynamics of the parameters over the course of GD training is shown in~\Cref{fig:appendix_low_dims_sim_mog} . It was observed that the fastest convergence occurs for a middle-valued $\lambda$ that emphasizes the positive role of having UME along with the global term, which in this case is MMD (See~\Cref{fig:appendix_low_dims_sim_mog} in the appendix for further details).

\subsubsection{Phase Portrait}
\label{sec:appendix_portrait}
\noindent~\Cref{fig:appendix_low_dims_phase_portrait} depicts the \emph{phase portrait} of the dynamical system enforced by UME loss function in a region encompassing its equilibrium ($m_q=0$) for different values of the kernel bandwidth. As can be seen, the system has a continuum of equilibria (depicted by the red line) rather than an isolated equilibrium point. This phenomenon is proved for the general conditions briefly in Section~\ref{sec:theory} and in detail in Appendix~\ref{sec_appendix:equilibrium_existence_proof}. This continuum is a stable equilibrium as it has an enclosing vicinity that every trajectory starting in this region converges to the equilibrium. This observation is also proved for the general case in Section~\ref{sec:theory} and Appendix~\ref{sec_appendix:local_stability_proof}. It is noteworthy in~\Cref{fig:appendix_low_dims_phase_portrait} that the value of the kernel bandwidth changes the shape of the trajectories but leaves the locations of the equilibria unchanged. This observation suggests that the stability properties of the resulting dynamical system are not too sensitive to the choice of kernel.

\end{document}